\documentclass[english]{article}
\pdfoutput=1
\usepackage[T1]{fontenc}
\usepackage[latin9]{inputenc}
\usepackage{color}
\usepackage{babel}
\usepackage{float}
\usepackage{booktabs}
\usepackage{textcomp}
\usepackage{amsmath}
\usepackage{amsthm}
\usepackage{amssymb}
\usepackage{graphicx}
\usepackage[authoryear,round]{natbib}
\usepackage{xargs}[2008/03/08]
\usepackage[unicode=true]
 {hyperref}

\makeatletter

\providecommand{\tabularnewline}{\\}
\floatstyle{ruled}
\newfloat{algorithm}{tbp}{loa}
\providecommand{\algorithmname}{Algorithm}
\floatname{algorithm}{\protect\algorithmname}

\usepackage{algolyx}
\usepackage{algolyx}

\usepackage[accepted]{icml2020}

\usepackage{mathptmx}
\usepackage{algolyx}
\usepackage{microtype}
\usepackage{graphicx}
\usepackage{subfigure}
\usepackage{booktabs} 
\usepackage{flushend}
\usepackage{hyperref}

\usepackage{amsfonts}       
\usepackage{nicefrac}       
\icmltitlerunning{Knowing The What But Not The Where in Bayesian Optimization}

\makeatother

\begin{document}
\twocolumn[ 
\icmltitle{Knowing The What But Not The Where in Bayesian Optimization}

\begin{icmlauthorlist}
\icmlauthor{Vu Nguyen}{to} 
\hskip 0.4in
\icmlauthor{Michael A Osborne}{to} 
\end{icmlauthorlist}

\icmlaffiliation{to}{University of Oxford,  UK} 

\icmlcorrespondingauthor{Vu Nguyen}{vu@robots.ox.ac.uk} 

\vskip 0.3in
]
\printAffiliationsAndNotice{}

\emph{}

\global\long\def\se{\hat{\text{se}}}%

\global\long\def\interior{\text{int}}%

\global\long\def\boundary{\text{bd}}%

\global\long\def\new{\text{*}}%

\global\long\def\stir{\text{Stirl}}%

\global\long\def\dist{d}%

\global\long\def\HX{\entro\left(X\right)}%
 
\global\long\def\entropyX{\HX}%

\global\long\def\HY{\entro\left(Y\right)}%
 
\global\long\def\entropyY{\HY}%

\global\long\def\HXY{\entro\left(X,Y\right)}%
 
\global\long\def\entropyXY{\HXY}%

\global\long\def\mutualXY{\mutual\left(X;Y\right)}%
 
\global\long\def\mutinfoXY{\mutualXY}%

\global\long\def\xnew{y}%

\global\long\def\bx{\mathbf{x}}%

\global\long\def\bz{\mathbf{z}}%

\global\long\def\bu{\mathbf{u}}%

\global\long\def\bs{\boldsymbol{s}}%

\global\long\def\bk{\mathbf{k}}%

\global\long\def\bX{\mathbf{X}}%

\global\long\def\tbx{\tilde{\bx}}%

\global\long\def\by{\mathbf{y}}%

\global\long\def\bY{\mathbf{Y}}%

\global\long\def\bZ{\boldsymbol{Z}}%

\global\long\def\bU{\boldsymbol{U}}%

\global\long\def\bv{\boldsymbol{v}}%

\global\long\def\bn{\boldsymbol{n}}%

\global\long\def\bV{\boldsymbol{V}}%

\global\long\def\bK{\boldsymbol{K}}%

\global\long\def\bw{\vt w}%

\global\long\def\bbeta{\gvt{\beta}}%

\global\long\def\bmu{\gvt{\mu}}%

\global\long\def\btheta{\boldsymbol{\theta}}%

\global\long\def\blambda{\boldsymbol{\lambda}}%

\global\long\def\bgamma{\boldsymbol{\gamma}}%

\global\long\def\bpsi{\boldsymbol{\psi}}%

\global\long\def\bphi{\boldsymbol{\phi}}%

\global\long\def\bpi{\boldsymbol{\pi}}%

\global\long\def\eeta{\boldsymbol{\eta}}%

\global\long\def\bomega{\boldsymbol{\omega}}%

\global\long\def\bepsilon{\boldsymbol{\epsilon}}%

\global\long\def\btau{\boldsymbol{\tau}}%

\global\long\def\bSigma{\gvt{\Sigma}}%

\global\long\def\realset{\mathbb{R}}%

\global\long\def\realn{\realset^{n}}%

\global\long\def\integerset{\mathbb{Z}}%

\global\long\def\natset{\integerset}%

\global\long\def\integer{\integerset}%

\global\long\def\natn{\natset^{n}}%

\global\long\def\rational{\mathbb{Q}}%

\global\long\def\rationaln{\rational^{n}}%

\global\long\def\complexset{\mathbb{C}}%

\global\long\def\comp{\complexset}%

\global\long\def\compl#1{#1^{\text{c}}}%

\global\long\def\and{\cap}%

\global\long\def\compn{\comp^{n}}%

\global\long\def\comb#1#2{\left({#1\atop #2}\right) }%

\global\long\def\nchoosek#1#2{\left({#1\atop #2}\right)}%

\global\long\def\param{\vt w}%

\global\long\def\Param{\Theta}%

\global\long\def\meanparam{\gvt{\mu}}%

\global\long\def\Meanparam{\mathcal{M}}%

\global\long\def\meanmap{\mathbf{m}}%

\global\long\def\logpart{A}%

\global\long\def\simplex{\Delta}%

\global\long\def\simplexn{\simplex^{n}}%

\global\long\def\dirproc{\text{DP}}%

\global\long\def\ggproc{\text{GG}}%

\global\long\def\DP{\text{DP}}%

\global\long\def\ndp{\text{nDP}}%

\global\long\def\hdp{\text{HDP}}%

\global\long\def\gempdf{\text{GEM}}%

\global\long\def\ei{\text{EI}}%

\global\long\def\rfs{\text{RFS}}%

\global\long\def\bernrfs{\text{BernoulliRFS}}%

\global\long\def\poissrfs{\text{PoissonRFS}}%

\global\long\def\grad{\gradient}%
 
\global\long\def\gradient{\nabla}%

\global\long\def\cpr#1#2{\Pr\left(#1\ |\ #2\right)}%

\global\long\def\var{\text{Var}}%

\global\long\def\Var#1{\text{Var}\left[#1\right]}%

\global\long\def\cov{\text{Cov}}%

\global\long\def\Cov#1{\cov\left[ #1 \right]}%

\global\long\def\COV#1#2{\underset{#2}{\cov}\left[ #1 \right]}%

\global\long\def\corr{\text{Corr}}%

\global\long\def\sst{\text{T}}%

\global\long\def\SST{\sst}%

\global\long\def\ess{\mathbb{E}}%

\global\long\def\Ess#1{\ess\left[#1\right]}%

\newcommandx\ESS[2][usedefault, addprefix=\global, 1=]{\underset{#2}{\ess}\left[#1\right]}%

\global\long\def\fisher{\mathcal{F}}%

\global\long\def\bfield{\mathcal{B}}%
 
\global\long\def\borel{\mathcal{B}}%

\global\long\def\bernpdf{\text{Bernoulli}}%

\global\long\def\betapdf{\text{Beta}}%

\global\long\def\dirpdf{\text{Dir}}%

\global\long\def\gammapdf{\text{Gamma}}%

\global\long\def\gaussden#1#2{\text{Normal}\left(#1, #2 \right) }%

\global\long\def\gauss{\mathbf{N}}%

\global\long\def\gausspdf#1#2#3{\text{Normal}\left( #1 \lcabra{#2, #3}\right) }%

\global\long\def\multpdf{\text{Mult}}%

\global\long\def\poiss{\text{Pois}}%

\global\long\def\poissonpdf{\text{Poisson}}%

\global\long\def\pgpdf{\text{PG}}%

\global\long\def\wshpdf{\text{Wish}}%

\global\long\def\iwshpdf{\text{InvWish}}%

\global\long\def\nwpdf{\text{NW}}%

\global\long\def\niwpdf{\text{NIW}}%

\global\long\def\studentpdf{\text{Student}}%

\global\long\def\unipdf{\text{Uni}}%

\global\long\def\transp#1{\transpose{#1}}%
 
\global\long\def\transpose#1{#1^{\mathsf{T}}}%

\global\long\def\mgt{\succ}%

\global\long\def\mge{\succeq}%

\global\long\def\idenmat{\mathbf{I}}%

\global\long\def\trace{\mathrm{tr}}%

\global\long\def\argmax#1{\underset{_{#1}}{\text{argmax}} }%

\global\long\def\argmin#1{\underset{_{#1}}{\text{argmin}\ } }%

\global\long\def\diag{\text{diag}}%

\global\long\def\norm{}%

\global\long\def\spn{\text{span}}%

\global\long\def\vtspace{\mathcal{V}}%

\global\long\def\field{\mathcal{F}}%
 
\global\long\def\ffield{\mathcal{F}}%

\global\long\def\inner#1#2{\left\langle #1,#2\right\rangle }%
 
\global\long\def\iprod#1#2{\inner{#1}{#2}}%

\global\long\def\dprod#1#2{#1 \cdot#2}%

\global\long\def\norm#1{\left\Vert #1\right\Vert }%

\global\long\def\entro{\mathbb{H}}%

\global\long\def\entropy{\mathbb{H}}%

\global\long\def\Entro#1{\entro\left[#1\right]}%

\global\long\def\Entropy#1{\Entro{#1}}%

\global\long\def\mutinfo{\mathbb{I}}%

\global\long\def\relH{\mathit{D}}%

\global\long\def\reldiv#1#2{\relH\left(#1||#2\right)}%

\global\long\def\KL{KL}%

\global\long\def\KLdiv#1#2{\KL\left(#1\parallel#2\right)}%
 
\global\long\def\KLdivergence#1#2{\KL\left(#1\ \parallel\ #2\right)}%

\global\long\def\crossH{\mathcal{C}}%
 
\global\long\def\crossentropy{\mathcal{C}}%

\global\long\def\crossHxy#1#2{\crossentropy\left(#1\parallel#2\right)}%

\global\long\def\breg{\text{BD}}%

\global\long\def\lcabra#1{\left|#1\right.}%

\global\long\def\lbra#1{\lcabra{#1}}%

\global\long\def\rcabra#1{\left.#1\right|}%

\global\long\def\rbra#1{\rcabra{#1}}%

\begin{abstract}
Bayesian optimization has demonstrated impressive success in finding
the optimum input $\bx^{*}$ and output $f^{*}=f(\bx^{*})=\max f(\bx)$
of a black-box function $f$. In some applications, however, the optimum
output $f^{*}$ is known in advance and the goal is to find the corresponding
optimum input $\bx^{*}$. In this paper, we consider a new setting
in BO in which the knowledge of the optimum output $f^{*}$ is available.
Our goal is to exploit the knowledge about $f^{*}$ to search for
the input $\bx^{*}$ efficiently. To achieve this goal, we first transform
the Gaussian process surrogate using the information about the optimum
output. Then, we propose two acquisition functions, called confidence
bound minimization and expected regret minimization. We show that
our approaches work intuitively and give quantitatively  better performance
against standard BO methods. We demonstrate real applications in tuning
a deep reinforcement learning algorithm on the CartPole problem and
XGBoost on Skin Segmentation dataset in which the optimum values are
publicly available.

\end{abstract}

\section{Introduction}

Bayesian optimization (BO) \citep{Brochu_2010Tutorial,Shahriari_2016Taking,oh2018bock,cocabo}
is an efficient method for the global optimization of a black-box
function. BO has been successfully employed in selecting chemical
compounds \citep{Hernandez_2017Parallel}, material design \citep{Frazier_2016Bayesian,li2018accelerating_ICDM},
algorithmic assurance \citep{gopakumar2018algorithmic_NIPS}, and
in search for hyperparameters of machine learning algorithms \citep{Snoek_2012Practical,klein2017fast,chen2018bayesian}.
These recent results suggest BO is more efficient than manual, random,
or grid search.

Bayesian optimization finds the global maximizer $\bx^{*}=\arg\max_{\bx\in\mathcal{X}}f(\bx)$
of the black-box function $f$  by incorporating prior beliefs about
$f$ and updating the prior with evaluations where $\mathcal{X}\subset\mathcal{\mathbb{R}}^{d}$
is the search domain. The model used for approximating the black-box
function is called the surrogate model. A popular choice for a surrogate
model is the Gaussian process (GP) \citep{Rasmussen_2006gaussian}
although there are existing alternative options, such as random forests
\citep{Hutter_2011Sequential}, deep neural networks \citep{Snoek_2015Scalable},
Bayesian neural networks \citep{Springenberg_2016Bayesian} and Mondrian
trees \citep{wang2018batched}. This surrogate model is then used
to define an acquisition function which determines the next query
of the black-box function.

In some settings, the optimum output $f^{*}=f(\bx^{*})$ is known
in advance. For example, the optimal reward is available for common
reinforcement learning benchmarks or we know the optimum accuracy
is $100$ in tuning classification algorithm for specific datasets.
As another example in inverse optimization, we retrieve the input
resulting the given target \citep{ahuja2001inverse,perdikaris2016model}.
The question is how to efficiently utilize such prior knowledge to
find the optimal inputs  using the fewest number of queries.

In this paper, we give the first BO approach to this setting in which
we know what we are looking for, but we do not know where it is. Specifically,
we know the optimum output $f^{*}=\max_{\bx\in\mathcal{X}}f(\bx)$
and aim to search for the unknown optimum input $\bx^{*}=\arg\max_{\bx\in\mathcal{X}}f(\bx)$
by utilizing $f^{*}$ value.

We incorporate the information about $f^{*}$ into Bayesian optimization
in the following ways. First, we use the knowledge of $f^{*}$ to
build a transformed GP surrogate model. Our intuition in transforming
a GP is based on the fact that the black-box function value $f(\bx)$
should not be above the threshold $f^{*}$ (since $f^{*}\ge f(\bx),\forall\bx\in\mathcal{X}$,
by definition). As a result, the GP surrogate should also follow this
property. Second, we propose two acquisition functions which  make
 decisions informed by the $f^{*}$ value, namely \emph{confidence
bound minimization} and \emph{expected regret minimization}.

We validate our model using  benchmark functions and tuning a deep
reinforcement learning algorithm where we observe the optimum value
in advance. These experiments demonstrate that our proposed framework
works both intuitively better and experimentally outperforms the baselines.
Our main contributions are summarized as follows:
\begin{itemize}
\item a first study of Bayesian optimization for exploiting the known optimum
output $f^{*}$;
\item a transformed Gaussian process surrogate using the knowledge of $f^{*}$;
and 
\item two novel acquisition functions to efficiently select the optimum
location given $f^{*}$.
\end{itemize}

\section{Preliminaries}

In this section, we review some of the existing acquisition functions
from the Bayesian optimization literature which can readily incorporate
the known $f^{*}$ value. Then, we summarize the possible transformation
techniques used to control the Gaussian process using  $f^{*}$.

\subsection{Available acquisition functions for the known $f^{*}$\label{subsec:Available-acquisition-functions}}

Bayesian optimization uses an acquisition function to make a query.
Among many existing acquisition functions \citep{Hennig_2012Entropy,Hernandez_2014Predictive,Wang_2016Optimization,letham2019constrained,astudillo2019bayesian,Nguyen_ICDM2019},
we review two acquisition functions which can incorporate the known
optimum output $f^{*}$ directly in their forms. We then use the
two acquisition functions as the baselines for comparison.

\paragraph{Expected improvement with known incumbent $f^{*}$.}

EI \citep{Mockus_1978Application} considers the expectation over
the improvement function which is defined over the \emph{incumbent}
$\xi$ as $\mathbb{E}\left[I_{t}\left(\bx\right)\right]=\mathbb{E}\left[\max\left\{ 0,f\left(\bx\right)-\xi\right\} \right]$.
One needs to define the incumbent to improve upon. Existing research
has considered modifying this incumbent with various choices \citep{Wang_2014Theoretical,Berk_2018Exploration_ECML}.
The typical choice of the incumbent is the best observed value so
far in the observation set $\xi=\max_{y_{i}\in\mathcal{D}_{t-1}}y_{i}$
where $\mathcal{D}_{t-1}$ is the dataset upto iteration $t$. Given
the known optimum output $f^{*}$, one can readily use it as the incumbent,
i.e., setting $\xi=f^{*}$ to have the following forms:
\begin{align}
\alpha^{\textrm{EI}^{*}}\left(\bx\right) & =\sigma\left(\bx\right)\phi\left(z\right)+\left[\mu\left(\bx\right)-f^{*}\right]\Phi\left(z\right)\label{eq:EI_Acq}
\end{align}
 where $\mu(\bx)$ is the GP predictive mean, $\sigma(\bx)$ is the
GP predictive variance, $z=\frac{\mu\left(\bx\right)-f^{*}}{\sigma\left(\bx\right)}$,
$\phi$ is the standard normal p.d.f. and $\Phi$ is the c.d.f. 

\paragraph{Output entropy search with known $f^{*}$.}

The second group of acquisition functions, which are readily to incorporate
the known optimum, include several approaches gaining information
about the output, such as output-space PES \citep{Hoffman_2015Output},
MES \citep{Wang_2017Max} and FITBO \citep{ru2018fast}. These approaches
consider different ways to gain information about the optimum output
$f^{*}$. When $f^{*}$ is not known in advance, \citet{Hoffman_2015Output,Wang_2017Max}
utilize Thompson sampling to sample $f^{*}$, or a collection of $f_{m}^{*},\forall m\le M$,
while \citet{ru2018fast} consider $f^{*}$ as a hyperparameter. After
generating optimum value samples, the above approaches consider different
approximation strategies. 

Since the optimum output $f^{*}$ is available in our setting, we
can use it directly within the above approaches. We select to review
the MES due to its simplicity and closed-form computation. Given the
known $f^{*}$ value, MES approximates $I(\bx,y;f^{*})$ using a truncated
Gaussian distribution such that the distribution of $y$ needs to
satisfy $y<f^{*}$, to obtain,
\begin{align*}
I(\bx,y;f^{*}) & =H\left[p(y|D_{t},\bx)\right]-\mathbb{E}[H\left(p(y|D_{t},\bx,f^{*})\right)]{p(f^{*}|D_{t})}.
\end{align*}
Let  $\gamma(\bx,f^{*})=\frac{f^{*}-\mu(\bx)}{\sigma(\bx)}$, we
have the MES$^{*}$ as
\begin{align*}
\alpha^{\textrm{MES}{}^{*}}(\bx\mid f^{*})= & \frac{\gamma(\bx,f^{*})\phi\left[\gamma(\bx,f^{*})\right]}{2\Phi\left[\gamma(\bx,f^{*})\right]}-\log\Phi\left[\gamma(\bx,f^{*})\right].
\end{align*}

\subsection{Gaussian process transformation for $f\le f^{*}$\label{subsec:GP_transformation}}

\begin{figure*}[t]
\begin{centering}
\includegraphics[width=1\columnwidth]{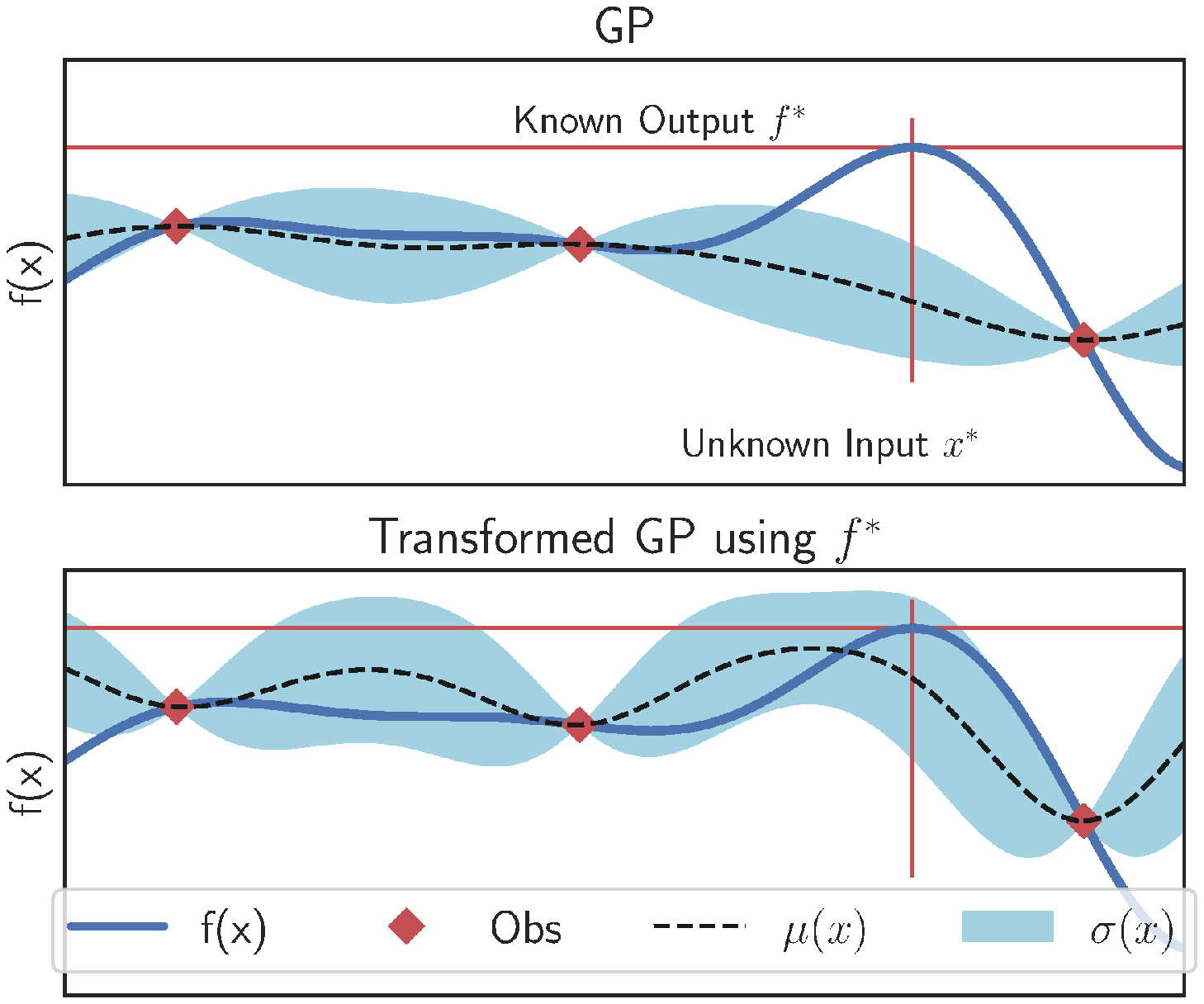}\hfill{}\includegraphics[width=1\columnwidth]{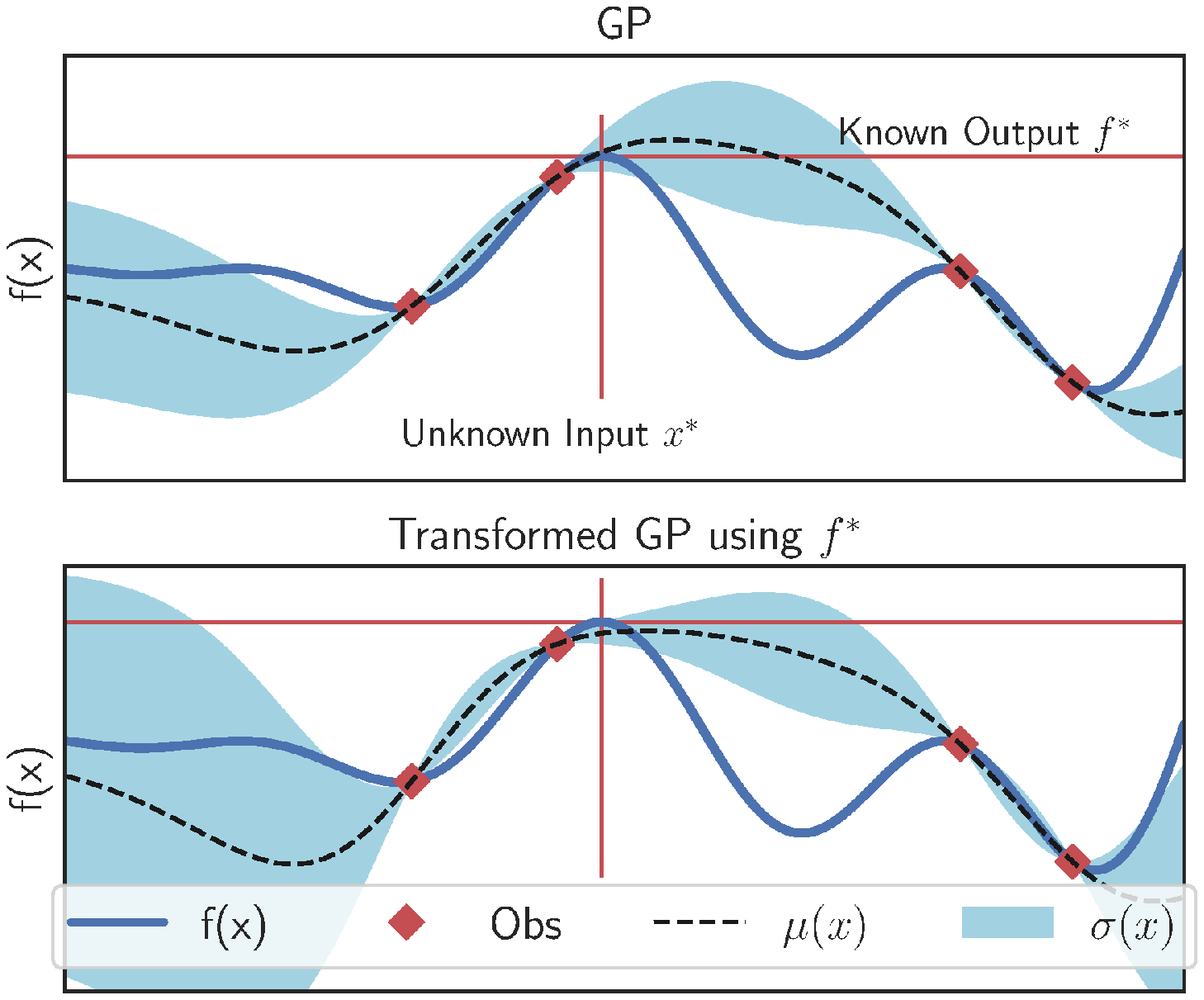}
\par\end{centering}
\caption{Comparison of the transformed GP with the GP using two different functions
in left and right. The known $f^{*}$ output and unknown input $\protect\bx^{*}$
are highlighted by horizontal and vertical \textcolor{red}{red} lines
respectively. Top: the GP allows $\mu(\protect\bx)$ to go above and
below $f^{*}$. Bottom: the transformed GP will lift up the surrogate
model closer to the known optimum output $f^{*}$ (left) and not go
above $f^{*}$ (right).\label{fig:Comparison_GP_TGP}}

\vspace{-0pt}
\end{figure*}
We summarize several transformation approaches which can be potentially
used to enforce that the function $f$ is everywhere below $f^{*}$,
given the upper bound $f^{*}=\max_{\forall\bx}f(\bx)$.

The first category is to use functions such as sigmoid and tanh. However,
there are two problems with such functions. The first problem is that
they both require the knowledge of the lower bound, $\min f(\bx)$,
and the upper bound, $\max f(\bx)$, for the normalization to the
predefined ranges, i.e. $[0,1]$ for sigmoid and $[-1,1]$ for tanh.
However, we do not know the lower bound in our setting. The second
problem is that exact inference for a GP is analytically intractable
under these transformations. Particularly, this will become the Gaussian
process classification problem \citep{nickisch2008approximations}
where approximation must be made, such as using expectation propagation
\citep{kuss2005assessing,riihimaki2013nested,hernandez2016scalable}.

The second category is to transform the output of a GP using warping
\citep{mackay1998introduction,snelson2004warped}. However, the warped
GP is less efficient in the context of Bayesian optimization. This
is because a warped GP requires more data points\footnote{using the datasets with 800 to 1000 samples for learning.}
to learn the mapping from original to transformed space while we only
have a small number of observations in BO setting.

The third category makes use of a linearization trick \citep{osborne2012active,gunter2014sampling}
as GPs are closed under linear transformations. This linearization
ensures that we arrive at another GP after transformation given our
existing GP. In this paper, we shall follow this linearization trick
to transform the surrogate model given $f^{*}$.

\section{Bayesian Optimization When The True Optimum Output Is Known}

\begin{figure}[t]
\begin{centering}
\includegraphics[width=1\columnwidth]{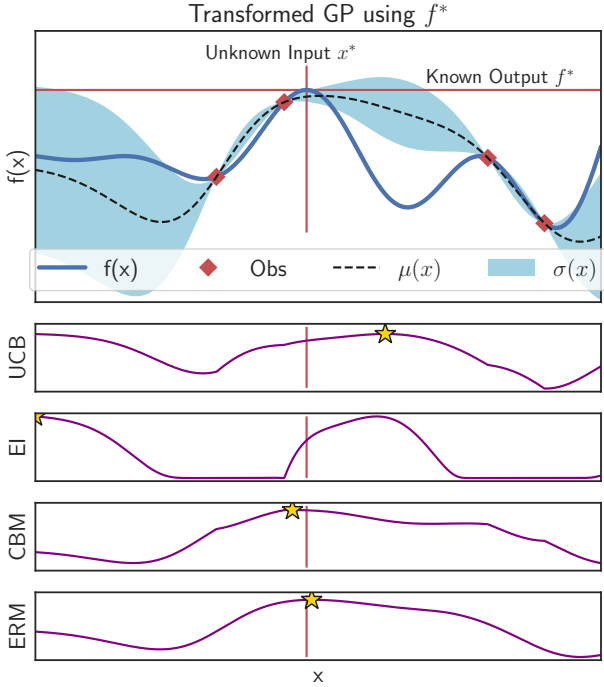}
\par\end{centering}
\caption{Illustration of the proposed acquisition functions CBM and ERM. A
yellow star indicates the maximum of the acquisition function and
thus is the selected point. Using the knowledge of $f^{*}$, CBM
and ERM can better identify $\protect\bx^{*}$ while EI and UCB cannot.
\label{fig:Illustration-of_ERM}}

\vspace{-0pt}
\end{figure}
We present a new approach for Bayesian optimization given situations
where the knowledge of optimum output (value) $f^{*}=\max_{\bx\in\mathcal{X}}f(\bx)$
is available. Our goal is to utilize this knowledge to improve BO
performance in finding the unknown optimum input (location) $\bx^{*}=\arg\max_{\bx\in\mathcal{X}}f(\bx)$.
We first encode $f^{*}$ to build an informed GP surrogate model through
transformation and then we propose two acquisition functions which
effectively exploit knowledge of $f^{*}$.

\subsection{Transformed Gaussian process \label{subsec:Transformed-Gaussian-Process}}

We make use of the knowledge about the optimum output to control the
GP surrogate model through transformation. Our transformation starts
with two key observations that firstly the function value $f(\bx)$
should reach the optimum output; but secondly never be greater than
the optimal value $f^{*}$, by definition of $f^{*}$ being a maximum
value. Therefore, the desired GP surrogate should not go above this
threshold. Based on this intuition, we propose the GP transformation
given $f^{*}$ as follows

\vspace{-17pt}

\begin{minipage}[t]{0.48\columnwidth}%
\begin{align*}
f(\bx) & =f^{*}-\frac{1}{2}g^{2}(\bx)
\end{align*}
\end{minipage}%
\begin{minipage}[t]{0.48\columnwidth}%
\begin{align*}
g(\bx) & \sim GP(m_{0},K).
\end{align*}
\end{minipage}

Our above transformation avoids the potential issues described in
Sec. \ref{subsec:GP_transformation}. That is we don't need  a lot
of samples to learn the transformation mapping for the desired property
that the function is always held $f^{*}\ge f(\bx),\forall\bx\in\mathcal{X}$
as $g^{2}(\bx)\ge0$. The prior mean for $g(\bx)$ can be used either
$m_{0}=0$ or $m_{0}=\sqrt{2f^{*}}$. These choices will bring two
different effects. A zero mean prior $m_{0}=0$ will tend to lift
up the surrogate model closer to $f^{*}$ as $f(\bx)=f^{*}$ when
$g(\bx)=0$. On the other hand, non-zero mean $m_{0}=\sqrt{2f^{*}}$
will encourage the mean prior of $f$ closer to zero -- as a common
practice in GP modeling where the output is standardized around zero
$y\sim\mathcal{N}(0,1)$. 

Given the observations $\mathcal{D}_{f}=(\bx_{i},y_{i})_{i=1}^{N}$,
we can compute the observations for $g$, i.e., $\mathcal{D}_{g}=(\bx_{i},g_{i})_{i=1}^{N}$
where $g_{i}=\sqrt{2(f^{*}-y_{i})}$. Then, we can write the posterior
of $p\left(g\mid\mathcal{D}_{g},f^{*}\right)\sim\mathcal{N}\left(\mu_{g}(\bx),\sigma_{g}(\bx)\right)$
as $\mu_{g}(\bx)=m_{0}+\mathbf{k}_{*}\mathbf{K}^{-1}\left(\mathbf{g}-m_{0}\right)$
and $\sigma_{g}^{2}\left(\bx\right)=k_{**}-\mathbf{k}_{*}\mathbf{K}^{-1}\mathbf{k}_{*}^{T}$
where $m_{0}$ is the prior mean of $g(\bx)$.

\begin{figure*}
\centering{}\includegraphics[width=0.66\columnwidth]{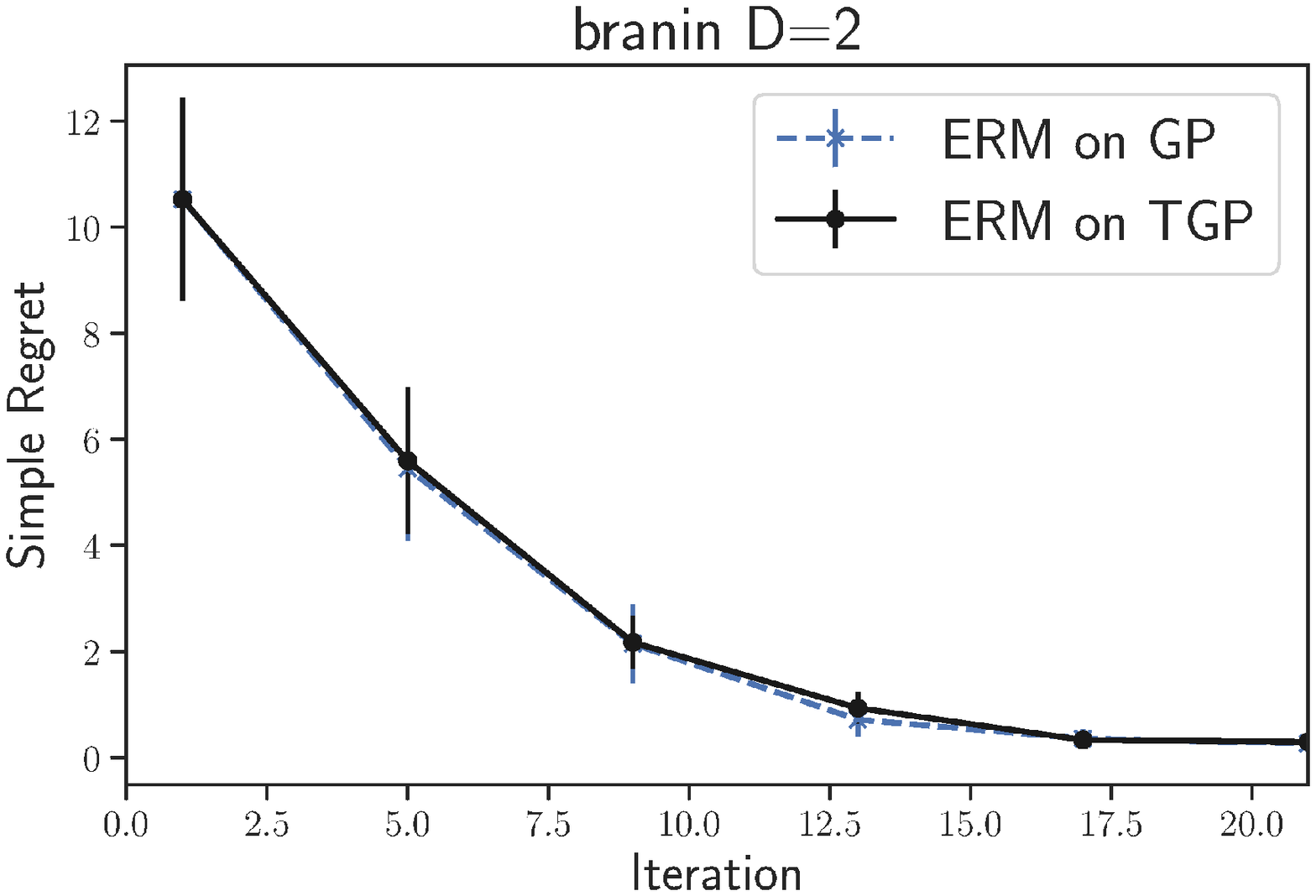}\includegraphics[width=0.67\columnwidth]{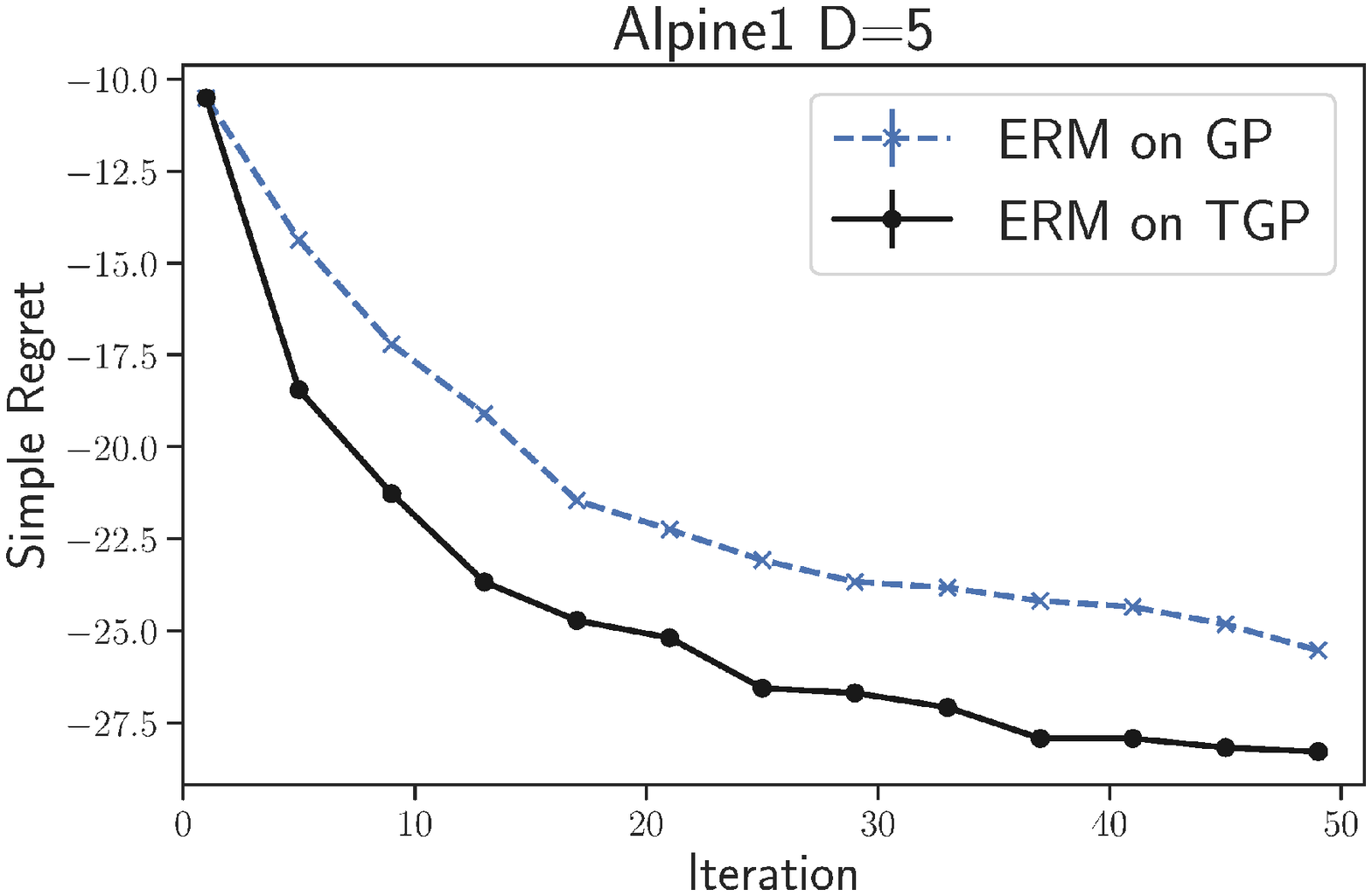}\includegraphics[width=0.66\columnwidth]{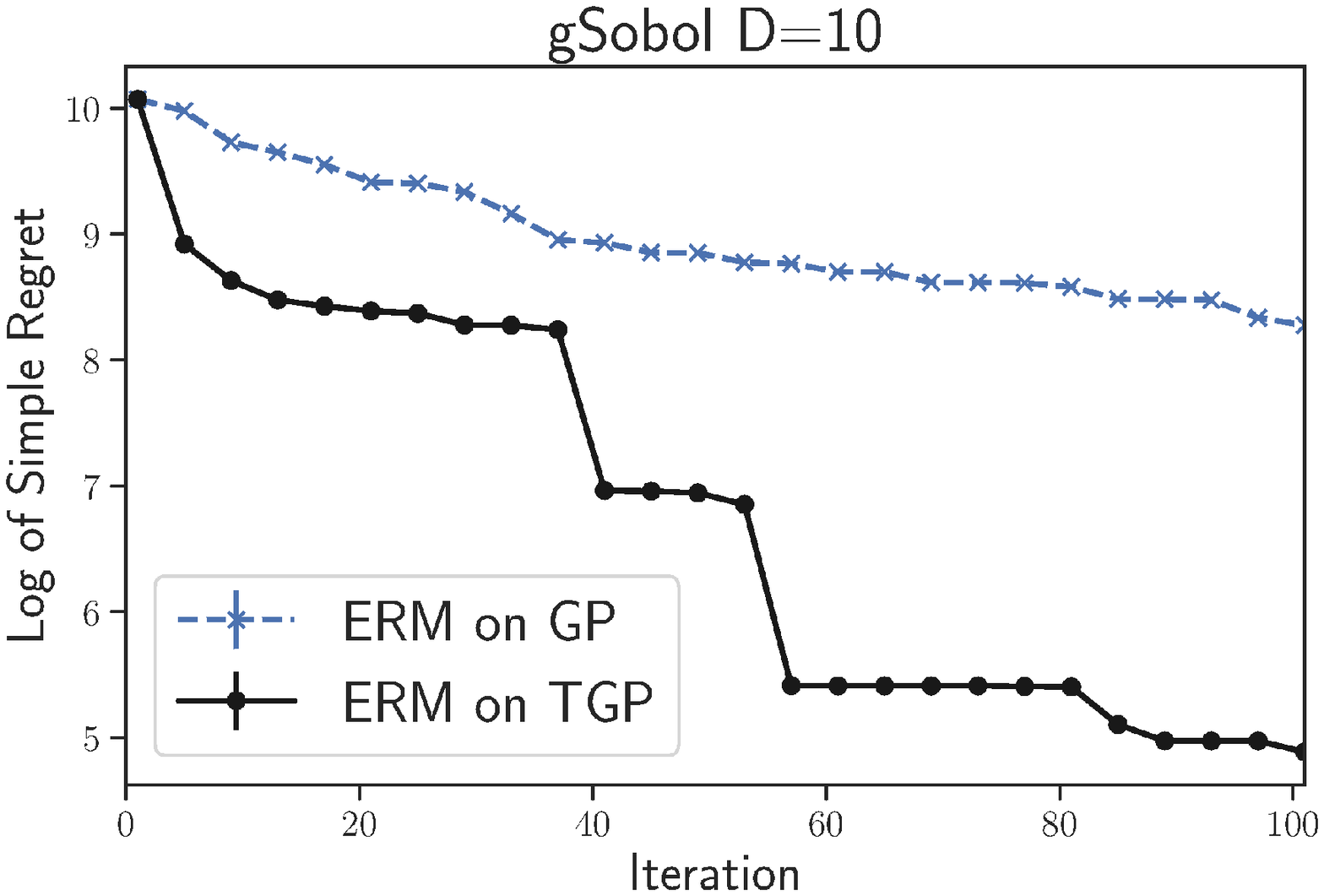}\caption{We show that our model performs much better using transformed Gaussian
process (TGP) than the vanilla GP. The knowledge of $f^{*}$ is useful
to inform the surrogate model for better optimization, especially
in high dimensional functions.\label{fig:Transforming-the-GPResults}}
\end{figure*}
We don\textquoteright t introduce any extra parameter for the above
transformation. However, the transformation causes the distribution
for any $f$ to become a non-central $\chi^{2}$ process, making the
analysis intractable. To tackle this problem and obtain a posterior
distribution $p\left(f\mid\mathcal{D}_{f},f^{*}\right)$ that is also
Gaussian, we employ an approximation technique presented in \citet{gunter2014sampling,ru2018fast}.
That is, we perform a local linearization of the transformation $h(g)=f^{*}-\frac{1}{2}g^{2}(\bx)$
around $g_{0}$ and obtain $f\approx h\left(g_{0}\right)+h'\left(g_{0}\right)\left(g-g_{0}\right)$
where the gradient $h'(g_{0})=-g$. Following \citet{gunter2014sampling,ru2018fast},
we set $g_{0}=\mu_{g}$ to the mode of the posterior distribution
$p\left(g\mid.\right)$ and obtain an expression for $f$ as
\begin{align*}
f(\bx) & \approx f^{*}-\frac{1}{2}\mu_{g}^{2}(\bx)-\mu_{g}(\bx)\left[g(\bx)\text{\textminus}\mu_{g}(\bx)\right]\\
 & =f^{*}+\frac{1}{2}\mu_{g}^{2}(\bx)-\mu_{g}(\bx)g(\bx).
\end{align*}

We have considered the mode $g_{0}$ of linear approximation to be
the multivariate function $\mu_{g}(\bx),\forall\bx$. As the property
of Taylor expansion, the approximation is very good at the mode $g_{0}$
and thus $\mu_{g}$. Since the linear transformation of a Gaussian
process remains Gaussian, the predictive posterior distribution for
$f$ now has a closed form for $p\left(f\mid.\right)=\mathcal{N}\left(f\mid\mu,\sigma\right)$
where the predictive mean and variance are given by
\begin{align}
\mu(\bx) & =f^{*}-\frac{1}{2}\mu_{g}^{2}(\bx),\label{eq:mu_f}\\
\sigma(\bx) & =\mu_{g}(\bx)\sigma_{g}(\bx)\mu_{g}(\bx).\label{eq:sigma_f}
\end{align}

These Eqs. (\ref{eq:mu_f}) and (\ref{eq:sigma_f}) are the key to
compute our acquisition functions in the next sections. As the effect
of transformation, the predictive uncertainty $\sigma(\bx)$ of the
transformed GP becomes larger than in the case of vanilla GP at the
location where $\mu(\bx)$ is low. This is because $\mu_{g}(\bx)$
is high when $\mu(\bx)$ is low and thus $\sigma(\bx)$ is high in
Eq. (\ref{eq:sigma_f}). This property may let other acquisition functions
(e.g., UCB, EI) explore more aggressively than they should. We further
examine these effects in the supplement.

We visualize the property of our transformed GP and compare with the
vanilla GP in Fig. \ref{fig:Comparison_GP_TGP}. By transforming the
GP using $f^{*}$, we encode the knowledge about $f^{*}$ into the
surrogate model, and thus are able to enforce that the surrogate model
gets close to but never above $f^{*}$, as desired, unlike the vanilla
GP. In the supplement, we provide further illustration   that transforming
the surrogate model  can help to find the optimum faster. We present
quantitative comparison of our transformed GP and vanilla GP in Fig.
\ref{fig:Transforming-the-GPResults} and in the supplement.

\subsection{Confidence bound minimization}

In this section, we introduce confidence bound minimization (CBM)
to efficiently select the (unknown) optimum location $\bx^{*}$ given
$f^{*}=f(\bx^{*})$. Our idea is based on the underlying concept of
GP-UCB \citep{Srinivas_2010Gaussian}. We consider the GP surrogate
at any location $\bx\in\mathcal{X}$ w.h.p.
\begin{align}
\mu(\bx)-\sqrt{\beta_{t}}\sigma(\bx) & \le f(\bx)\le\mu(\bx)+\sqrt{\beta_{t}}\sigma(\bx)\label{eq:GP_UCB}
\end{align}
where $\beta_{t}$ is a hyperparameter. Given the knowledge of $f^{*}$,
we can express this property at the optimum location $\bx^{*}$ where
$f^{*}=f(\bx^{*})$ to have w.h.p.
\begin{align*}
\mu(\bx^{*})-\sqrt{\beta_{t}}\sigma(\bx^{*}) & \le f^{*}\le\mu(\bx^{*})+\sqrt{\beta_{t}}\sigma(\bx^{*}).
\end{align*}
This is equivalent to write $|\mu(\bx^{*})-f^{*}|\le\sqrt{\beta_{t}}\sigma(\bx^{*})$.
Therefore, we can find the next point  $\bx_{t}$ by balancing the
posterior mean being close to the known optimum $f^{*}$ with having
low variance. That is
\begin{align*}
\alpha_{t}^{\textrm{CBM}}(\bx) & =|\mu(\bx)-f^{*}|+\sqrt{\beta_{t}}\sigma(\bx)
\end{align*}
 where $\mu(\bx)$ and $\sigma(\bx)$ are the GP mean and variance
 from Eq. (\ref{eq:mu_f}) and Eq. (\ref{eq:sigma_f}) respectively.
We select the next point by taking
\begin{align}
\bx_{t+1}= & \argmin{\bx\in\mathcal{X}}\alpha_{t}^{\textrm{CBM}}(\bx).\label{eq:MCB}
\end{align}

In the above objective function, we aim to quickly locate the area
potentially containing an optimum. Since the acquisition function
is non-negative, $\alpha^{\textrm{CBM}}(\bx)\ge0,\forall\bx\in\mathcal{X}$,
it takes the minimum value at the ideal location where $\mu(\bx_{t})=f^{*}$
and $\sigma(\bx_{t})=0$. When these two conditions are met, we can
conclude that $f(\bx_{t})=f(\bx^{*})$ and thus $\bx_{t}$ is what
we are looking for, as the property of Eq. (\ref{eq:GP_UCB}).

Because the CBM involves a hyperparameter $\beta$ to which performance
can be sensitive, we below propose another acquisition function incorporating
the knowledge of $f^{*}$ using no hyperparameter.

\subsection{Expected regret minimization}

We next develop our second acquisition function using $f^{*}$, called
expected regret minimization (ERM). We start with the regret function
$r\left(\bx\right)=f^{*}-f(\bx)$. The probability of regret $r\left(\bx\right)$
on a normal posterior distribution is as follows
\begin{align}
p\left(r\right)= & \frac{1}{\sqrt{2\pi}\sigma\left(\bx\right)}\exp\left(-\frac{1}{2}\frac{\left[f^{*}-\mu\left(\bx\right)-r(\bx)\right]^{2}}{\sigma^{2}\left(\bx\right)}\right).\label{eq:ELI_llk_improvement}
\end{align}
As the end goal in optimization is to minimize the regret, we consider
our acquisition function to minimize this expected regret as $\alpha^{\textrm{ERM}}\left(\bx\right)=\mathbb{E}\left[r\left(\bx\right)\right]$.
Using the likelihood function in Eq. (\ref{eq:ELI_llk_improvement}),
we write the expected regret minimization acquisition function as
\begin{align*}
\mathbb{E}\left[r\left(\bx\right)\right]= & \int\frac{r}{\sqrt{2\pi}\sigma\left(\bx\right)}\exp\left(-\frac{1}{2}\frac{\left[f^{*}-\mu\left(\bx\right)-r(\bx)\right]^{2}}{\sigma^{2}\left(\bx\right)}\right)dr.
\end{align*}
Let $z=\frac{f^{*}-\mu\left(\bx\right)}{\sigma\left(\bx\right)}$,
we obtain the closed-form computation as
\begin{align}
\alpha^{\textrm{ERM}}\left(\bx\right) & =\sigma\left(\bx\right)\phi\left(z\right)+\left[f^{*}-\mu\left(\bx\right)\right]\Phi\left(z\right)\label{eq:AcqFunc_ELI}
\end{align}
where $\phi\left(z\right)$ and $\Phi\left(z\right)$ are the standard
normal p.d.f. and c.d.f., respectively. To select the next point,
we minimize this acquisition function which is equivalent to minimizing
the expected regret,
\begin{align}
\bx_{t+1} & =\arg\min_{\bx\in\mathcal{X}}\alpha^{\textrm{ERM}}\left(\bx\right)=\arg\min_{\bx\in\mathcal{X}}\mathbb{E}\left[r\left(\bx\right)\right].\label{eq:ERM}
\end{align}
Our choice in Eq. (\ref{eq:ERM}) is where to minimize the expected
regret. We can see that this acquisition function is always positive
$\alpha^{\textrm{ERM}}(\bx)\ge0,\forall\bx\in\mathcal{X}$. It is
minimized at the ideal location $\bx_{t}$, i.e., $\alpha^{\textrm{ERM}}(\bx_{t})=\mathbb{E}\left[r\left(\bx\right)\right]=0$,
when $f^{*}-\mu(\bx_{t})=0$ and $\sigma(\bx_{t})=0$. This case happens
at the desired location  where the GP predictive value is equal to
the true $f^{*}$ with zero GP uncertainty.

\begin{algorithm}[t]
\caption{BO with known optimum output.\label{alg:BO_KOV}}

\begin{algor}
\item [{{*}}] Input: \#iter $T$, optimum value $f^{*}=\max_{\bx\in\mathcal{X}}f(\bx)$
\end{algor}
\begin{algor}[1]
\item [{while}] $t\le T$ and $f^{*}>\max_{\forall y_{i}\in D_{t}}y_{i}$
\item [{{*}}] Construct a transformed Gaussian process surrogate model
from $\mathcal{D}_{t}$ and $f^{*}$.
\item [{{*}}] Estimating $\mu$ and $\sigma$ from Eqs. (\ref{eq:mu_f})
and (\ref{eq:sigma_f}).
\item [{{*}}] Select $\bx_{t}=\arg\min_{\bx\in\mathcal{X}}\alpha_{t}^{\textrm{ERM}}(\bx)$,
or $\alpha_{t}^{\textrm{CBM}}(\bx)$, using the above transformed
GP model.
\item [{{*}}] Evaluate $y_{t}=f\left(\bx_{t}\right)$, set $g_{t}=\sqrt{2(f^{*}-y_{t})}$
and augment $\mathcal{D}_{t}=\mathcal{D}_{t-1}\cup\left(\bx_{t},y_{t},g_{t}\right)$.
\item [{endwhile}]~
\end{algor}
\end{algorithm}

Although our ERM is inspired by the  EI in the way that we define
the regret function and take the expectation, the resulting approach
is different in the following. The original EI strategy is to balance
exploration and exploitation, i.e., prefers high GP mean and high
GP variance. On the other hand, ERM will not encourage such trade-off
directly. Instead, ERM selects the point to minimize the expected
regret $\mathbb{E}\left[f^{*}-f(\bx)\right]$ with $\mu(\bx)$ being
closer to the known $f^{*}$ while having low variance to make sure
that the GP estimation at our chosen location is correct. Then, if
the chosen location turns out to be not expected (e.g., poor function
value), the GP is updated and ERM will move to another place which
minimizes the new expected regret. Therefore, these behaviors of
EI and our ERM are radically different.

\paragraph{Algorithm.}

We summarize all steps in Algorithm \ref{alg:BO_KOV}. Given the original
observation $\left\{ \bx_{i},y_{i}\right\} _{i=1}^{N}$ and $f^{*}$,
we compute $g_{i}=\sqrt{2(f^{*}-y_{i})}$, then build a transformed
GP using $\{\bx_{i},g_{i}\}_{i=1}^{N}$. Using a transformed GP, we
can predict the mean $\mu(\bx)$ and uncertainty $\sigma(\bx)$ at
any location $\bx$ from Eqs. (\ref{eq:mu_f}) and (\ref{eq:sigma_f})
which are used to compute the CBM and ERM acquisition functions in
Eq. (\ref{eq:MCB}) and Eq. (\ref{eq:ERM}). Our formulas are in closed-forms
and the algorithm is easy to implement. In addition, our computational
complexity is as cheap as the GP-UCB and EI.
\begin{figure*}[t]
\begin{centering}
\includegraphics[width=1\columnwidth]{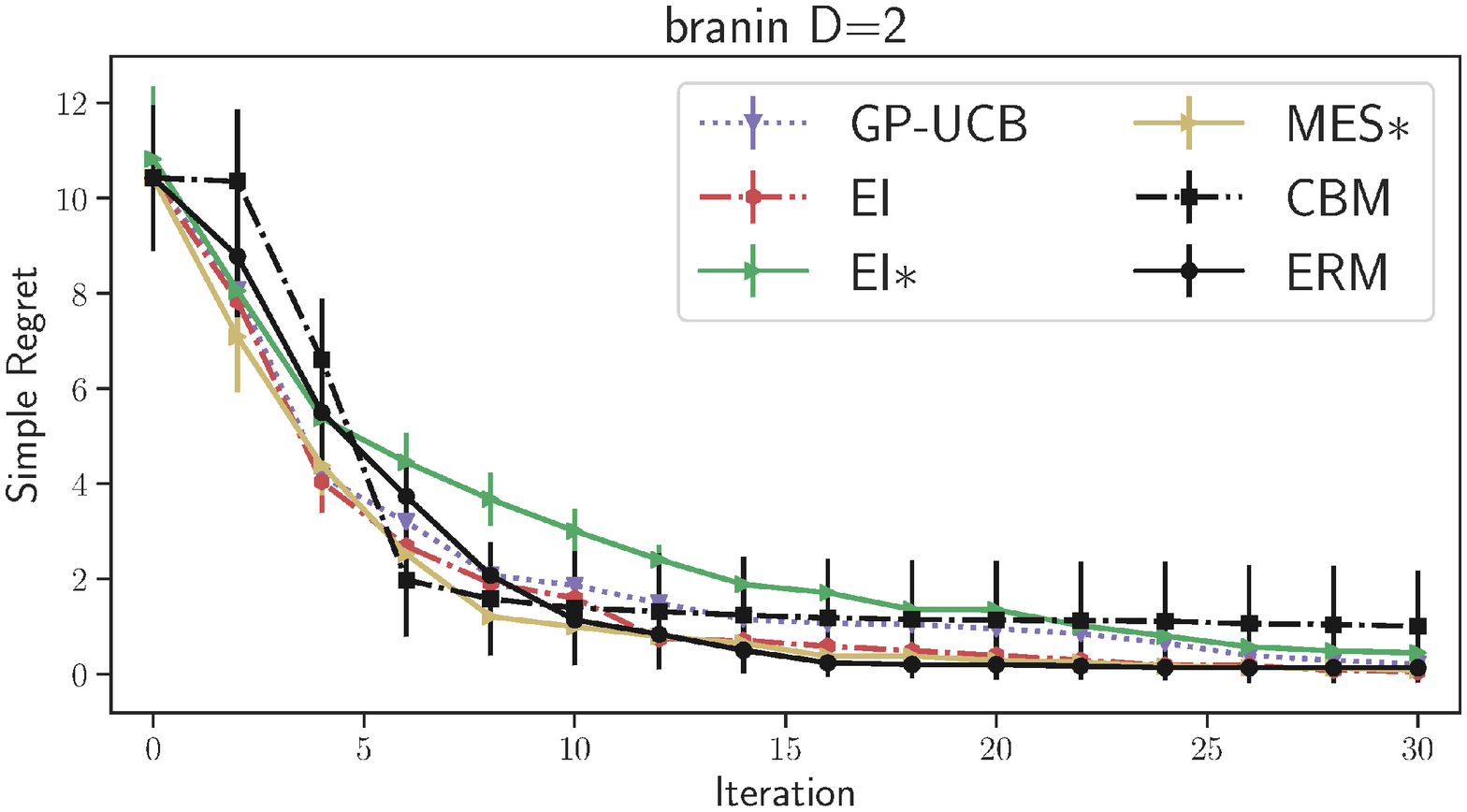}\includegraphics[width=1\columnwidth]{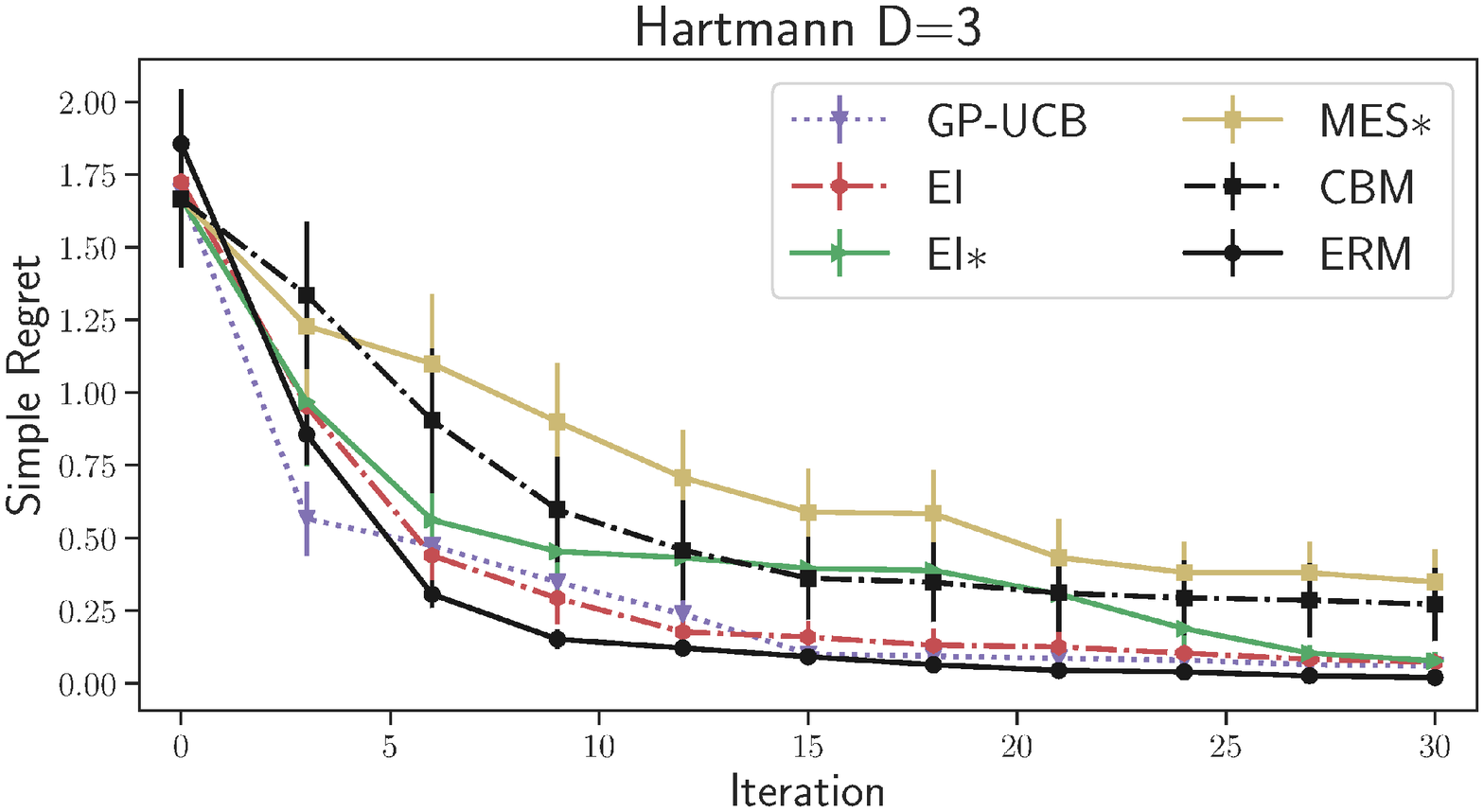}
\par\end{centering}
\begin{centering}
\includegraphics[width=1\columnwidth]{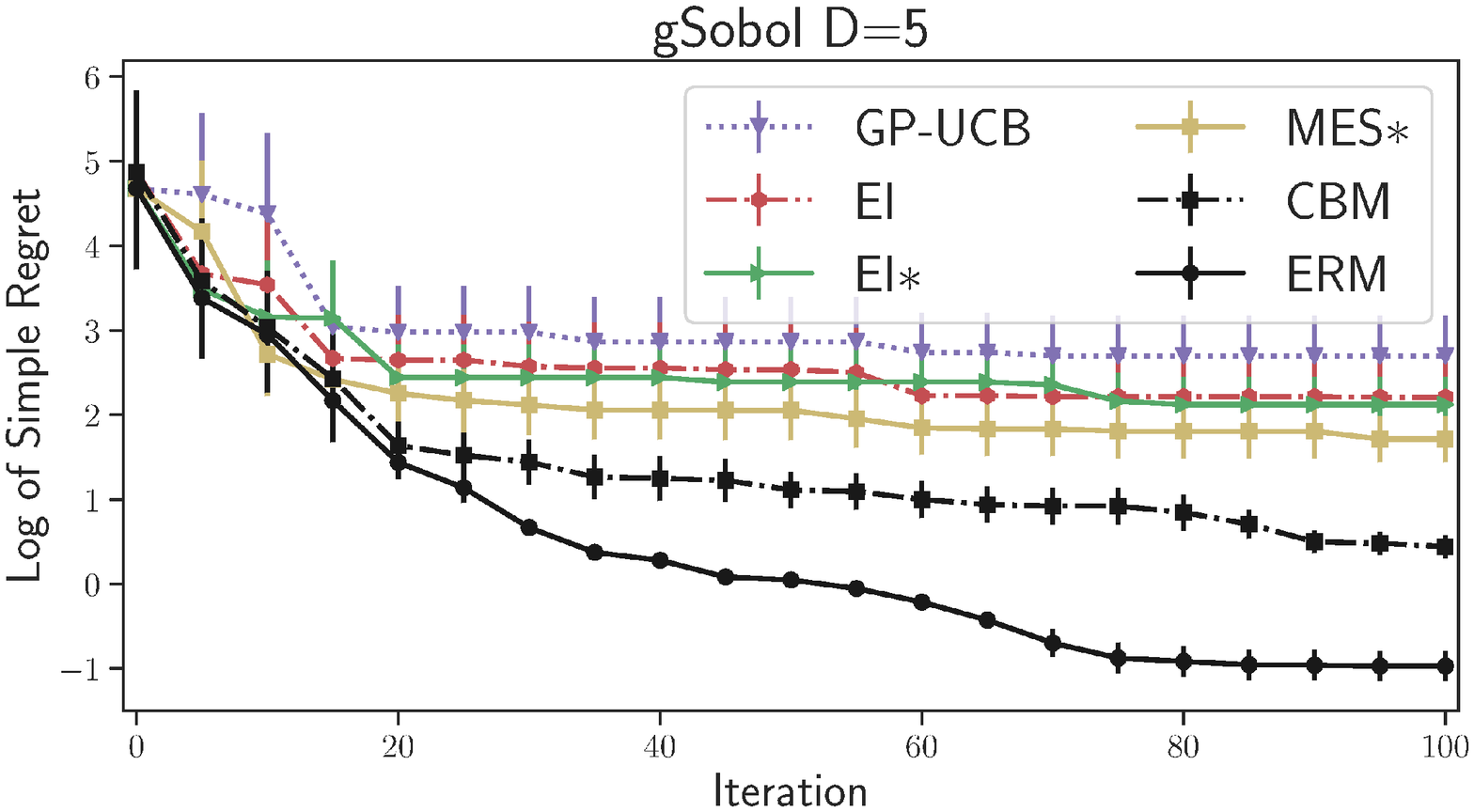}\includegraphics[width=1\columnwidth]{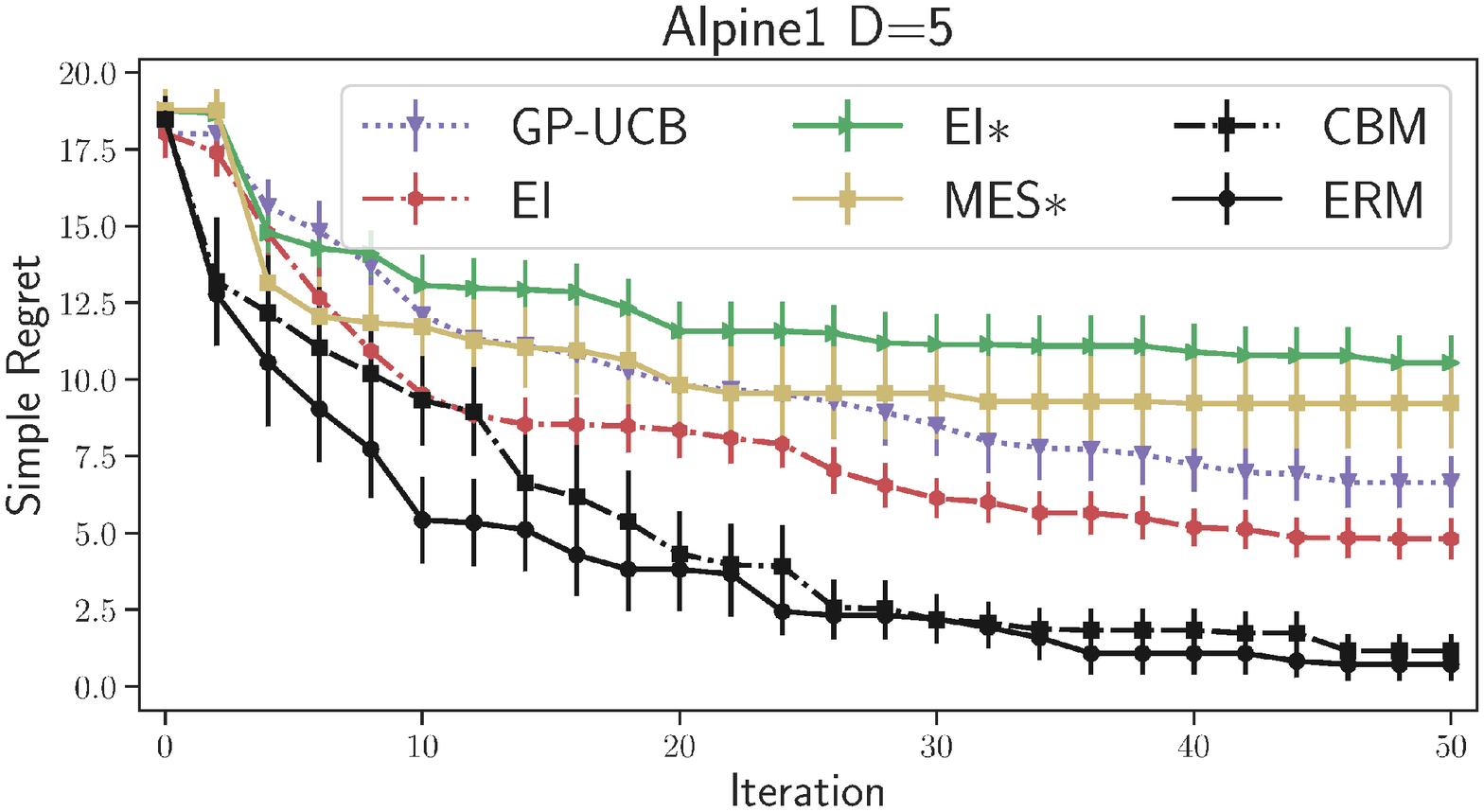}
\par\end{centering}
\begin{centering}
\includegraphics[width=1\columnwidth]{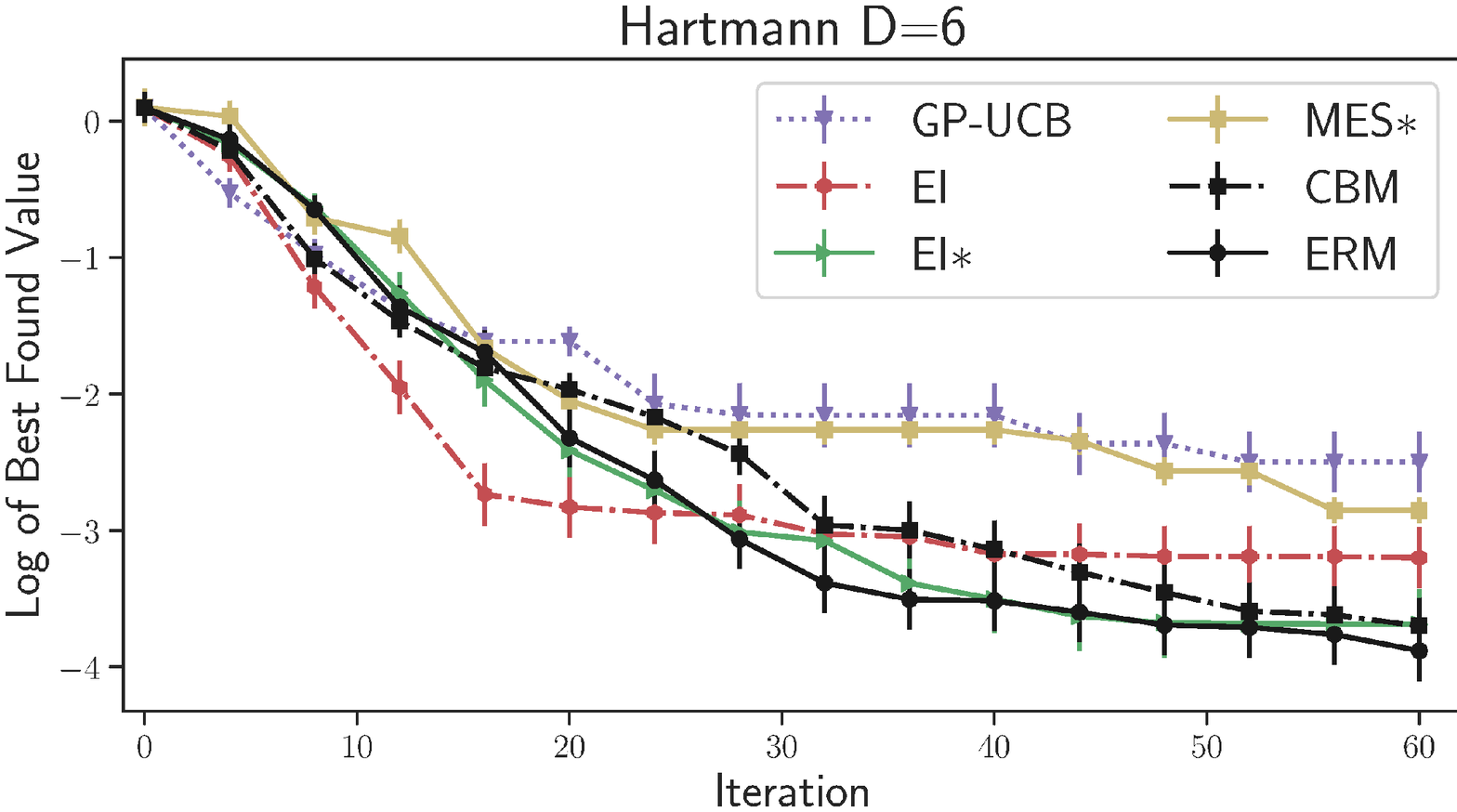}\includegraphics[width=1\columnwidth]{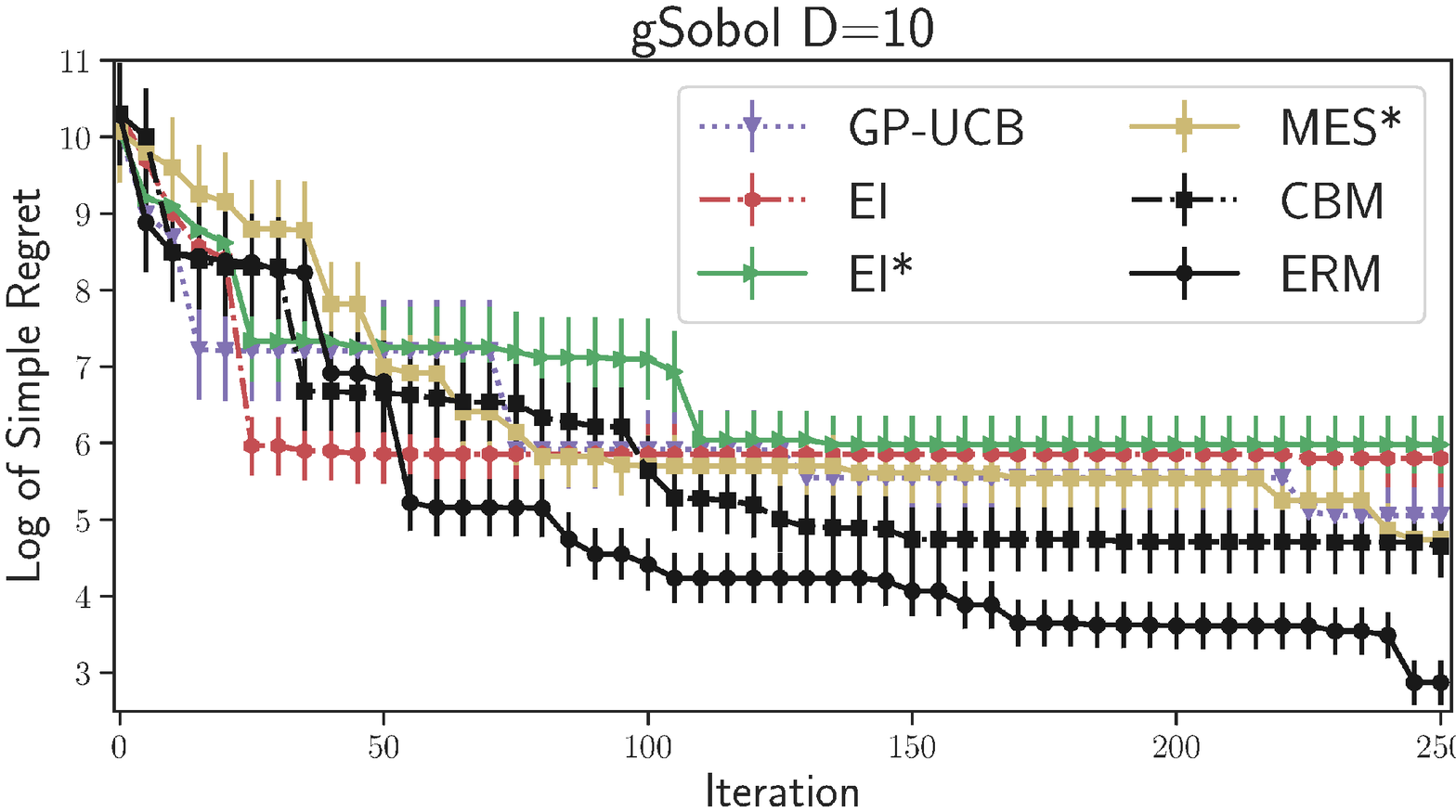}
\par\end{centering}
\vspace{-5pt}
\caption{Optimization comparison using benchmark functions from $D=2$ to $D=10$
dimensions. We demonstrate that the known optimum output $f^{*}$
will significantly boost the performances in high dimensions, such
as in Alpine1 $D=5$, gSobol $D=5$ and $D=10$.\label{fig:Optimization-comparison-benchmark}}

\vspace{-3pt}
\end{figure*}

\subsubsection*{Illustration of CBM and ERM}

We illustrate in Fig. \ref{fig:Illustration-of_ERM} our proposed
CBM and ERM comparing to the standard UCB and EI with both vanilla
GP and transformed GP settings. Our acquisition functions make use
of the knowledge about $f^{*}$ to make an informed decision about
where we should query. That is, CBM and ERM will select the location
where the GP mean $\mu(\bx)$ is close to the optimal value $f^{*}$
and we are highly certain about it -- or low $\sigma(\bx)$. On the
other hand, GP-UCB and EI will always keep exploring as the principle
of explore-exploit without using the knowledge of $f^{*}$. As the
results, GP-UCB and EI can not identify the unknown location $\bx^{*}$
efficiently as opposed to our acquisition functions.

\section{Experiments}

The main goal of our experiments is to show that we can effectively
exploit  the known optimum output to improve Bayesian optimization
performance. We first demonstrate the efficiency of our model
on benchmark functions. Then, we perform hyperparameter optimization
for  a XGBoost classification on Skin Segmentation dataset and a
deep reinforcement learning task on CartPole problem where the optimum
values are publicly available. We provide additional experiments in
the supplement and the code is released at.\footnote{\href{http://github.com/ntienvu/KnownOptimum_BO}{github.com/ntienvu/KnownOptimum\_BO}}

\paragraph{Settings.}

All implementations are in Python. The experiments are independently
performed $20$ times.  We use the squared exponential kernel $k\left(x,x'\right)=\exp\left(-||x-x'||^{2}/\sigma_{l}\right)$
where $\sigma_{l}$ is optimized from the GP marginal likelihood,
the input is scaled $x\sim\left[0,1\right]^{d}$ and the output is
standardized $y\sim\mathcal{N}\left(0,1\right)$ for robustness. We
follow Theorem 3 in \citet{Srinivas_2010Gaussian} to specify $\beta_{t}=2f^{*}+300\log^{3}(t/\delta)$.

To avoid our algorithms from the early exploitation, we use a standard
BO (with GP and EI) at the earlier iterations and our proposed algorithms
at later iterations once the $f^{*}$ value has been reached by the
upper confidence bound. The reaching $f^{*}$ condition can be checked
in each BO iteration using a global optimization toolbox, i.e., $\exists\bx\mid f^{*}\le\mu(\bx)+\sqrt{\beta_{t}}\sigma(\bx)$.

Our CBM and ERM use a transformed Gaussian process (Sec. \ref{subsec:Transformed-Gaussian-Process})
in all experiments. We learn empirically that using a transformed
GP as a surrogate will boost the performance for our CBM and ERM significantly
against the case of using vanilla GP. For other baselines, we use
both surrogates and report the best performance. We present further
details of experiments in the supplement.

\paragraph{Baselines.}

To the best of our knowledge, there is no baseline in directly using
the known optimum output for BO. We select to compare our model with
the vanilla BO without knowing the optimum value including the GP-UCB
\citep{Srinivas_2010Gaussian} and EI \citep{Mockus_1978Application}.
In addition, we use two other baselines using $f^{*}$ described in
Sec. \ref{subsec:Available-acquisition-functions}.

\subsection{Comparison on benchmark function given $f^{*}$}

We perform optimization tasks on $6$ common benchmark functions.\footnote{\href{https://www.sfu.ca/~ssurjano/optimization.html}{https://www.sfu.ca/$\sim$ssurjano/optimization.html}}
For these functions, we assume that the optimum value $f^{*}$ is
available in advance which will be given to the algorithm. We use
the simple regret  for comparison, defined as $f^{*}-\max_{\forall i\le t}f(x_{i})$
for maximization problem.

The experimental results are presented in Fig. \ref{fig:Optimization-comparison-benchmark}
which shows that our proposed CBM and ERM are among the best approaches
over all problems considered.  This is because our framework has
utilized the additional knowledge of $f^{*}$ to build an informed
surrogate model and decision functions. Especially, ERM outperforms
all methods by a wide margin. While CBM can be sensitive to the hyperparameter
$\beta_{t}$, ERM has no parameter and is thus more robust. 
\begin{table}
\begin{centering}
\caption{Hyperparameters for XGBoost.\label{tab:Hyper-parameters-for-XGBoost.}}
\par\end{centering}
\begin{centering}
\begin{tabular}{cccc}
\toprule 
\multicolumn{4}{c}{Known $f^{*}=100$ (Accuracy)}\tabularnewline
Variables & Min & Max & Found $\bx^{*}$\tabularnewline
\midrule
min child weight & $1$ & $20$ & $4.66$\tabularnewline
colsample bytree & $0.1$ & $1$ & $0.99$\tabularnewline
max depth & $5$ & $15$ & $9.71$\tabularnewline
subsample & $0.5$ & $1$ & $0.77$\tabularnewline
alpha & $0$ & $10$ & $0.82$\tabularnewline
gamma & $0$ & $10$ & $0.51$\tabularnewline
\bottomrule
\end{tabular}
\par\end{centering}
\vspace{0pt}
\end{table}

Particularly, our approaches with $f^{*}$ perform significantly better
than the baselines in gSobol and Alpine1 functions. The results
indicate that the knowledge of $f^{*}$ is particularly useful for
high dimensional functions.

\subsection{Tuning machine learning algorithms with $f^{*}$}

A popular application of BO is for hyperparameter tuning of machine
learning models. Some machine learning tasks come with the known optimal
value in advance. We consider tuning (1) a classification task using
XGBoost on a Skin dataset  and (2) a deep reinforcement learning
task on a CartPole problem \citep{barto1983neuronlike}. Further detail
of the experiment is described in the supplement.

\paragraph{XGBoost classification.}

We demonstrate a classification task using XGBoost \citep{chen2016xgboost}
on a Skin Segmentation dataset\footnote{\href{https://archive.ics.uci.edu/ml/datasets/skin\%2Bsegmentation}{https://archive.ics.uci.edu/ml/datasets/skin+segmentation}}
where we know the best accuracy is $f^{*}=100$, as shown in Table
1 of \citet{Le_2016Nonparametric}.
\begin{figure}
\begin{centering}
\includegraphics[width=1\columnwidth]{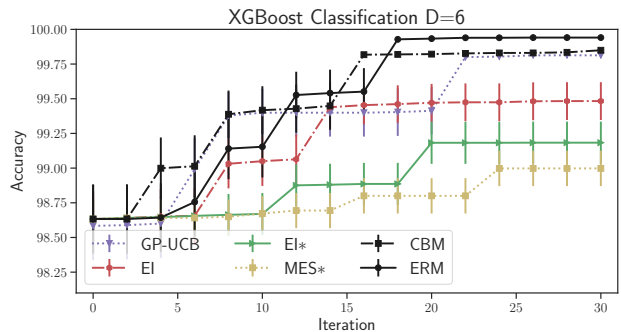}
\par\end{centering}
\begin{centering}
\vspace{-4pt}
\caption{Tuning performance on Skin dataset.\label{fig:Tuning-performance-Skin}}
\par\end{centering}
\vspace{-6pt}
\end{figure}

The Skin Segmentation dataset is split into $15\%$ for training and
$85\%$ for testing for a classification problem. There are $6$ hyperparameters
for XGBoost \citep{chen2016xgboost} which is summarized in Table
\ref{tab:Hyper-parameters-for-XGBoost.}. To optimize the integer
(ordinal) variables, we round the scalars to the nearest values in
the continuous space. We present the result in Fig. \ref{fig:Tuning-performance-Skin}.
Our proposed ERM is the best approach, outperforming all the baselines
by a wide margin. This demonstrates the benefit of exploiting the
 optimum value $f^{*}$ in BO.

\paragraph{Deep reinforcement learning.}

\begin{figure*}
\begin{centering}
\includegraphics[width=1\columnwidth]{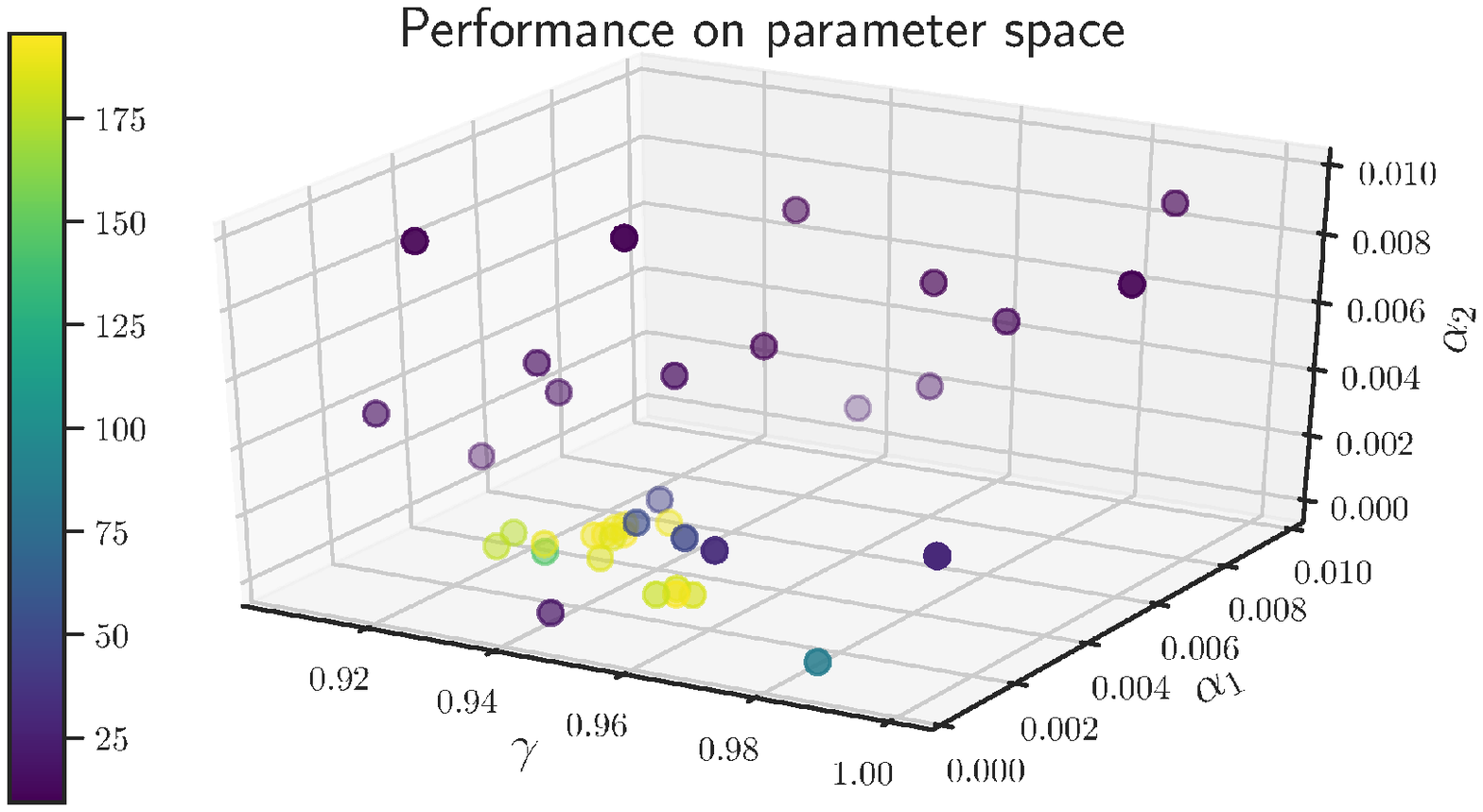}\includegraphics[width=1\columnwidth]{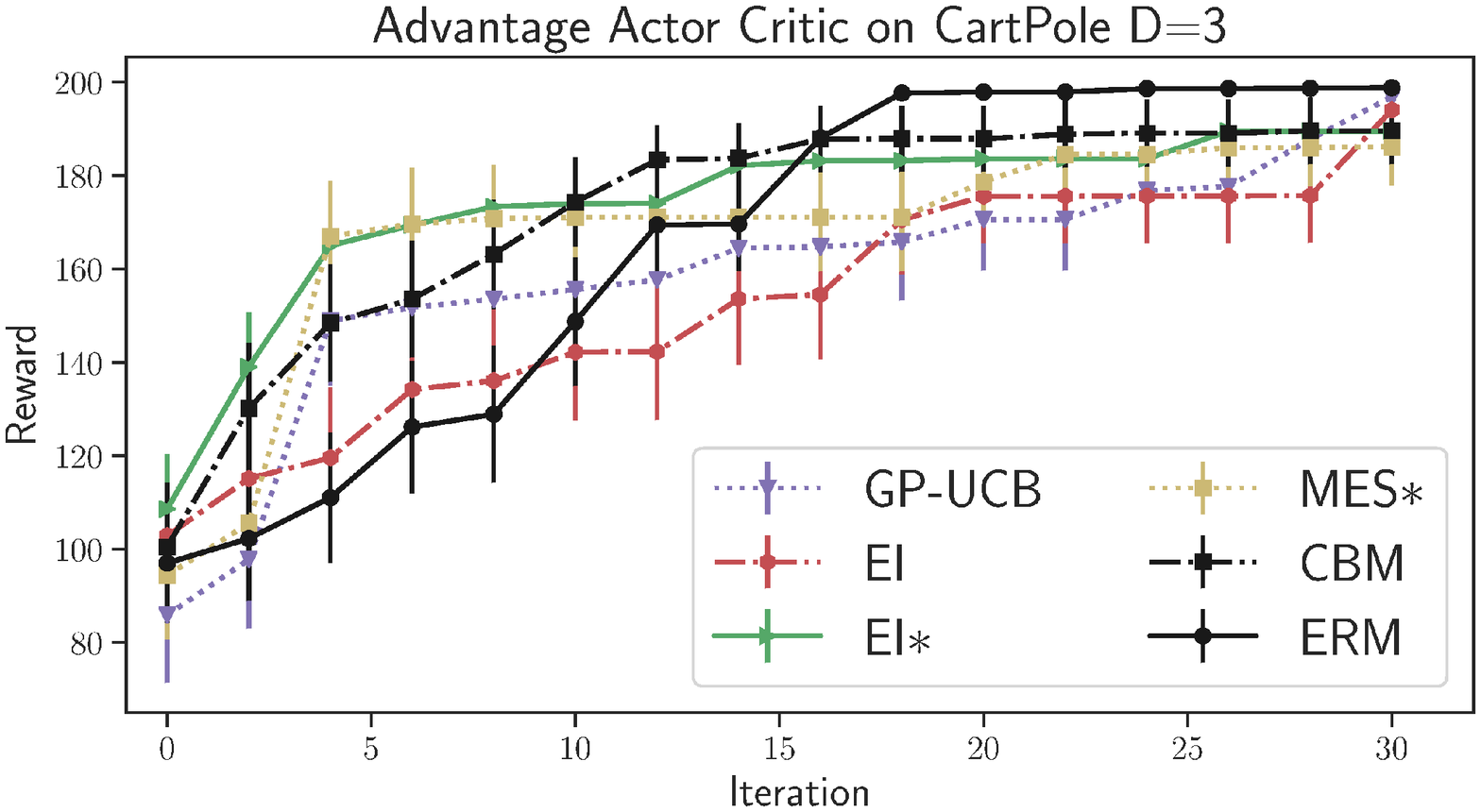}
\par\end{centering}
\vspace{-3pt}

\caption{Hyperparameter tuning for a deep reinforcement learning algorithm.
The optimum value is available $f^{*}=200$. Left: Selected points
by our algorithm on tuning DRL. Color indicates the reward $f(x)$
value. Right: Performance comparison with the baselines. \label{fig:Hyper-parameter-tuning_drl}}

\vspace{-5pt}
\end{figure*}
CartPole is a pendulum with a center of gravity above its pivot point.
The goal is to keep the cartpole balanced by controlling a pivot
point. The reward performance in CartPole is often averaged over
$100$ consecutive trials. The maximum reward is known from the literature\footnote{\href{https://gym.openai.com/envs/CartPole-v0/}{https://gym.openai.com/envs/CartPole-v0/}}
as $f^{*}=200$. 

We then use a deep reinforcement learning (DRL) algorithm to solve
the CartPole problem and use Bayesian optimization to optimize the
 hyperparameters. In particular, we use the advantage actor critic
(A2C) \citep{Sutton_1998Reinforcement} which possesses three sensitive
hyperparameters, including the discount factor $\gamma$, the learning
rate for actor model, $\alpha_{1}$, and the learning rate for critic
model, $\alpha_{2}$. We choose not to optimize the deep learning
architecture for simplicity. We use Bayesian optimization given the
known optimum output of $200$ to find the best hyperparameters for
the A2C algorithm. We present the results in Fig. \ref{fig:Hyper-parameter-tuning_drl}
where our ERM reaches the optimal performance after $20$ iterations
outperforming all other baselines. In Fig. \ref{fig:Hyper-parameter-tuning_drl}
Left, we visualize the selected point $\{\bx_{t}\}_{t=1}^{T}$ by
our ERM acquisition function. Our ERM initially explores at several
places and then exploits in the high value region (yellow dots).

\subsection{What happens if we misspecify the optimum value}

We now consider setting the $f^{*}$ to a value which is not the
true optimum of the black-box function. We show that our model's performance
will drop with misspecified value of $f^{*}$ with different effects.
Specifically, we both set $f^{*}$ larger (over-specify) and smaller
(under-specify) than the true value in a maximization problem.

\begin{figure}
\begin{centering}
\includegraphics[width=0.95\columnwidth]{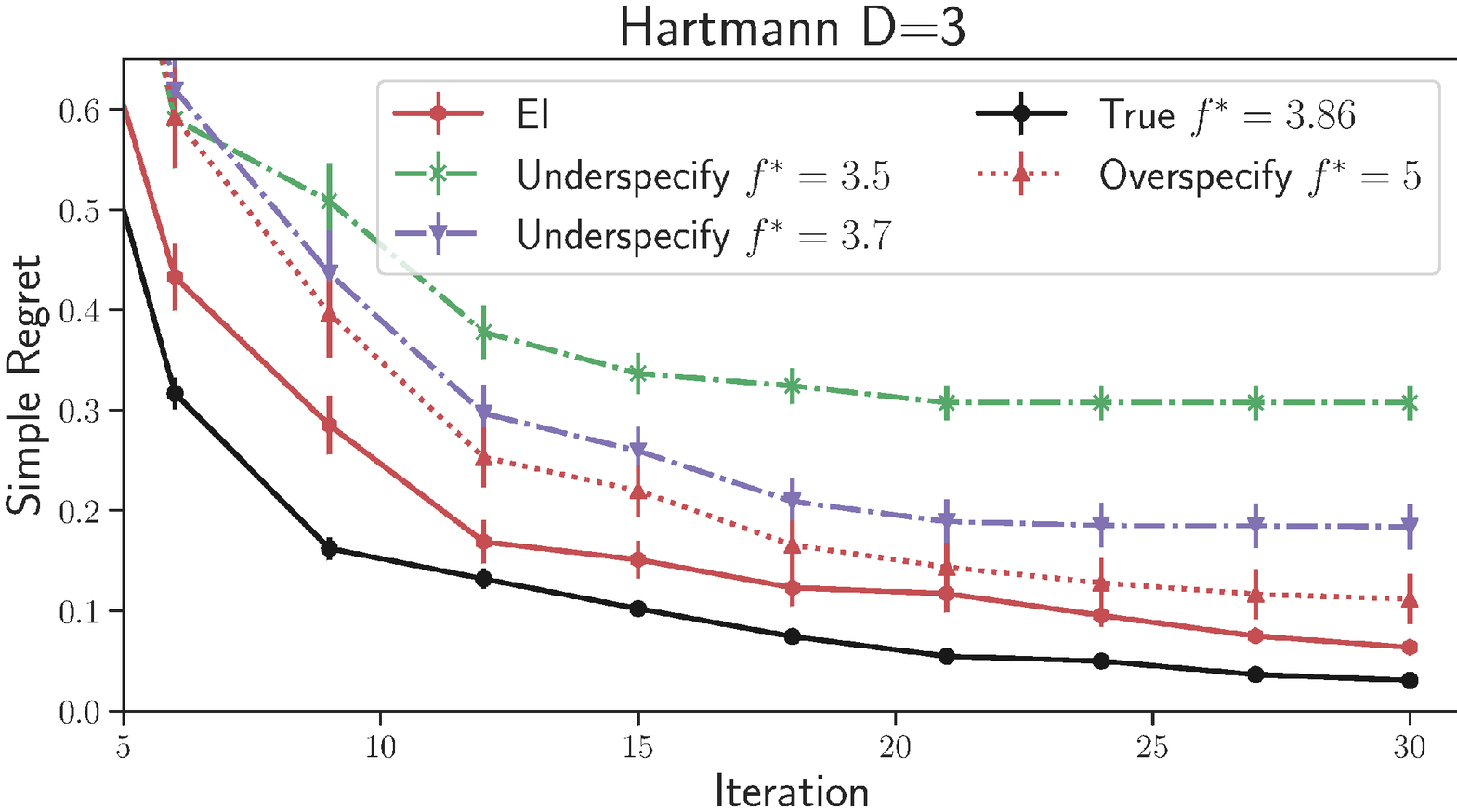}
\par\end{centering}
\begin{centering}
\includegraphics[width=0.95\columnwidth]{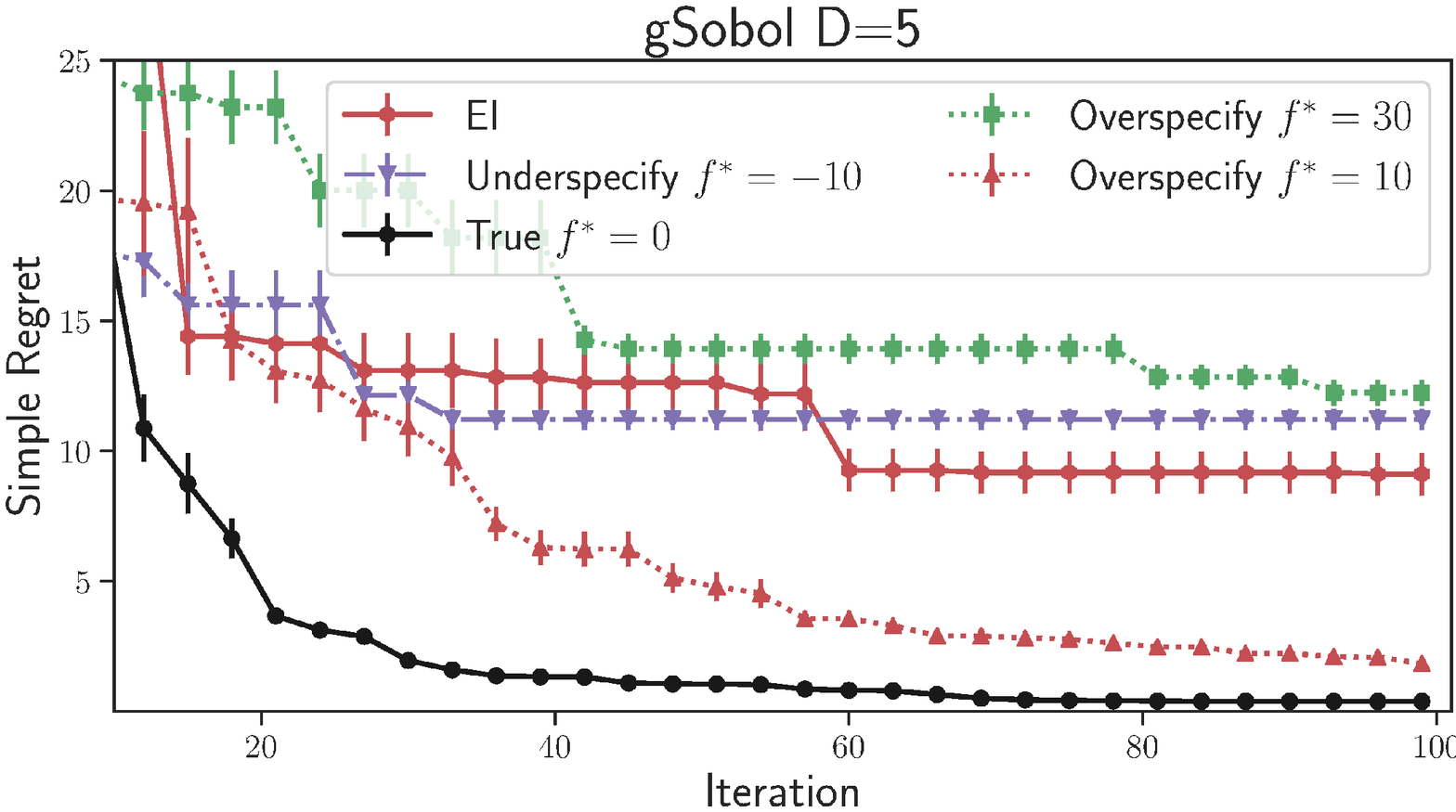}
\par\end{centering}
\caption{Experiments with ERM in the maximization problem. Over-specifying
is when the value of $f^{*}$ is larger than the true optimal value
and under-specifying is when the value of $f^{*}$ is smaller than
the true. Top: the true $f^{*}=3.86$ for Hartmann. Bottom: the
true $f^{*}=0$ for gSobol. Both cases of misspecifying $f^{*}$ will
degrade the performance. \label{fig:Experiments_misspecified_fstar}}

\vspace{-10pt}
\end{figure}

We experiment with our ERM using this misspecified setting of $f^{*}$
in Fig. \ref{fig:Experiments_misspecified_fstar}. 
 The results suggest that our algorithm using the true value ($f^{*}=3.86$
for Hartmann and $f^{*}=0$ for gSobol) will have the best performance.
Both over-specifying and under-specifying  $f^{*}$ will return worse
performance. These misspecified settings slightly perform worse than
the standard EI in Hartmann while it still performs better than EI
for overspecifying $f^{*}=10$ in gSobol. In particular, the under-specifying
case will result in worse performance than over-specifying. This is
because our acquisition function will get stuck at the area once being
found wrongly as the optimal. On the other hand, if we over-specify
$f^{*}$, our algorithm continues exploring to find the optimum because
it can not find the point where both conditions are met $\sigma(\bx_{t})=0$
and $f^{*}=\mu(\bx_{t})$. 

\paragraph{Discussion.}

We make the following observations. If we know the true value $f^{*}$,
ERM  will return the best result. If we do not know the exact $f^{*}$
value, the performance of our approach is degraded. Thus, we should
use the existing BO approaches, such as EI, for the best performance.

\section{Conclusion and Future Work}

In this paper, we have considered a new setting in Bayesian optimization
with known optimum output. We present a transformed Gaussian process
surrogate to model the objective function better by exploiting the
knowledge of $f^{*}$. Then, we propose two decision strategies which
can exploit the function optimum value to make informed decisions.
Our approaches are intuitively simple and easy to implement. By using
extra knowledge of $f^{*}$, we demonstrate that our ERM can converge
quickly to the optimum in benchmark functions and real-world applications.

In future work, we can expand our algorithm to handle batch setting
for parallel evaluations or extend this work to other classes of surrogate
functions such as Bayesian neural networks \citep{neal2012bayesian}
and deep GP \citep{damianou2013deep}. Moreover, we can extend the
model to handle $f^{*}$ within a range of $\epsilon$ from the true
output.

\bibliographystyle{icml2020}
\bibliography{vn}

\appendix

\section{Expected Regret Minimization Derivation}

We are given an optimization problem $\bx^{*}=\arg\max_{\bx\in\mathcal{X}}f(\bx)$
where $f$ is a black-box function that we can evaluate pointwise.
Let $\mathcal{D}_{t}=\left\{ \bx_{i}\in\mathcal{X},y_{i}\in\mathcal{R}\right\} _{i=1}^{t}$
be the observation set including an input $\bx_{i}$, an outcome $y_{i}$
and $\mathcal{X}\in\mathcal{R}^{d}$ be the bounded search space.
We define the regret function $r\left(\bx\right)=f^{*}-f(\bx)$ where
$f^{*}=\max_{\bx\in\mathcal{X}}f(\bx)$ is the known global optimum
value. The likelihood of the regret $r\left(\bx\right)$ on a normal
posterior distribution is as follows
\begin{align}
p\left(r(\bx)\right)= & \frac{1}{\sqrt{2\pi}\sigma\left(\bx\right)}\exp\left(-\frac{1}{2}\frac{\left[\mu\left(\bx\right)-f^{*}+r(\bx)\right]^{2}}{\sigma^{2}\left(\bx\right)}\right).\label{eq:ELI_llk_improvement}
\end{align}
The expected regret can be written using the likelihood function in
Eq. (\ref{eq:ELI_llk_improvement}), we obtain $\mathbb{E}\left[r\left(\bx\right)\right]$
\begin{align*}
 & =\int_{0}^{\infty}\frac{r(\bx)}{\sqrt{2\pi}\sigma\left(\bx\right)}\exp\left(-\frac{1}{2}\frac{\left[\mu\left(\bx\right)-f^{*}+r(\bx)\right]^{2}}{\sigma^{2}\left(\bx\right)}\right)dr(\bx).
\end{align*}
As the ultimate goal in optimization is to minimize the regret, we
consider our acquisition function to minimize this expected regret
as $\alpha^{\textrm{ERM}}\left(\bx\right)=\mathbb{E}\left[r\left(\bx\right)\right]$.
Let $t=\frac{\mu\left(\bx\right)-f^{*}+r(\bx)}{\sigma\left(\bx\right)}$,
then $r(\bx)=t\times\sigma\left(\bx\right)-\mu(\bx)+f^{*}$ and $dt=\frac{dr}{\sigma\left(\bx\right)}$.
We write $\alpha^{\textrm{ERM}}\left(\bx\right)$ 
\begin{align}
= & \int_{t=\frac{\mu\left(\bx\right)-f^{*}}{\sigma\left(\bx\right)}}^{\infty}\frac{t\times\sigma\left(\bx\right)+f^{*}-\mu\left(\bx\right)}{\sqrt{2\pi}}\exp(-\frac{1}{2}t^{2})dt\nonumber \\
= & \sigma\left(\bx\right)\int_{t=\frac{\mu\left(\bx\right)-f^{*}}{\sigma\left(\bx\right)}}^{\infty}\frac{t}{\sqrt{2\pi}}\exp(-\frac{1}{2}t^{2})dt\nonumber \\
 & +\left[f^{*}-\mu\left(\bx\right)\right]\int_{t=\frac{\mu\left(\bx\right)-f^{*}}{\sigma\left(\bx\right)}}^{\infty}\frac{1}{\sqrt{2\pi}}\exp(-\frac{1}{2}t^{2})dt.\label{eq:alpha_EI_derivation}
\end{align}
We compute the first term in Eq. (\ref{eq:alpha_EI_derivation}) as
\begin{align*}
 & \sigma\left(\bx\right)\int_{t=\frac{\mu\left(\bx\right)-f^{*}}{\sigma\left(\bx\right)}}^{\infty}\frac{t}{\sqrt{2\pi}}\exp(-\frac{1}{2}t^{2})dt\\
 & =\frac{\sigma\left(\bx\right)}{\sqrt{2\pi}}\left[-\exp\left(-\frac{1}{2}\left[\frac{\mu\left(\bx\right)-f^{*}+r(\bx)}{\sigma\left(\bx\right)}\right]^{2}\right)\right]_{r=0}^{r=\infty}\\
 & =\sigma(\bx)\mathcal{N}\left(\frac{\mu\left(\bx\right)-f^{*}}{\sigma\left(\bx\right)}\mid0,1\right).
\end{align*}
Next, we compute the second term in Eq. (\ref{eq:alpha_EI_derivation})
as 
\begin{align*}
 & \left[f^{*}-\mu\left(\bx\right)\right]\int_{t=\frac{\mu\left(\bx\right)-f^{*}}{\sigma\left(\bx\right)}}^{\infty}\frac{1}{\sqrt{2\pi}}\exp(-\frac{1}{2}t^{2})dt\\
 & =\left[f^{*}-\mu\left(\bx\right)\right]\left\{ \int_{-\infty}^{\infty}\mathcal{N}\left(t\mid0,1\right)dt-\int_{-\infty}^{\frac{\mu\left(\bx\right)-f^{*}}{\sigma\left(\bx\right)}}\mathcal{N}\left(t\mid0,1\right)dt\right\} \\
 & =\left[f^{*}-\mu\left(\bx\right)\right]\left[1-\Phi\left(\frac{\mu\left(\bx\right)-f^{*}}{\sigma\left(\bx\right)}\right)\right]\\
 & =\left[f^{*}-\mu\left(\bx\right)\right]\Phi\left(\frac{f^{*}-\mu\left(\bx\right)}{\sigma\left(\bx\right)}\right).
\end{align*}
Let $z=\frac{f^{*}-\mu\left(\bx\right)}{\sigma\left(\bx\right)}$,
we obtain the acquisition function
\begin{align}
\alpha^{\textrm{ERM}}\left(\bx\right) & =\sigma\left(\bx\right)\phi\left(z\right)+\left[f^{*}-\mu\left(\bx\right)\right]\Phi\left(z\right)\label{eq:AcqFunc_ELI}
\end{align}
where $\phi\left(z\right)=\mathcal{N}\left(z\mid0,1\right)$ is the
standard normal pdf and $\Phi\left(z\right)$ is the cdf. To select
the next point, we minimize this acquisition function which is equivalent
to minimize the expected regret $\mathbb{E}\left[r\left(\bx\right)\right]$
\begin{align*}
\bx_{t+1} & =\arg\min_{\bx\in\mathcal{X}}\alpha^{\textrm{ERM}}\left(\bx\right)=\arg\min_{\bx\in\mathcal{X}}\mathbb{E}\left[r\left(\bx\right)\right].
\end{align*}
We can see that this acquisition function is minimized $\alpha^{\textrm{ERM}}\left(\bx_{t}\right)=\mathbb{E}[r(\bx_{t})]=0$
when $f^{*}=\mu(\bx_{t})$ and $\sigma(\bx_{t})=0$. Our chosen point
$\bx_{t}$ is the one which offers the smallest expected regret. We
aim to find the point with the desired property of $\mathbb{E}[r(\bx_{t})]=0$.

\section{Additional Experiments}

We first illustrate the BO with and without the knowledge of $f^{*}$.
Then, we provide  additional information about the deep reinforcement
learning experiment in the main paper. Next, we compare the effect
of using the vanilla GP and transformed GP with different acquisition
functions.

\subsection{Illustration per iteration}

\begin{figure*}
\includegraphics[width=0.5\textwidth]{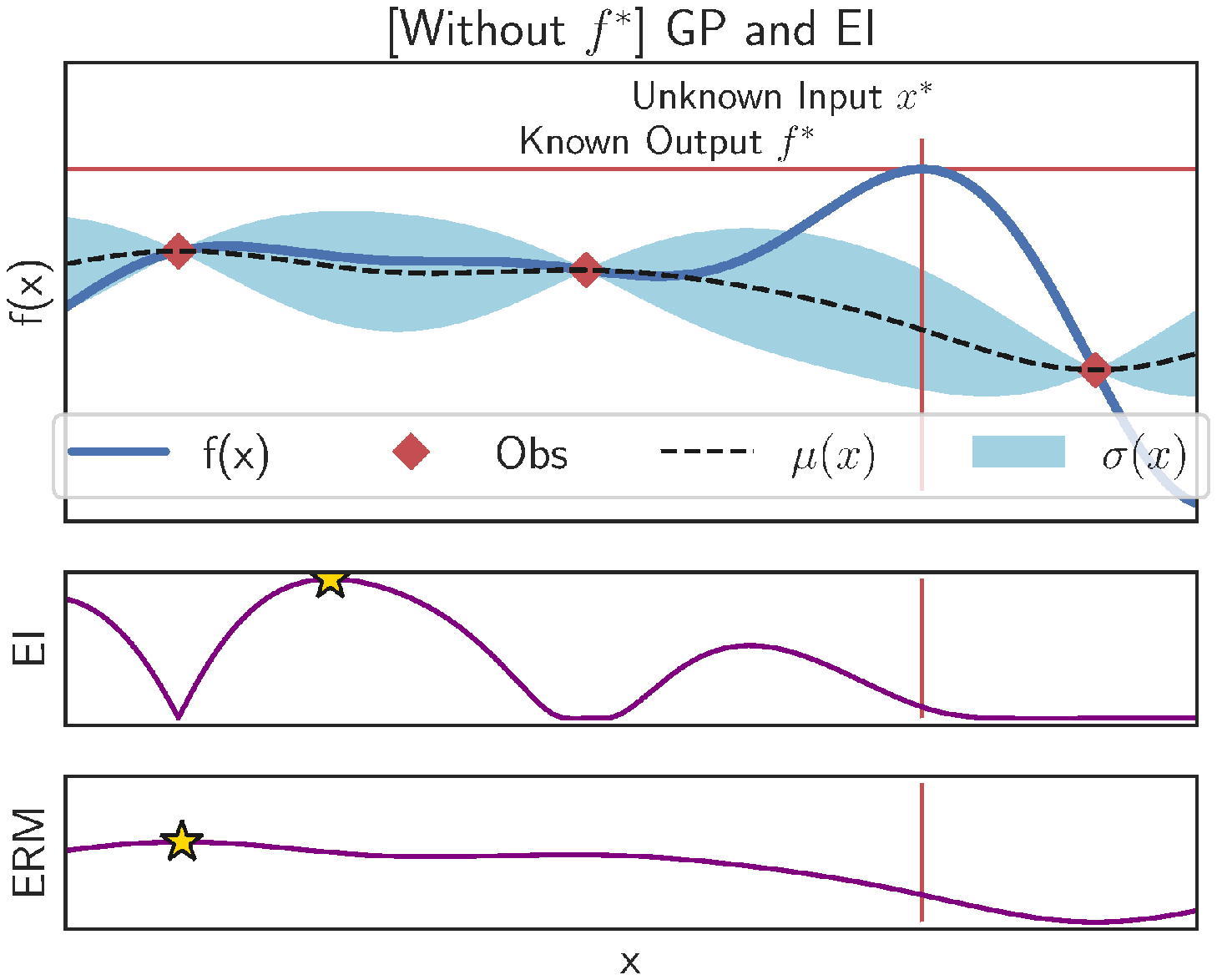}\includegraphics[width=0.5\textwidth]{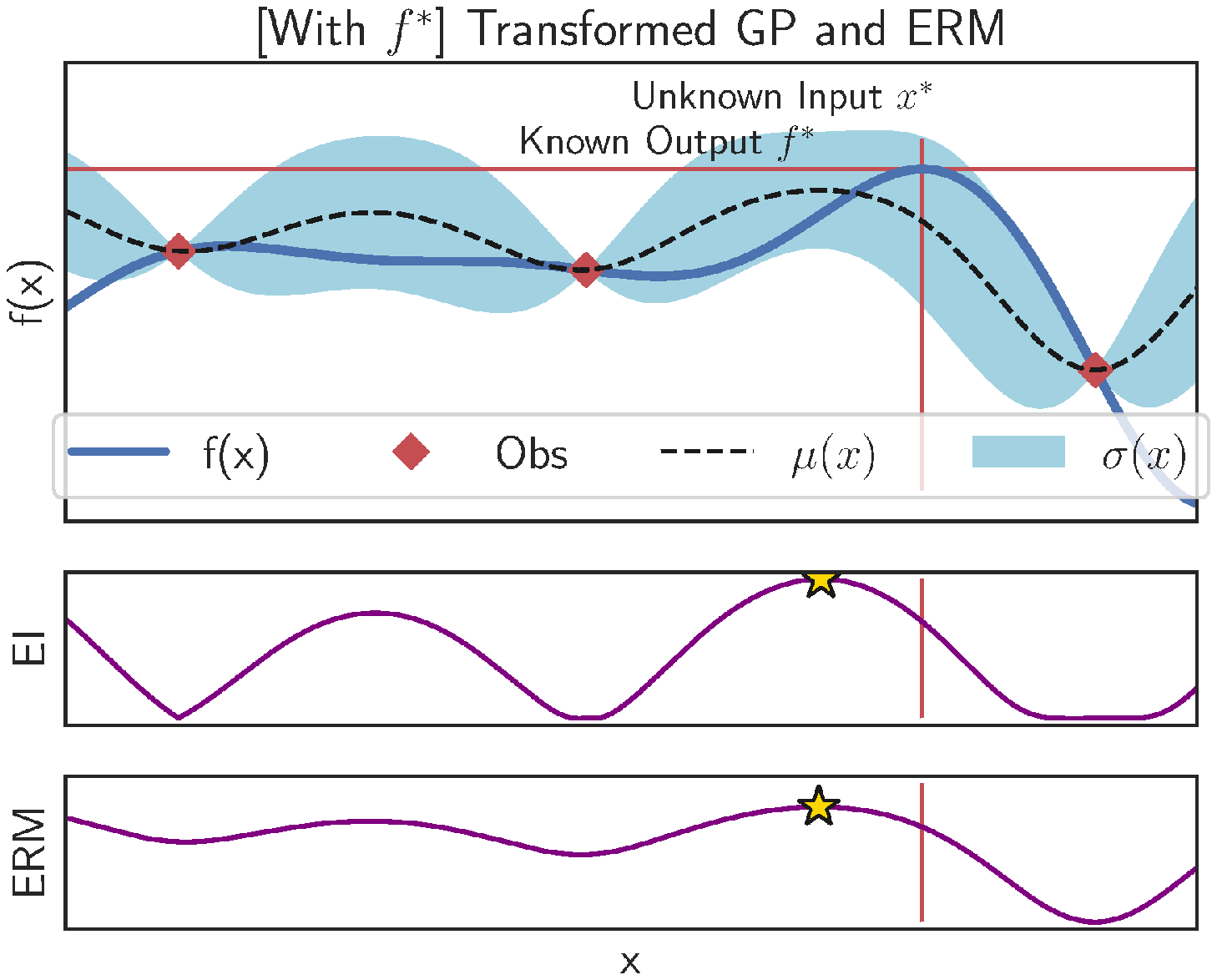}

\includegraphics[width=0.5\textwidth]{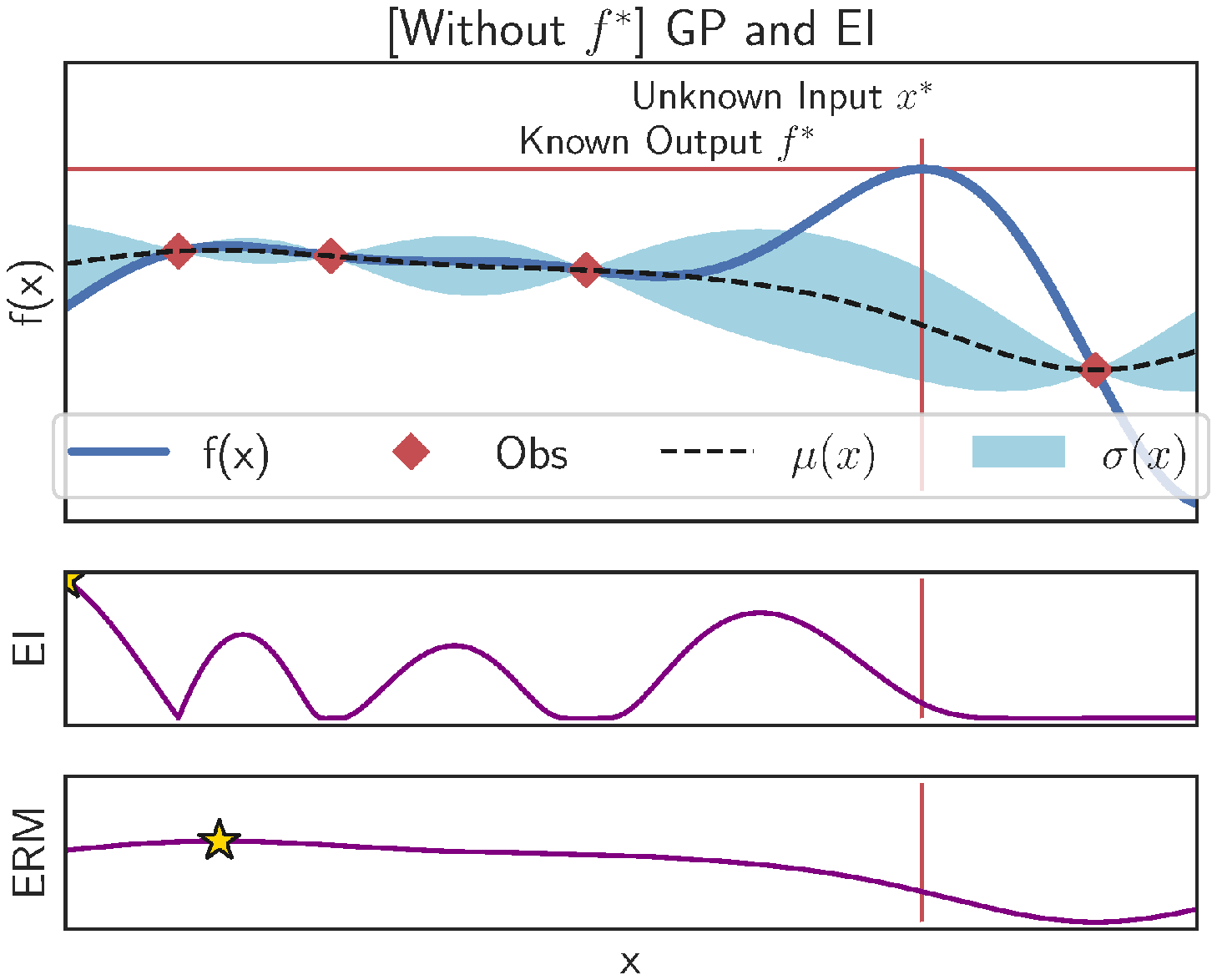}\includegraphics[width=0.5\textwidth]{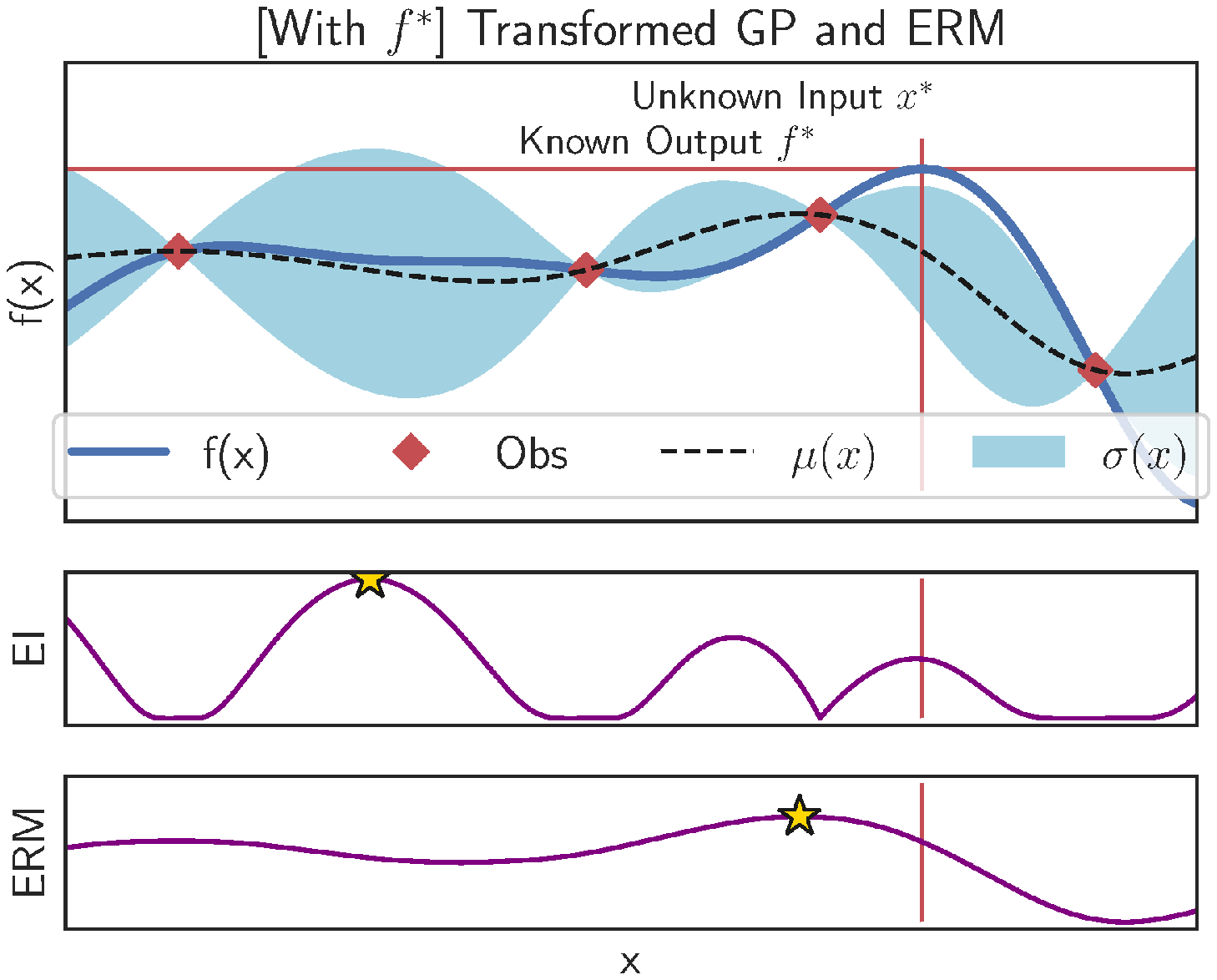}

\includegraphics[width=0.5\textwidth]{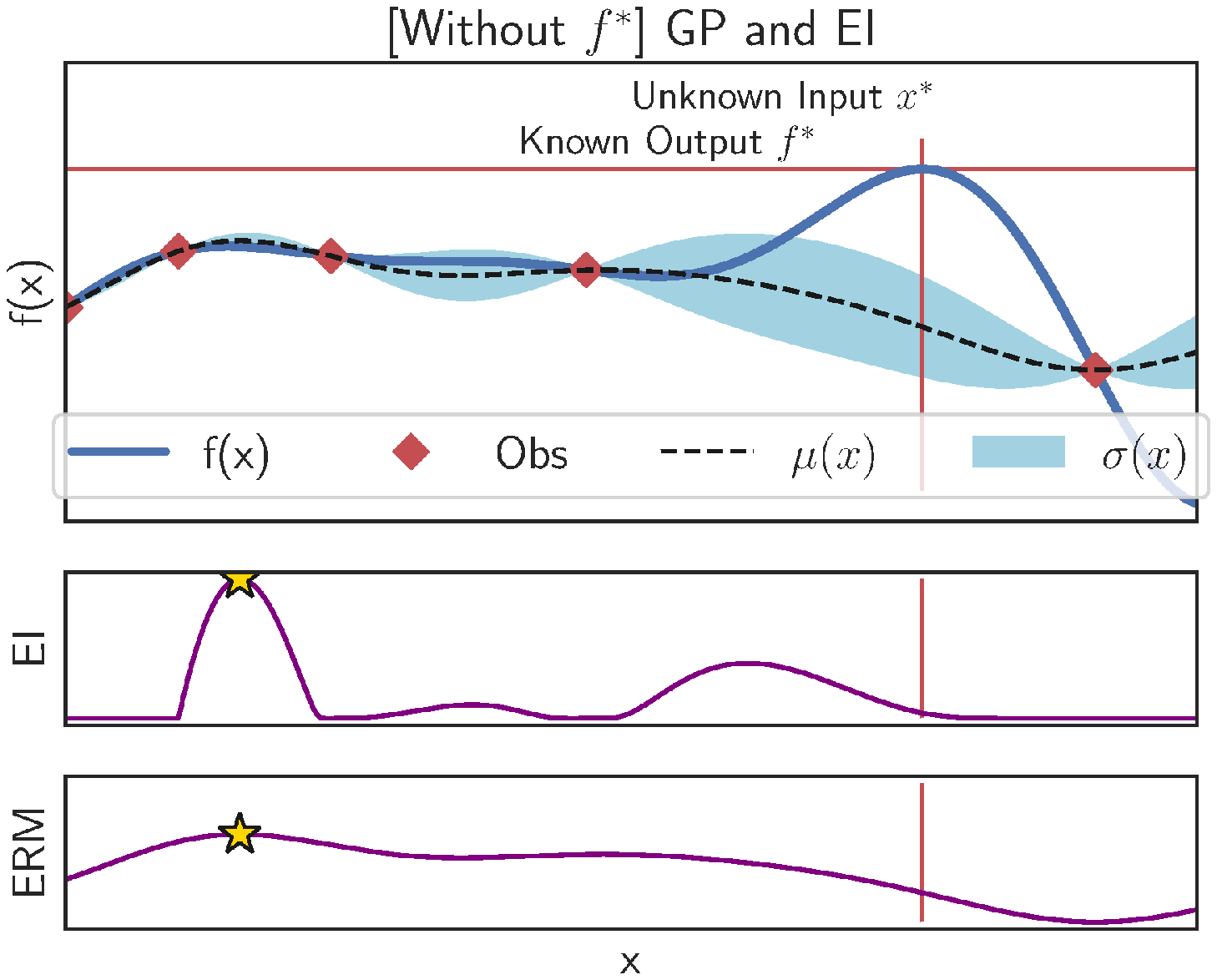}\includegraphics[width=0.5\textwidth]{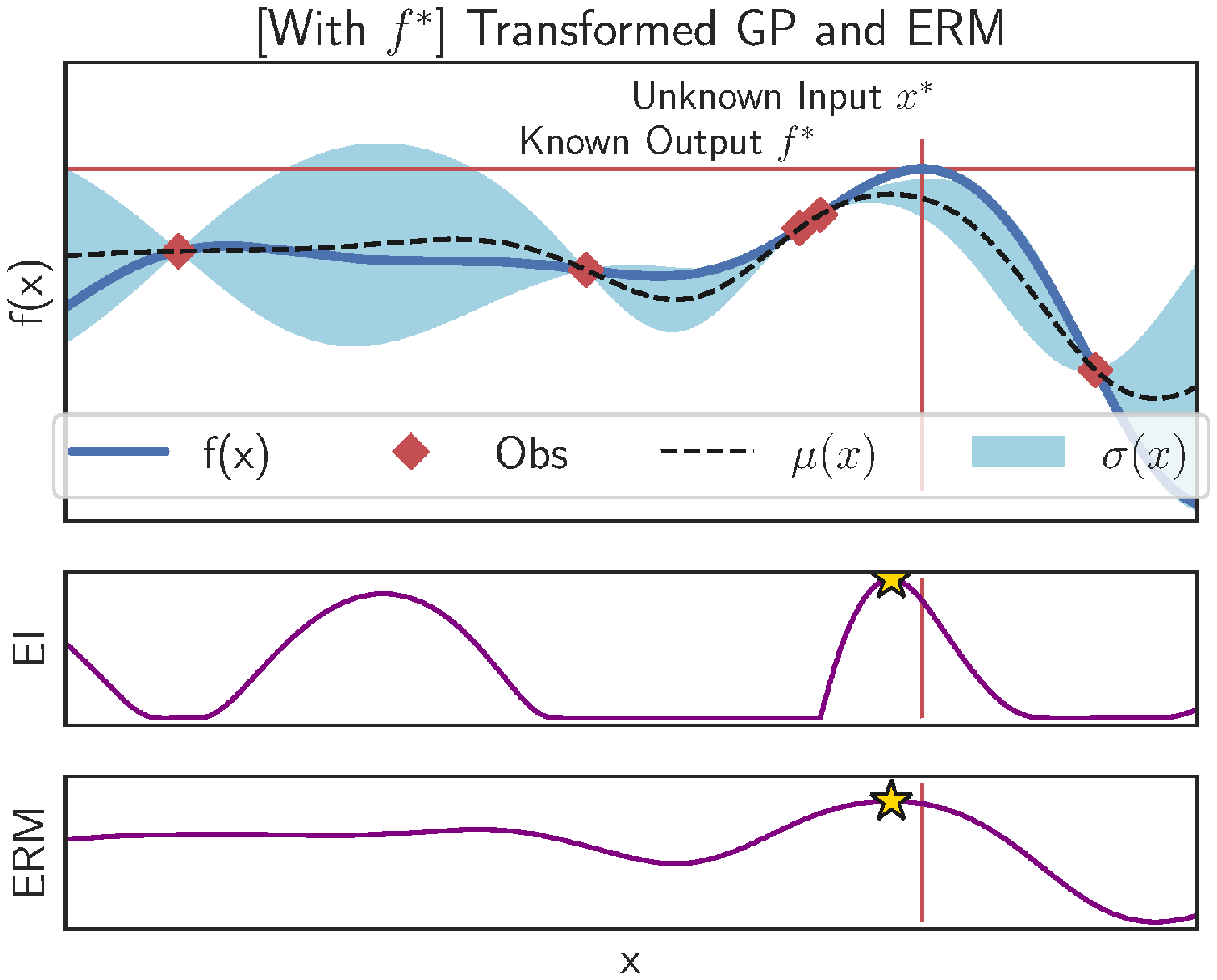}

\caption{Illustration of the optimization process per iteration $1-3$ starting
given the same initialization. Left: BO using GP as surrogate and
EI as acquisition function. Right: BO using TGP as surrogate and ERM
as acquisition function. Given the known optimum $f^{*}$ value, the
transformed GP can lift up the surrogate model closer to the known
value. Then, the ERM will make informed decision given $f^{*}$. We
also show that the EI may not make the best decision as ERM. To be
continue in the next figure. \label{fig:Illustration1}}

\end{figure*}

\begin{figure*}
\includegraphics[width=0.5\textwidth]{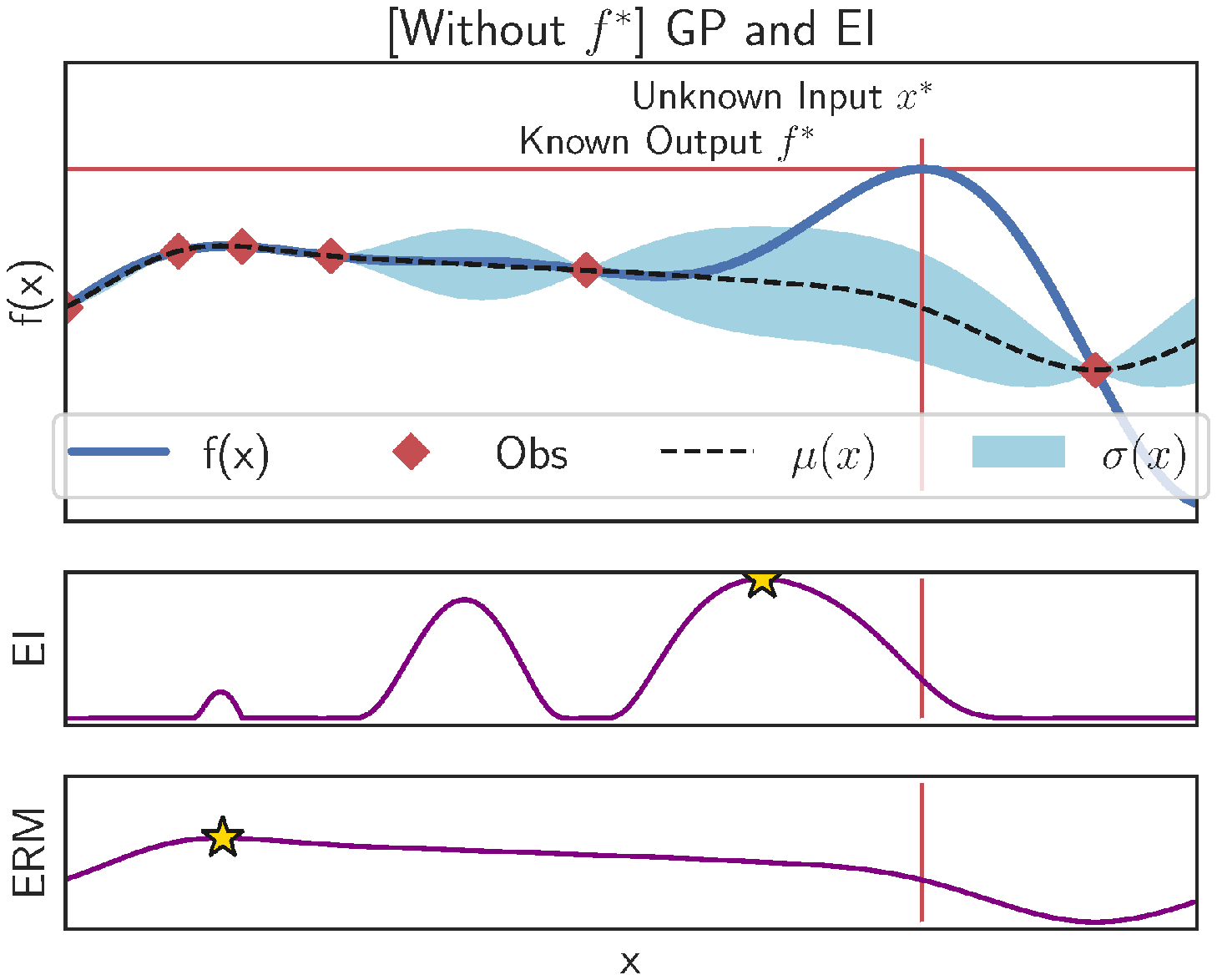}\includegraphics[width=0.5\textwidth]{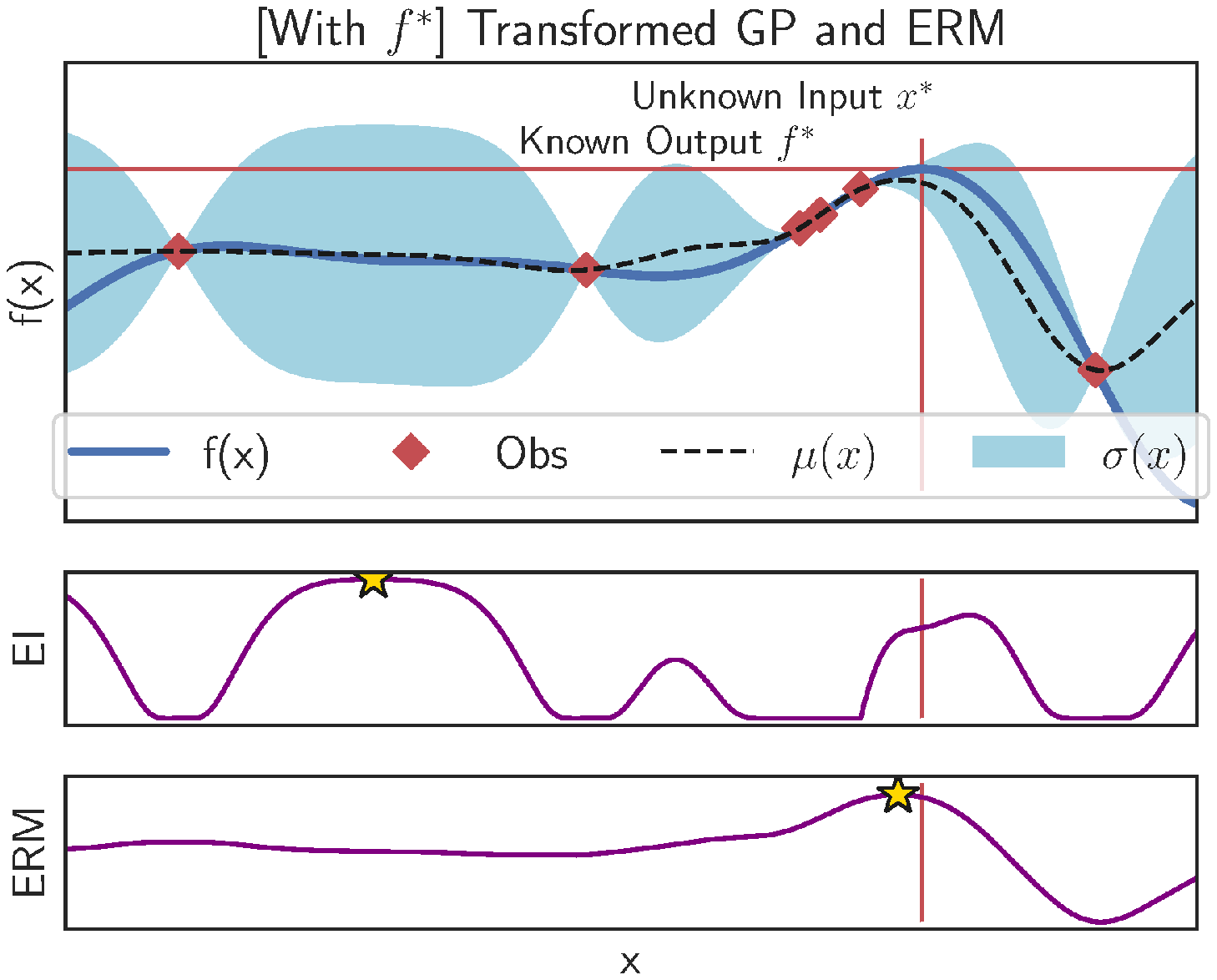}

\includegraphics[width=0.5\textwidth]{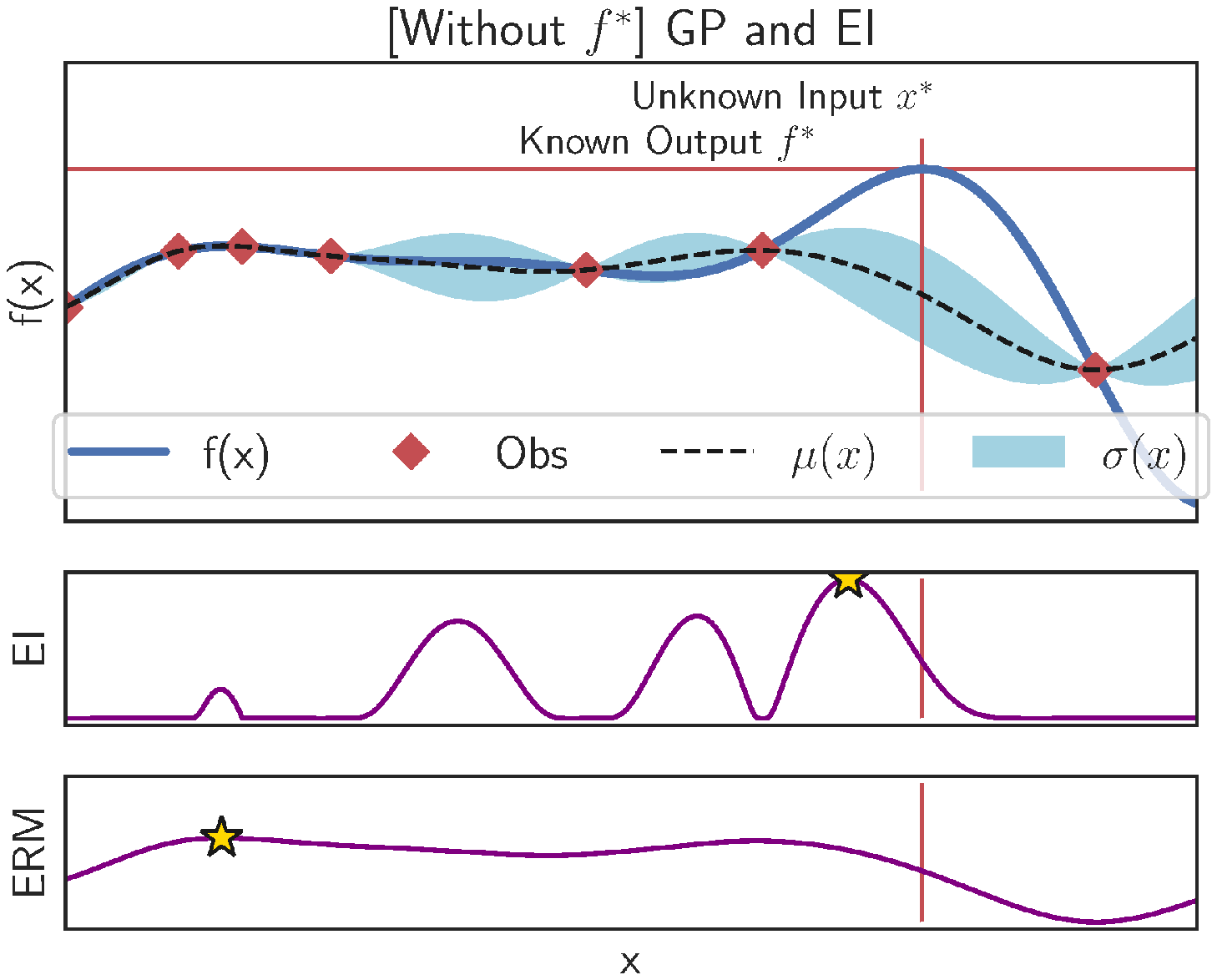}\includegraphics[width=0.5\textwidth]{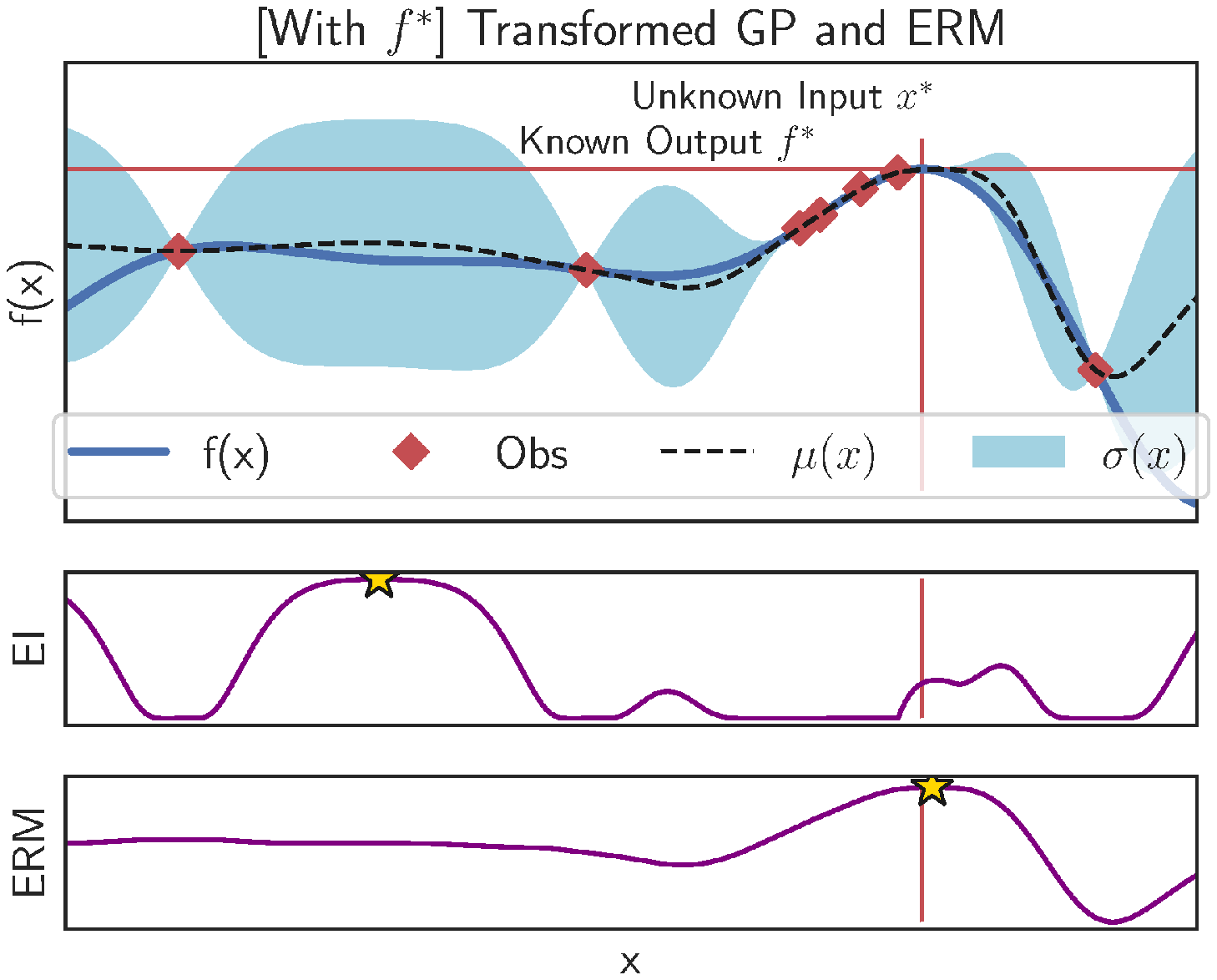}

\includegraphics[width=0.5\textwidth]{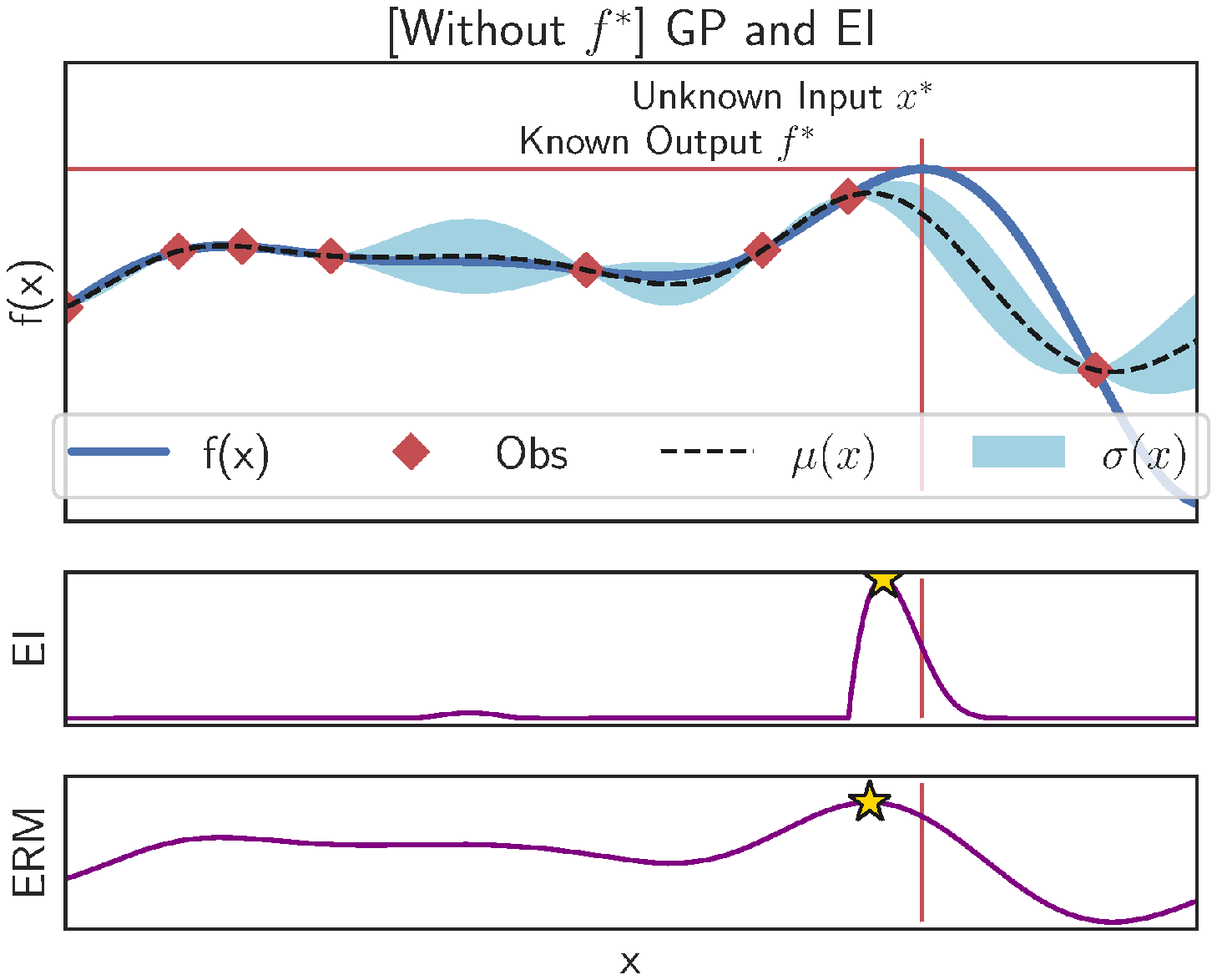}\includegraphics[width=0.5\textwidth]{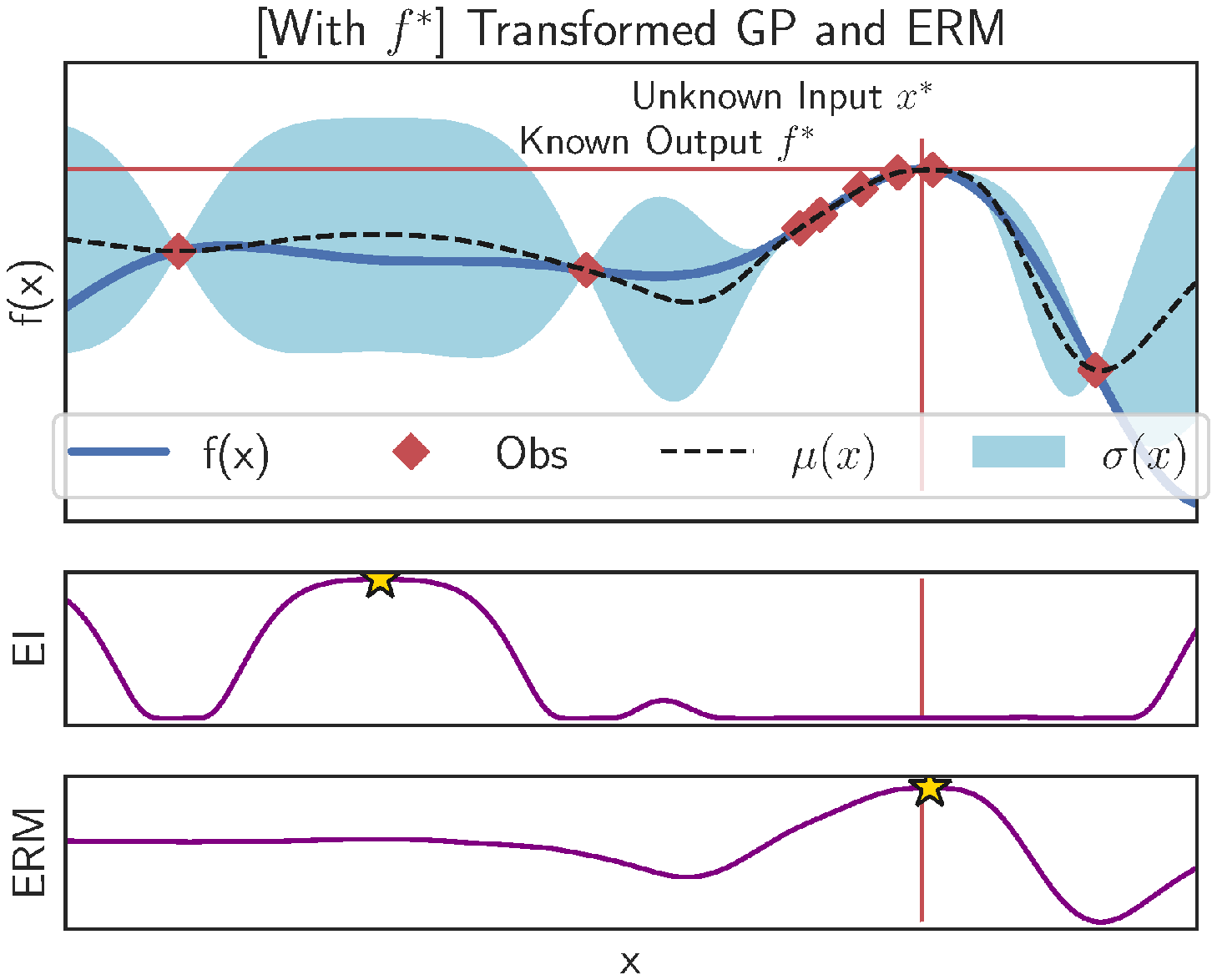}

\caption{Continuing from the previous figure. Illustration of the optimization
process per iteration $4-6$ starting given the same initialization.
Left: BO using GP as surrogate and EI as acquisition function. Right:
BO using TGP as surrogate and ERM as acquisition function. Given the
known optimum $f^{*}$ value, the transformed GP can lift up the surrogate
model closer to the known value. Then, the ERM will make informed
decision given $f^{*}$. We also show that the EI may not make the
best decision as ERM.\label{fig:Illustration2}}
\end{figure*}

We provide the illustration of BO with and without the knowledge of
$f^{*}$ for comparison in Figs. \ref{fig:Illustration1} and \ref{fig:Illustration2}.
We show the GP and EI in the left (without $f^{*}$) and the transformed
GP and ERM in the right (with $f^{*}$). As the effect of transformation
using $f^{*}$, the transformed GP (right) can lift up the surrogate
model closer to the true value $f^{*}$ (\textcolor{red}{red} horizontal
line) encouraging the acquisition function to select at these potential
locations. On the other hand, without $f^{*}$, the GP surrogate (left)
is less informative. As a result, the EI operating on GP (left) is
less efficient as opposed to the transformed GP. We demonstrate visually
that using TGP our model can finally find the optimum input within
the evaluation budget while the standard GP does not.

\subsection{Details of A2C on CartPole problem}

\begin{figure*}
\begin{centering}
\includegraphics[width=0.13\textwidth]{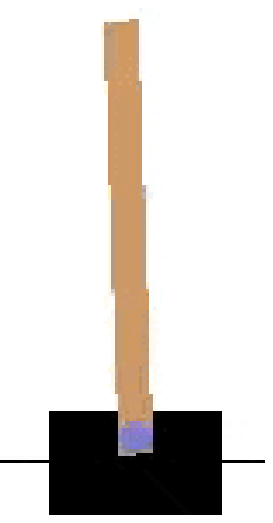}\includegraphics[width=0.43\textwidth]{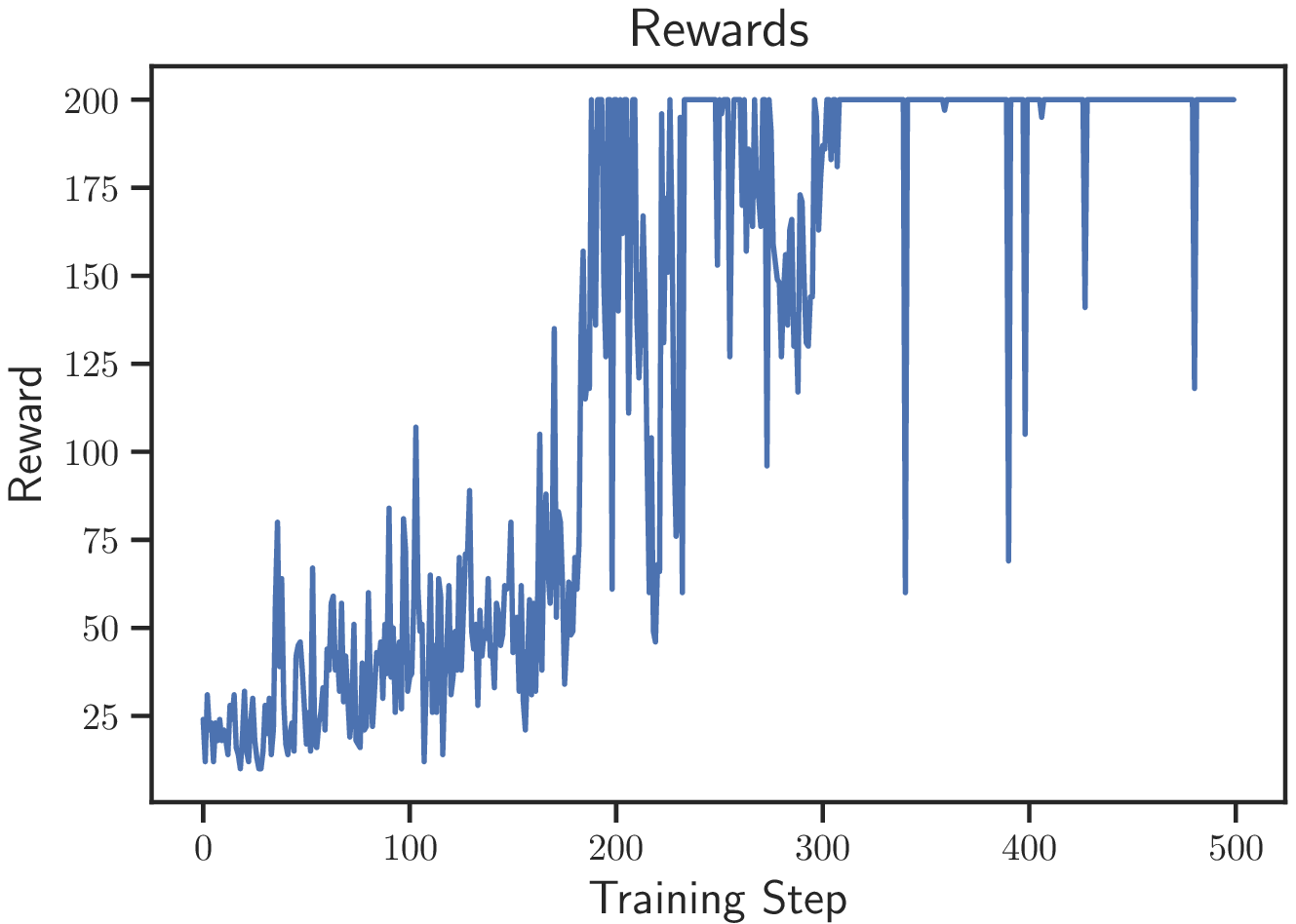}\includegraphics[width=0.43\textwidth]{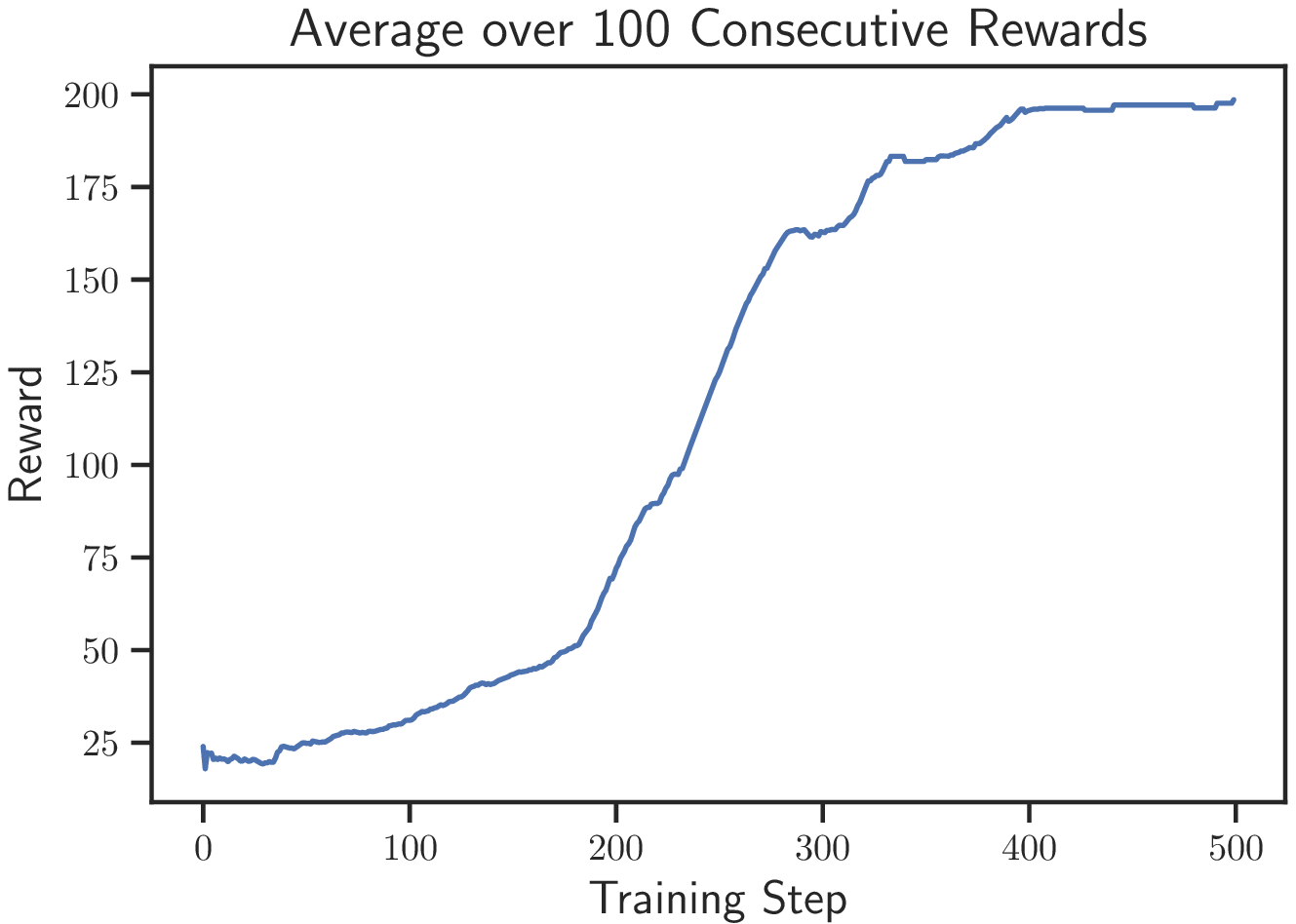}
\par\end{centering}
\caption{Left: visualization of a CartPole. Middle and Right: visualization
of the reward curve using the best found parameter value $\protect\bx^{*}$.
We have used the Advantage Actor Critic (A2C) algorithm to solve the
CartPole problem. The known optimum value is $f^{*}=200$.\label{fig:A2C_Reward}}
\end{figure*}

We use the advantage actor critic (A2C) \citep{Sutton_1998Reinforcement}
as the deep reinforcement learning algorithm to solve the CartPole
problem \citep{barto1983neuronlike}. This A2C is implemented in Tensorflow
\citet{abadi2016tensorflow} and run on a NVIDIA GTX 2080 GPU machine.
In A2C, we use two neural network models to learn $Q(s,a)$ and $V(s)$
separately. In particular, we use a simple neural network architecture
with $2$ layers and $10$ nodes in each layer. The range of the used
hyperparameters in A2C and the found optimal parameter are summarized
in Table \ref{tab:Hyperparameters-of-A2C}.

We illustrate the reward performance over $500$ training episodes
using the found optimal parameter $\bx^{*}=\argmax{\bx\in\mathcal{X}}f(\bx)$
value in Fig. \ref{fig:A2C_Reward}. In particular, we plot the raw
reward and the average reward over 100 consecutive episodes - this
average score is used as the evaluation output. Our A2C with the found
hyperparameter will take around $300$ episodes to reach the optimum
value $f^{*}=200$.

\begin{table}
\centering{}\caption{Hyperparameters of Advantage Actor Critic.\label{tab:Hyperparameters-of-A2C}}
\begin{tabular}{cccc}
\toprule 
Variables & Min & Max & Best Parameter $\bx^{*}$\tabularnewline
\midrule
$\gamma$ discount factor & $0.9$ & $1$ & $0.95586$\tabularnewline
learning rate $q$ model & $1e^{-6}$ & $0.01$ & $0.00589$\tabularnewline
learning rate $v$ model & $1e^{-6}$ & $0.01$ & $0.00037$\tabularnewline
\bottomrule
\end{tabular}
\end{table}

\subsection{Comparison using vanilla GP and transformed GP}

We empirically compare the proposed transformed Gaussian process (using
the knowledge of $f^{*})$  and the vanilla Gaussian process \citep{Rasmussen_2006gaussian}
as the surrogate model for Bayesian optimization. We then test our
ERM and EI on the two surrogate models. After the experiment, we learn
that the transformed GP is more suitable for our ERM while it may
not be ideal for the EI.

\paragraph{ERM. }

\begin{figure*}
\includegraphics[width=0.5\textwidth]{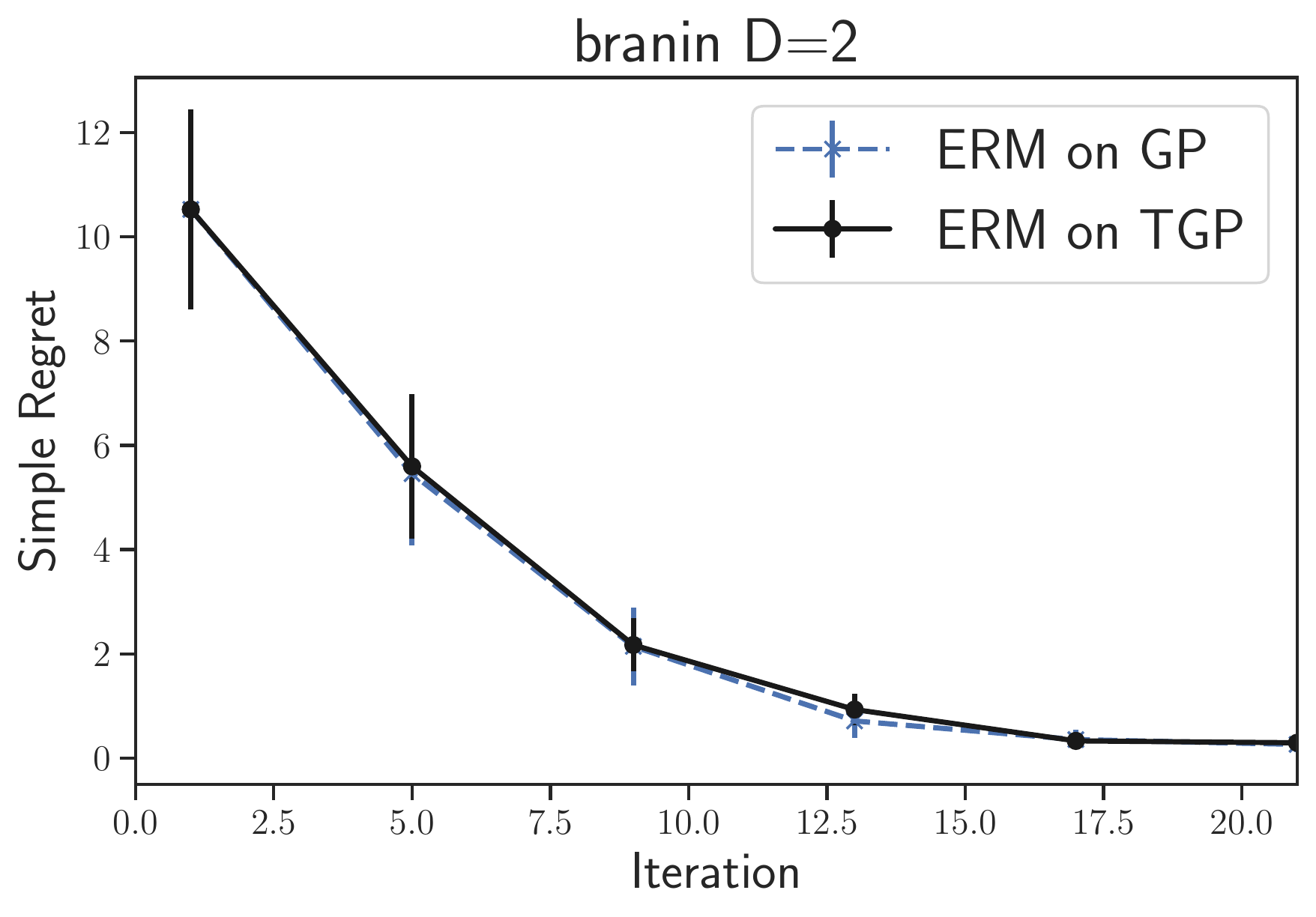}\includegraphics[width=0.5\textwidth]{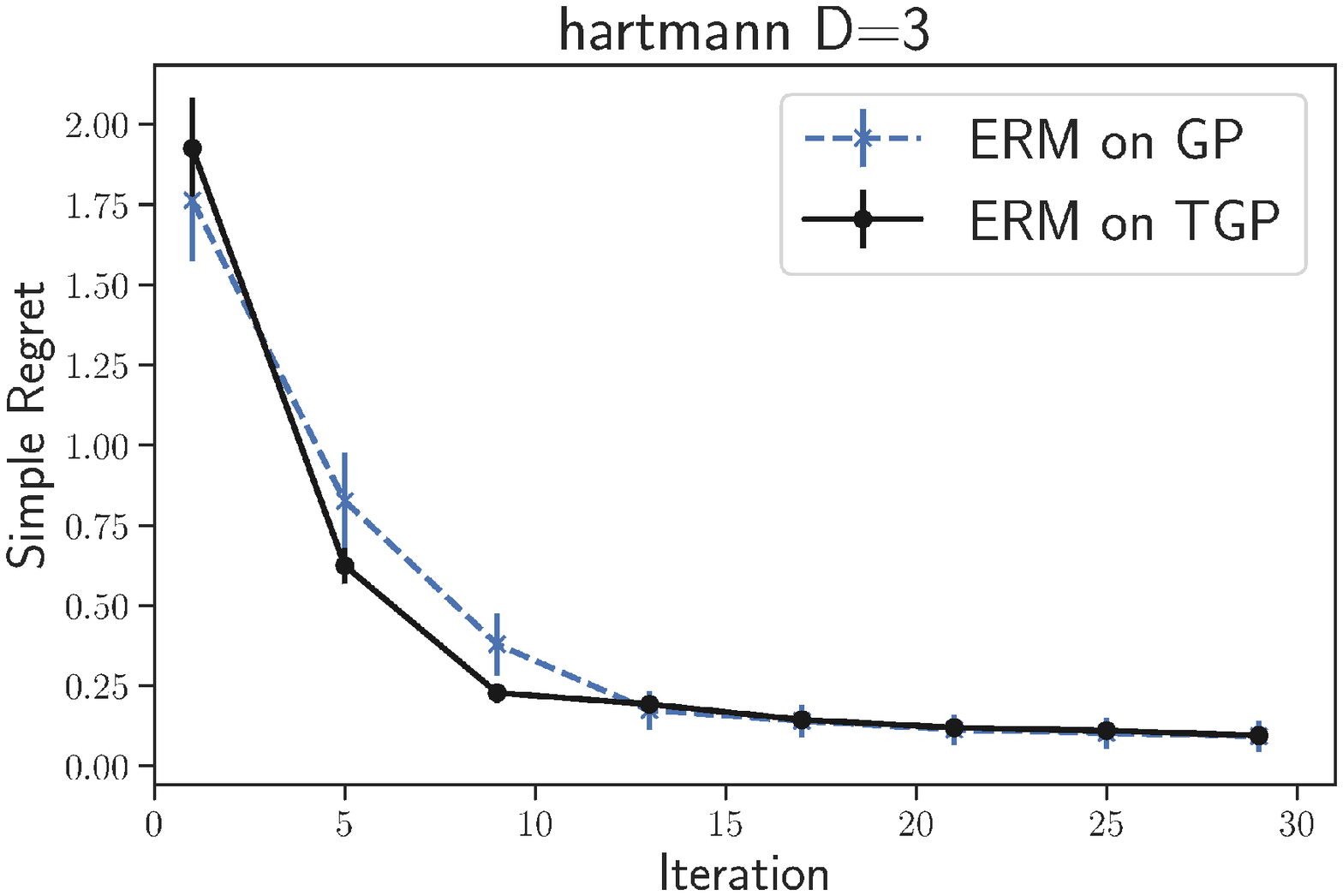}

\includegraphics[width=0.49\textwidth]{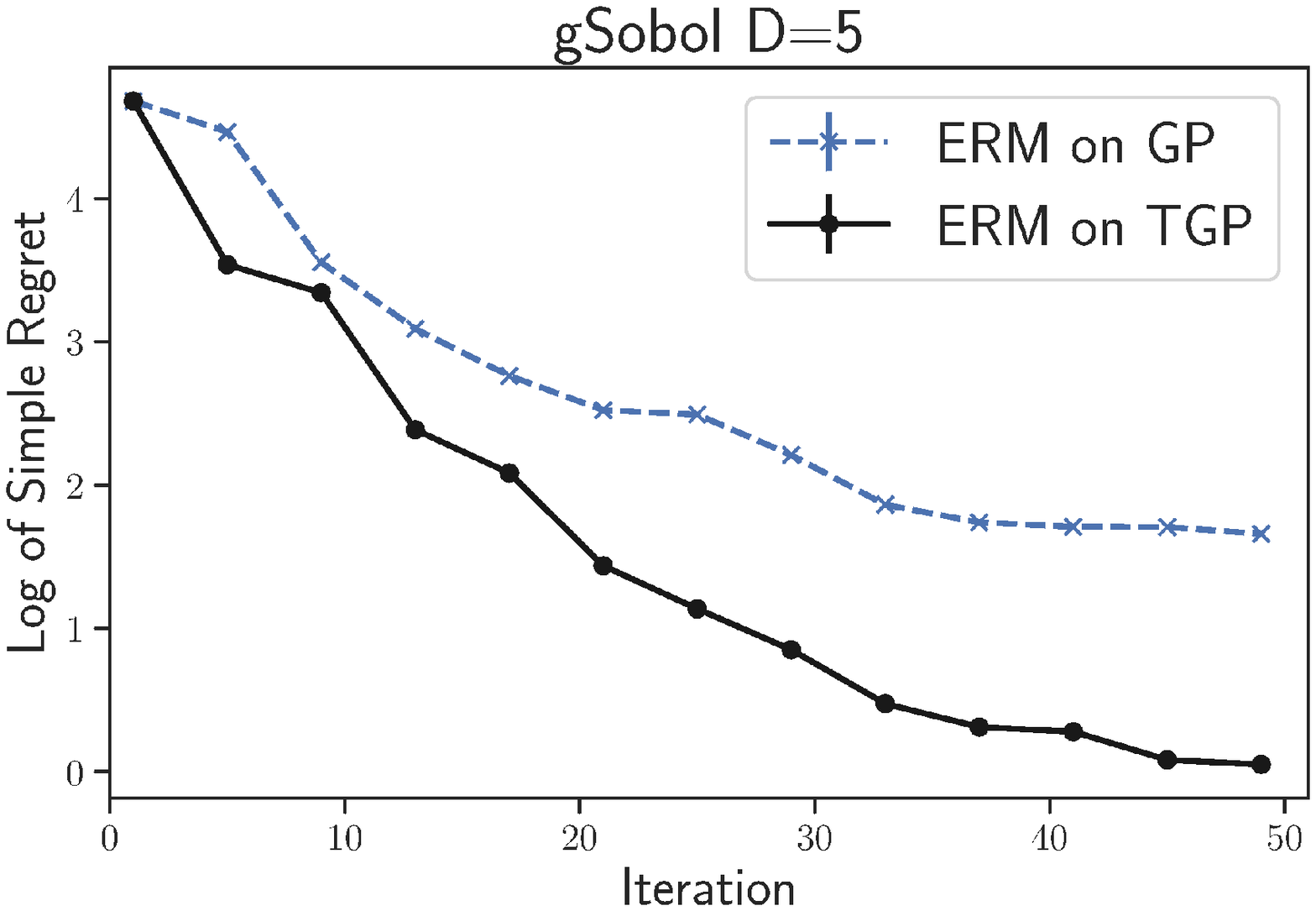}\includegraphics[width=0.5\textwidth]{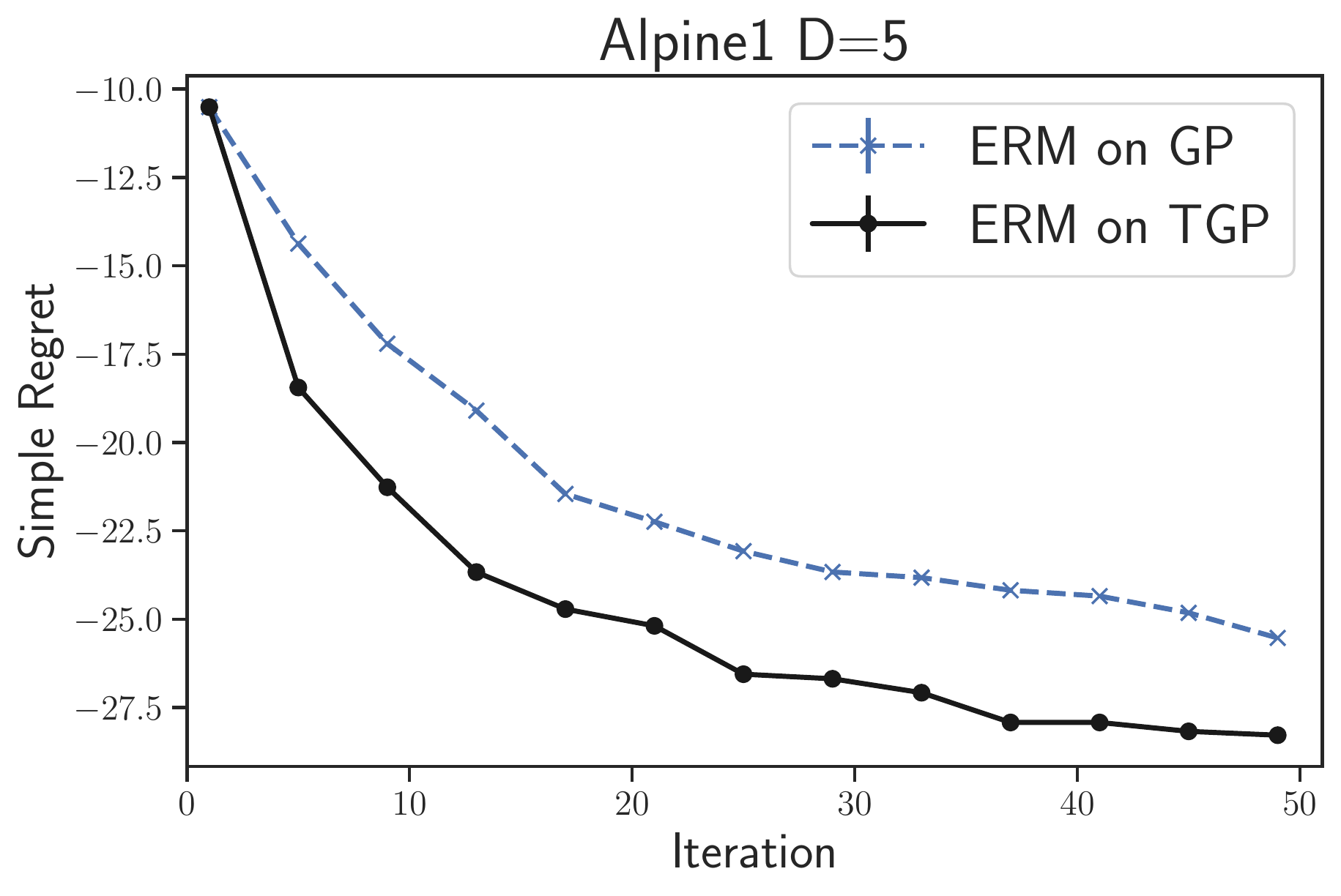}

\includegraphics[width=0.5\textwidth]{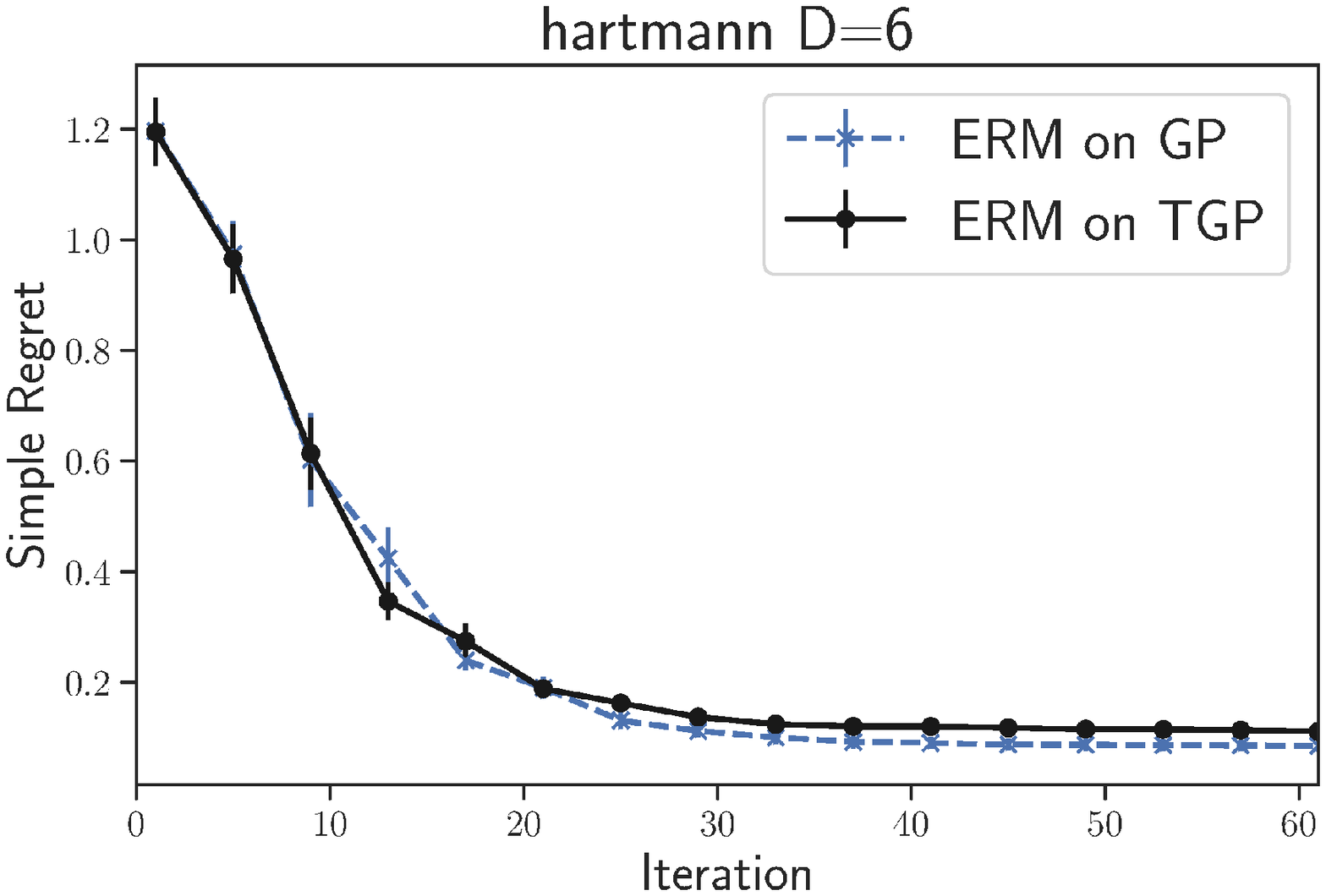}\includegraphics[width=0.5\textwidth]{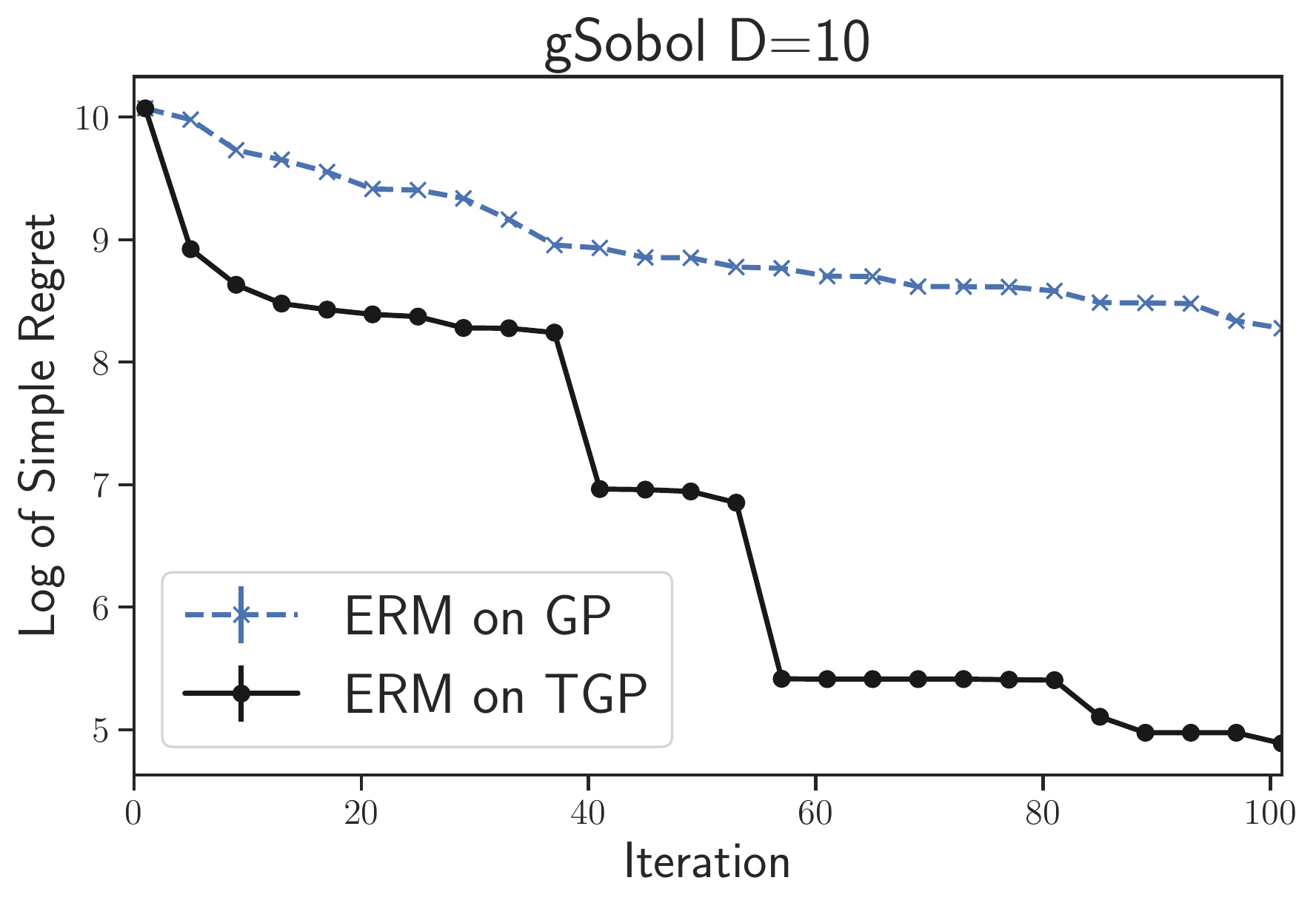}

\caption{Experiments with ERM acquisition function on vanilla Gaussian process
(GP) and transformed Gaussian process (TGP). Our acquisition function
using the transformed GP consistently performs better than using the
vanilla GP. Particularly, the TGP will be more useful in high-dimensional
functions of Alpine1 $D=5$ and gSobol $D=5$, $D=10$ functions.
In these functions, ERM on TGP will outperform ERM on GP by a wide
margin. \label{fig:Experiments-on-ERM_TGP}}
\end{figure*}

We perform experiments on ERM acquisition function using two surrogate
models as vanilla Gaussian process (GP) and transformed Gaussian process
(TGP). Our acquisition function performs better with the TGP. The
TGP exploits the knowledge about the optimum value $f^{*}$ to construct
the surrogate model. Thus, it is more informative and can be helpful
in high dimension functions, such as Alpine1 $D=5$ and gSobol $D=5$,
$D=10$, in which the ERM on TGP achieves much better performance
than ERM on GP. On the simpler functions, such as branin and hartmann,
the transformed GP surrogate achieves comparable performances with
the vanilla GP. We visualize all results in Fig. \ref{fig:Experiments-on-ERM_TGP}.

\paragraph{Expected Improvement (EI).}

\begin{figure*}
\includegraphics[width=0.5\textwidth]{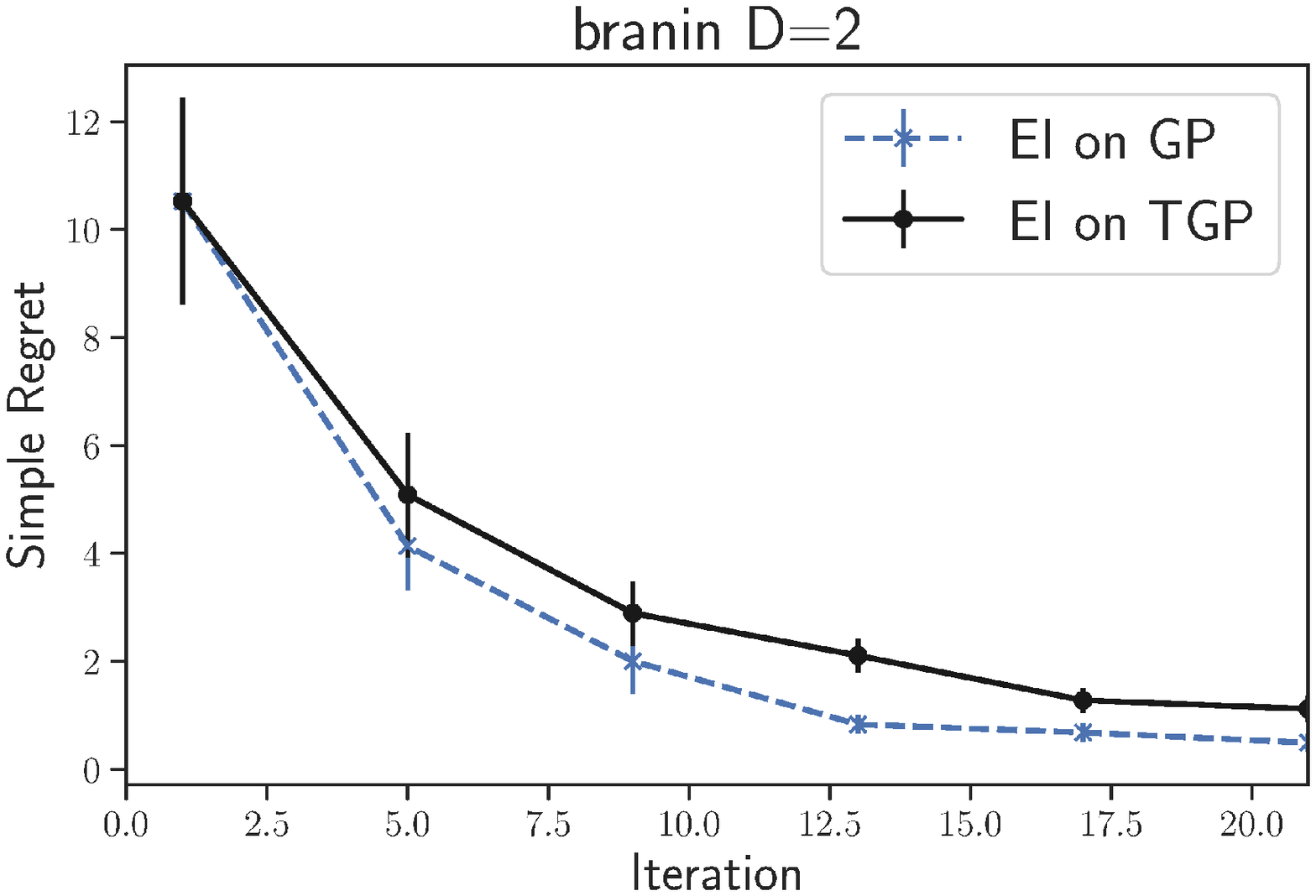}\includegraphics[width=0.5\textwidth]{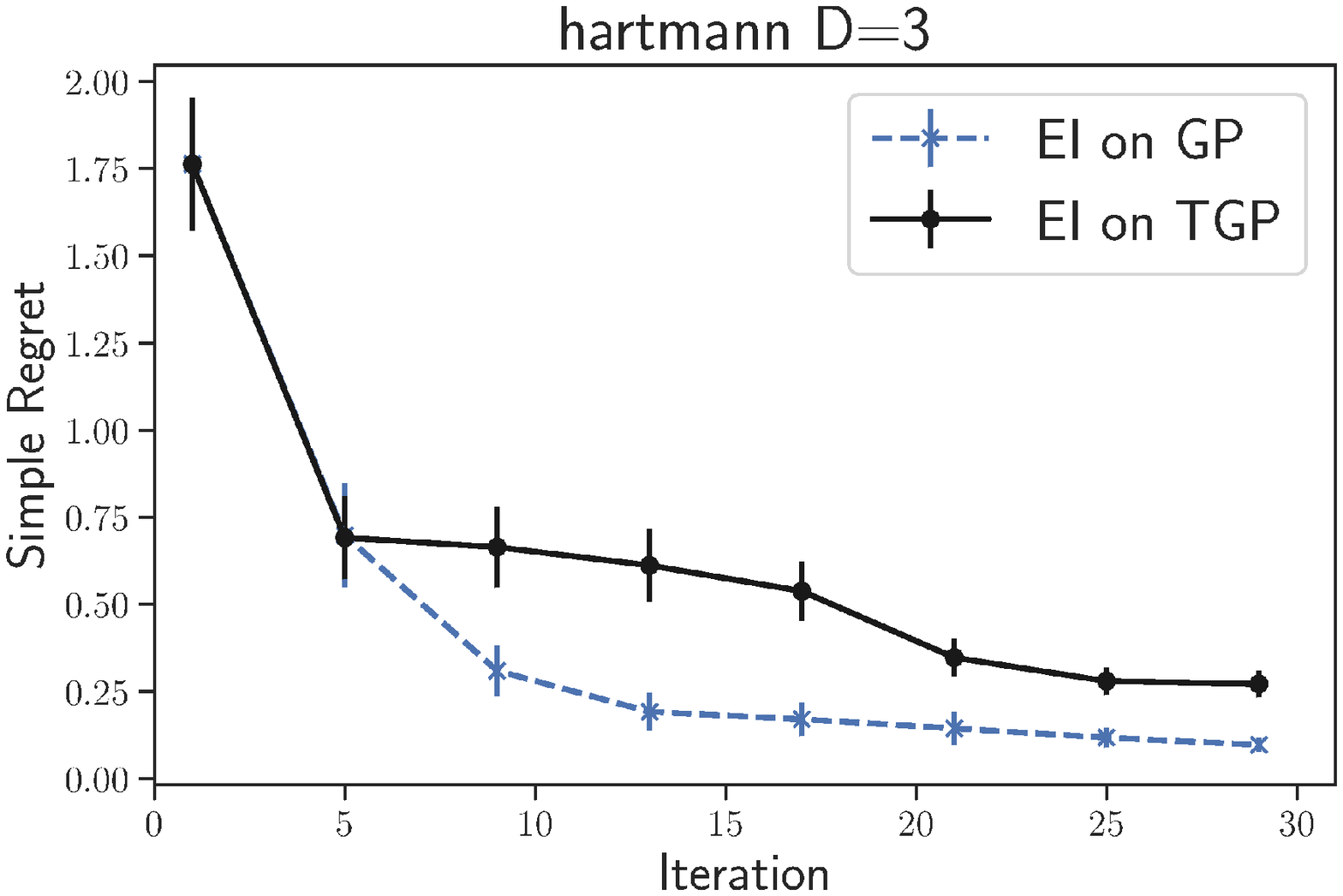}

\includegraphics[width=0.5\textwidth]{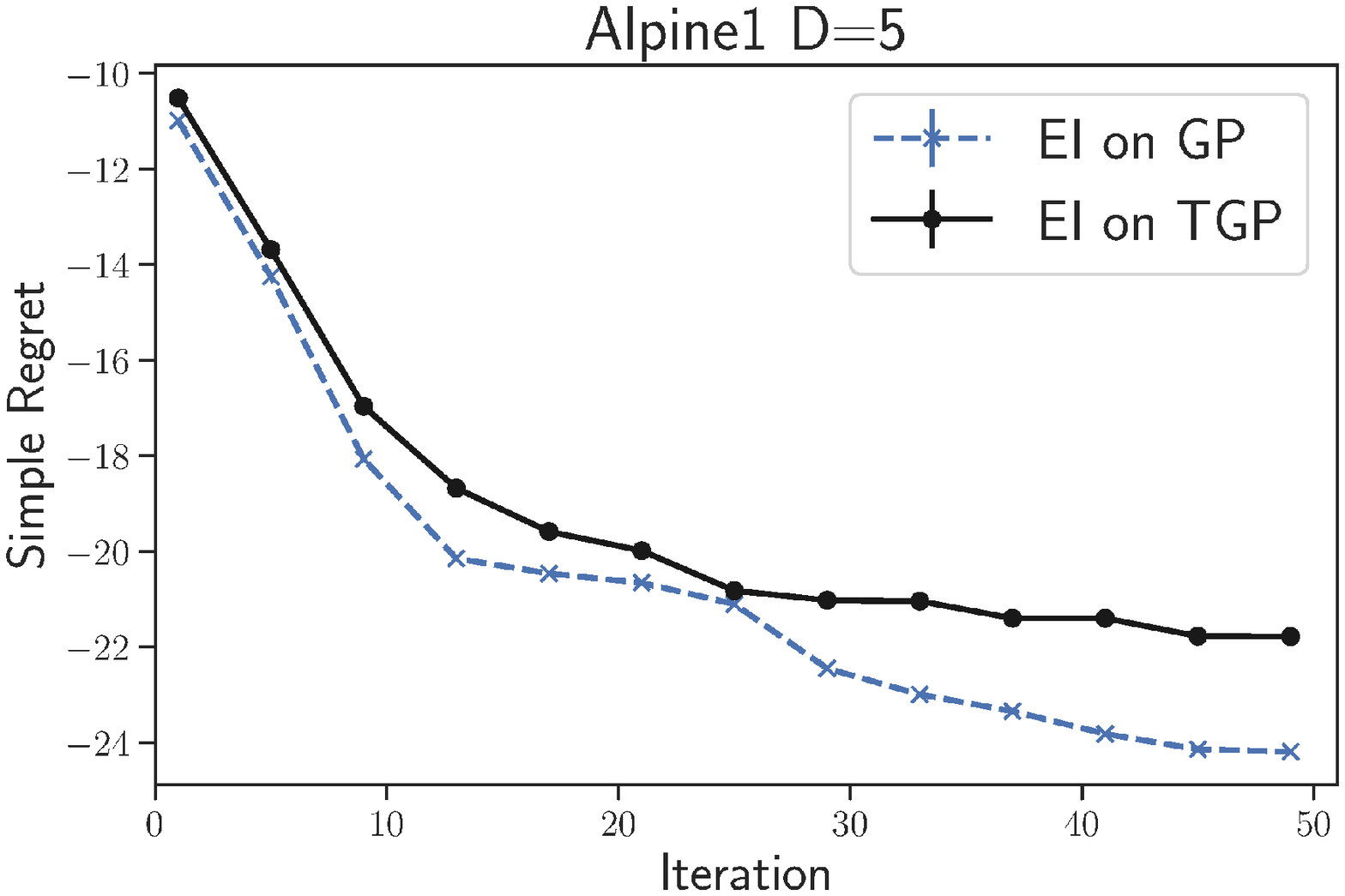}\includegraphics[width=0.5\textwidth]{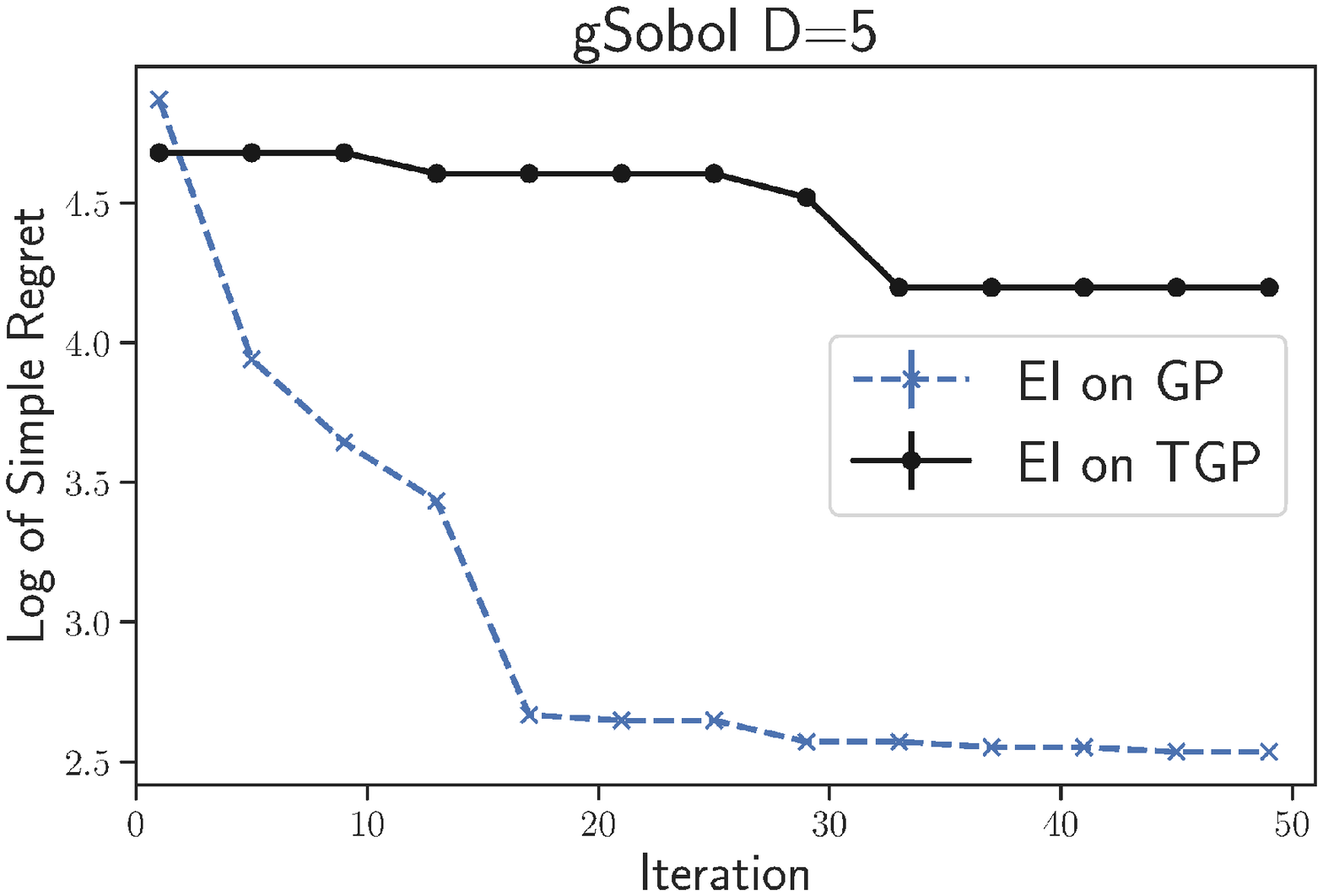}

\includegraphics[width=0.5\textwidth]{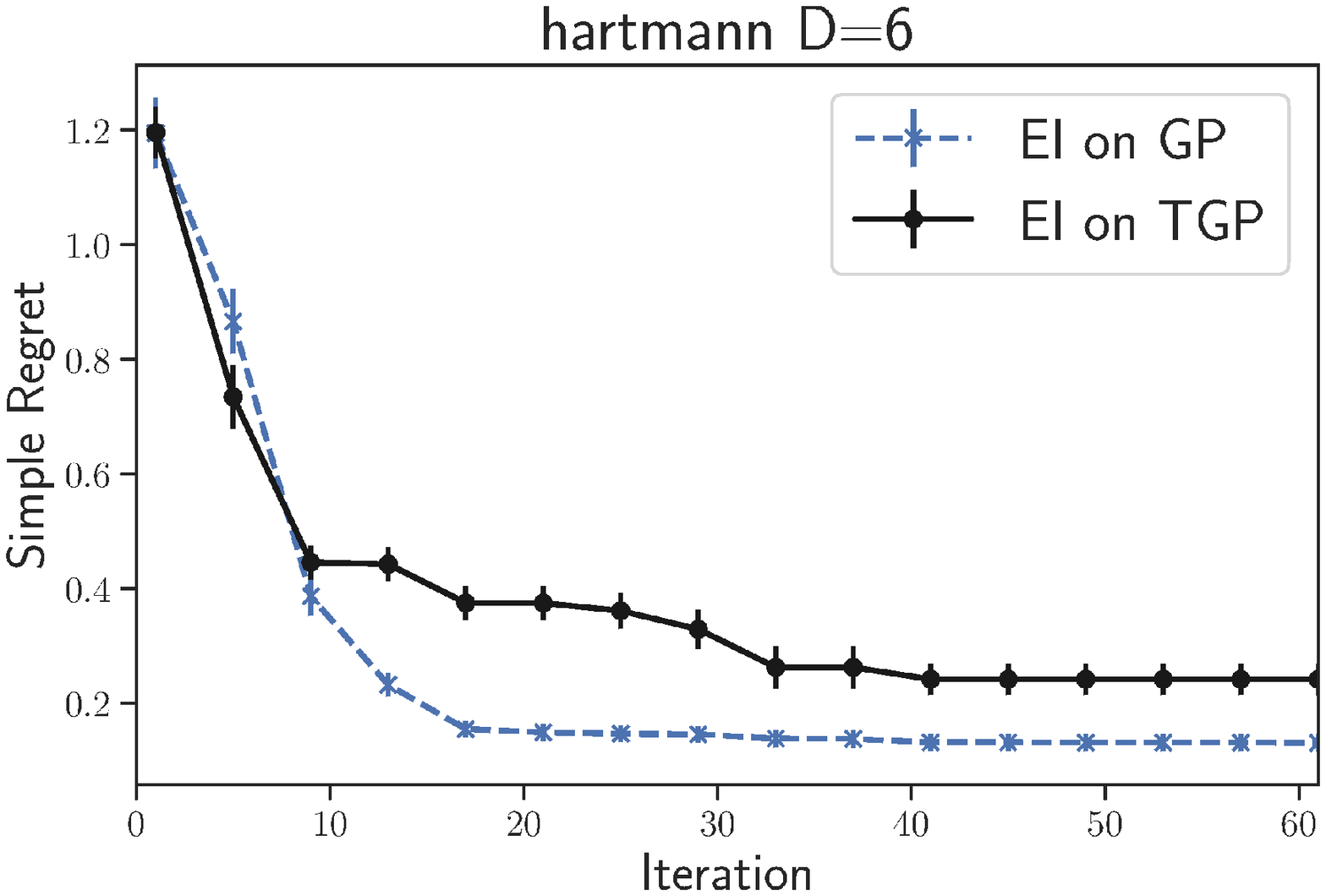}\includegraphics[width=0.5\textwidth]{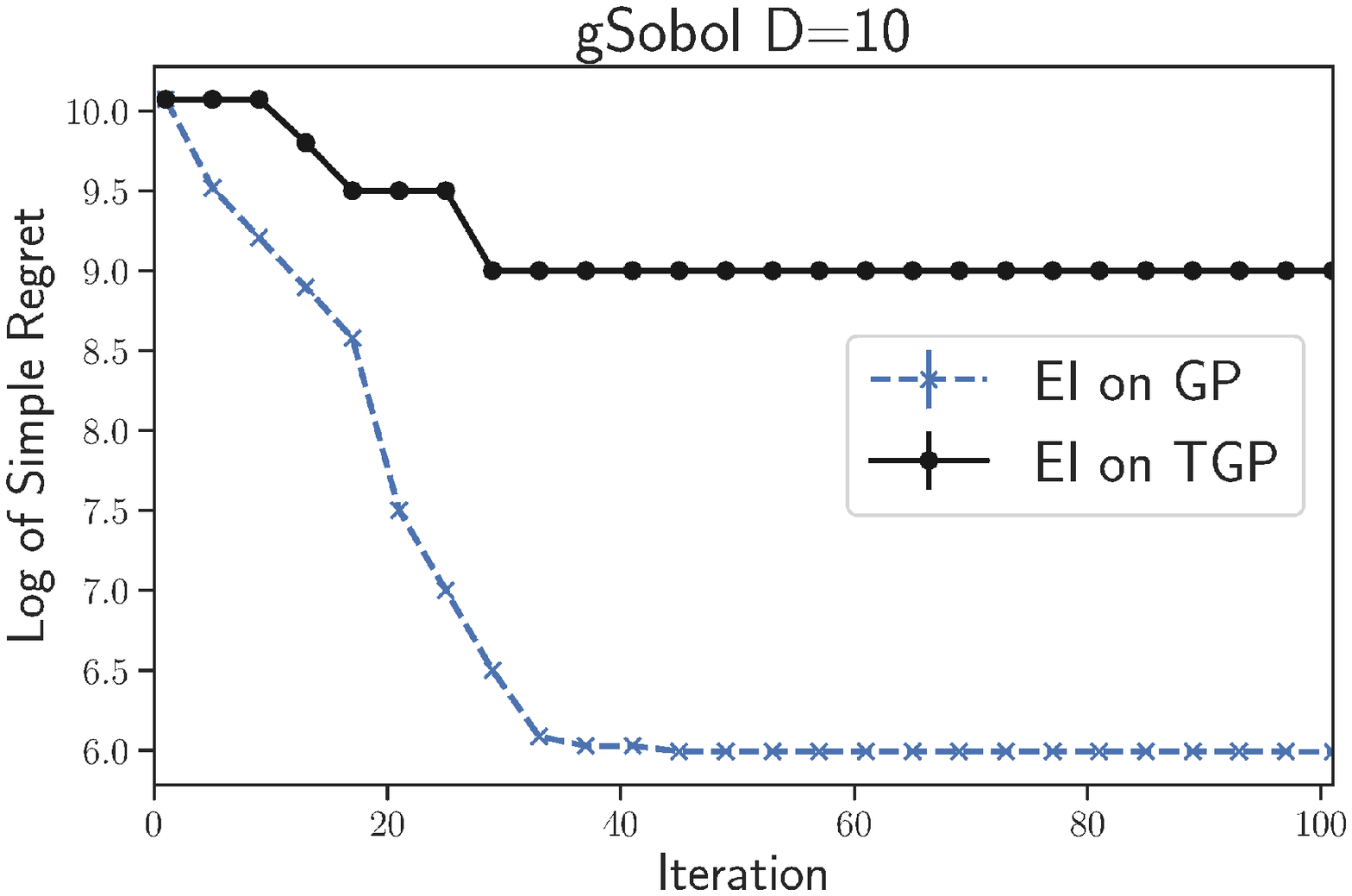}

\caption{Experiments with EI acquisition function using the surrogate models
as GP and TGP. Although the TGP exploits the knowledge about the optimum
value $f^{*}$ to construct the informed surrogate model, it brings
the side effect of transformation in making additional GP predictive
uncertainty. As a result, the EI will explore more aggressively using
TGP and thus obtain worse performance comparing to the case of using
vanilla GP. \label{fig:Experiments-with-EI_TGP}}
\end{figure*}

We then test the EI acquisition function on two surrogate models of
vanilla Gaussian process and our transformed Gaussian process (using
$f^{*}$) in Fig. \ref{fig:Experiments-with-EI_TGP}. In contrast
to the case of ERM above, we show that the EI will perform well on
the vanilla GP, but not on the TGP. This can be explained by the side
effect of the GP transformation as follows. From Eq. (1) in the main
paper, when the location has poor (or low) prediction value $\mu(\bx)=f^{*}-\frac{1}{2}\mu_{g}^{2}(\bx)$,
we will have large value  $\mu_{g}(\bx)$. As a result, this large
value of $\mu_{g}(\bx)$ will make the uncertainty larger $\sigma(\bx)=\mu_{g}(\bx)\sigma_{g}(\bx)\mu_{g}(\bx)$
from Eq. (2) in the main paper. Therefore, TGP will make an additional
uncertainty $\sigma(\bx)$ at the location where $\mu(\bx)$ is low.

Under the additional uncertainty effect of TGP, the expected improvement
may spend more iterations to explore these uncertainty area and take
more time to converge than the case of using the vanilla GP. We note
that this effect will also happen to the GP-UCB and other acquisition
functions, which rely on exploration-exploitation trade-off.

In high dimensional function of gSobol $D=10$, TGP will make the
EI explore aggressively due to the high uncertainty effect (described
above) and thus result in worse performance. That is, it keeps exploring
at poor region in the first $100$ iterations (see bottom row of Fig.
\ref{fig:Experiments-with-EI_TGP}).

\paragraph{Discussion.}

The transformed Gaussian process (TGP) surrogate takes into account
the knowledge of optimum value $f^{*}$ to inform the surrogate. However,
this transformation may create additional uncertainty at the area
where function value is low. While our proposed acquisition function
ERM and CBM will not suffer this effect, the existing acquisition
functions of EI and UCB will. Therefore, we only recommend to use
this TGP with our acquisition functions for the best optimization
performance.

\section{Other known optimum value settings}

To highlight the applicability of the proposed model, we list several
other settings where the optimum values are known in Table \ref{tab:Examples-of-knownoptimumvalue}.

\begin{table}
\begin{centering}
\caption{Examples of known optimum value settings.\label{tab:Examples-of-knownoptimumvalue}}
\begin{tabular}{ccc}
\toprule 
Environment & $f^{*}$ & Source\tabularnewline
\midrule
\midrule 
Pong & $18$ & Gym.OpenAI\tabularnewline
\midrule 
Frozen Lake & $0.79$ & Gym.OpenAI\tabularnewline
\midrule 
Inverted Pendulum v1 & $135.91$ & Gym.OpenAI\tabularnewline
\midrule 
CartPole & $200$ & Gym.OpenAI\tabularnewline
\bottomrule
\end{tabular}
\par\end{centering}
\end{table}

\end{document}